\documentclass{MITcsail}

\usepackage[utf8]{inputenc}
\usepackage[margin=1in]{geometry}
\graphicspath{{images/}}
\usepackage{graphicx}
\usepackage{amsmath}
\usepackage{amssymb}
\usepackage{booktabs}
\usepackage{hyperref}
\usepackage{cite}
\usepackage{natbib}
\usepackage{times}
\usepackage{setspace}
\usepackage{enumitem}
\usepackage{float}
\usepackage{caption}
\usepackage{xcolor}
\usepackage{multirow}   
\usepackage{array}      
\usepackage{longtable}  
\usepackage{booktabs}   
\usepackage{tabularx}        
\usepackage{booktabs}        
\usepackage{ragged2e}        
\usepackage{array}           

\usepackage{tikz}
\usetikzlibrary{positioning, shapes.geometric, arrows.meta,shapes.geometric, arrows, positioning, fit, calc} 
\usepackage{orcidlink}
\usepackage{wrapfig}
\usepackage{wasysym}
\usepackage{marvosym}
\usepackage{attachfile2}
\usepackage{prisma-flow-diagram}

\PassOptionsToPackage{table}{xcolor}
\usepackage{xcolor}

\usepackage[table]{xcolor} 
\definecolor{tealcol}{RGB}{0,128,128} 
\usepackage{siunitx}

\usepackage{makecell}
\usepackage{enumitem} 
 \usepackage{longtable}

\usepackage{placeins}

\title{Exploring the Relationship between Brain Hemisphere States and Frequency Bands through Classical Machine Learning and Deep Learning Optimization Techniques with Neurofeedback}

\author
{
Robiul Islam~$^{1}~\orcidlink{0000-0002-3704-8409}$\footnote{Correspondence E-mail: r\_islam@ieee.org},  
Dmitry I. Ignatov~$^{2}~\orcidlink{0000-0002-6584-8534}$, 
Karl Kaberg~$^{1}$, 
Roman Nabatchikov~$^{2}$~\orcidlink{0009-0004-7587-2978}\\
\vspace{1em} 
\normalfont{\small $^{1}$Innopolis University, Universitetskaya, 1, Innopolis, 420500, Russia}\\
\normalfont{\small $^{2}$Faculty of Computer Science, Higher School of Economics, Pokrovsky Boulevard, 11, Moscow, Russia} \vspace{2em}
}

\begin{document}

\maketitle
\thispagestyle{firstpagestyle} 

\begin{abstract}
This study investigates the performance of classifiers across EEG frequency bands, evaluating efficient class prediction for the left and right hemispheres using various optimisers. Three neural network architectures a deep dense network, a shallow three-layer network, and a convolutional neural network (CNN) are implemented and compared using the TensorFlow and PyTorch frameworks. Adagrad and RMSprop optimisers consistently outperformed others across frequency bands, with Adagrad excelling in the $\beta$ band and RMSprop achieving superior performance in the $\gamma$ band. Classical machine learning methods (Linear SVM and Random Forest) achieved perfect classification with 50--100${\times}$ faster training times than deep learning models. However, in neurofeedback simulations with real-time performance requirements, the deep neural network demonstrated superior feedback-signal generation (a 44.7\% regulation rate versus 0\% for classical methods). SHAP analysis reveals the nuanced contributions of EEG frequency bands to model decisions. Overall, the study highlights the importance of selecting a model dependent on the task: classical methods for efficient offline classification and deep learning for adaptive, real-time neurofeedback applications.

\textbf{Keywords:} Interpretable Machine Learning, EEG, Frequency Bands, Optimizers, Neurofeedback, Brain-Computer Interfaces
\end{abstract}

\section{Introduction}\label{sec:introduction}

The brain is divided into two hemispheres, the left and the right~\citep{sperry1975left}. Each hemisphere has specific functions and controls the opposite side of the body. The left hemisphere is generally associated with language, logic, and analytical thinking. It also controls the right side of the body. The right hemisphere is associated with creativity, intuition, and spatial awareness, and controls the left side of the body~\citep{mccrea2010intuition}. Both hemispheres work together and communicate through the corpus callosum, allowing for integrated functioning and coordination of various cognitive and motor functions.

Neuroscience has studied the relationship between brain hemisphere states and frequency bands. Delta waves are linked to deep sleep and memory consolidation, theta waves to cognitive processing and spatial navigation, alpha waves to relaxation, beta waves to sensorimotor functions, and gamma waves to perceptual binding and conscious awareness and memory~\citep{muhl2014survey}. Deep learning~\citep{raza2025deep} optimization can help explore this connection further.

In our previous work~\citep{islam2022explainable}, we used intensity-based classification on a visual perception dataset. This previous study classified brain hemispheres on a dataset originating from 5 individuals, while the present dataset spans 10 individuals. The pre-processing methods to filter frequency and learning model are analogous to our previous work with minor modifications of the architecture. Here we introduced batch normalization~\citep{bjorck2018understanding} (in addition to drop-out) to every layer to avoid overfitting as a machine learning problem~\citep{ying2019overview}.

\begin{figure}[!ht]
    \begin{minipage}{0.45\textwidth}
        \centering
        \includegraphics[width=0.5\textwidth]{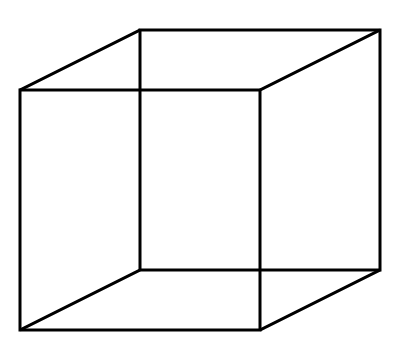}
        \caption{A line drawing of a transparent cube (known as the \textbf{Necker cube}), with opposite sides drawn parallel, so that the perspective is ambiguous}
        \label{fig:cube}
    \end{minipage} \hfill
    \begin{minipage}{0.45\textwidth}
        \centering
        \includegraphics[width=0.4\textwidth]{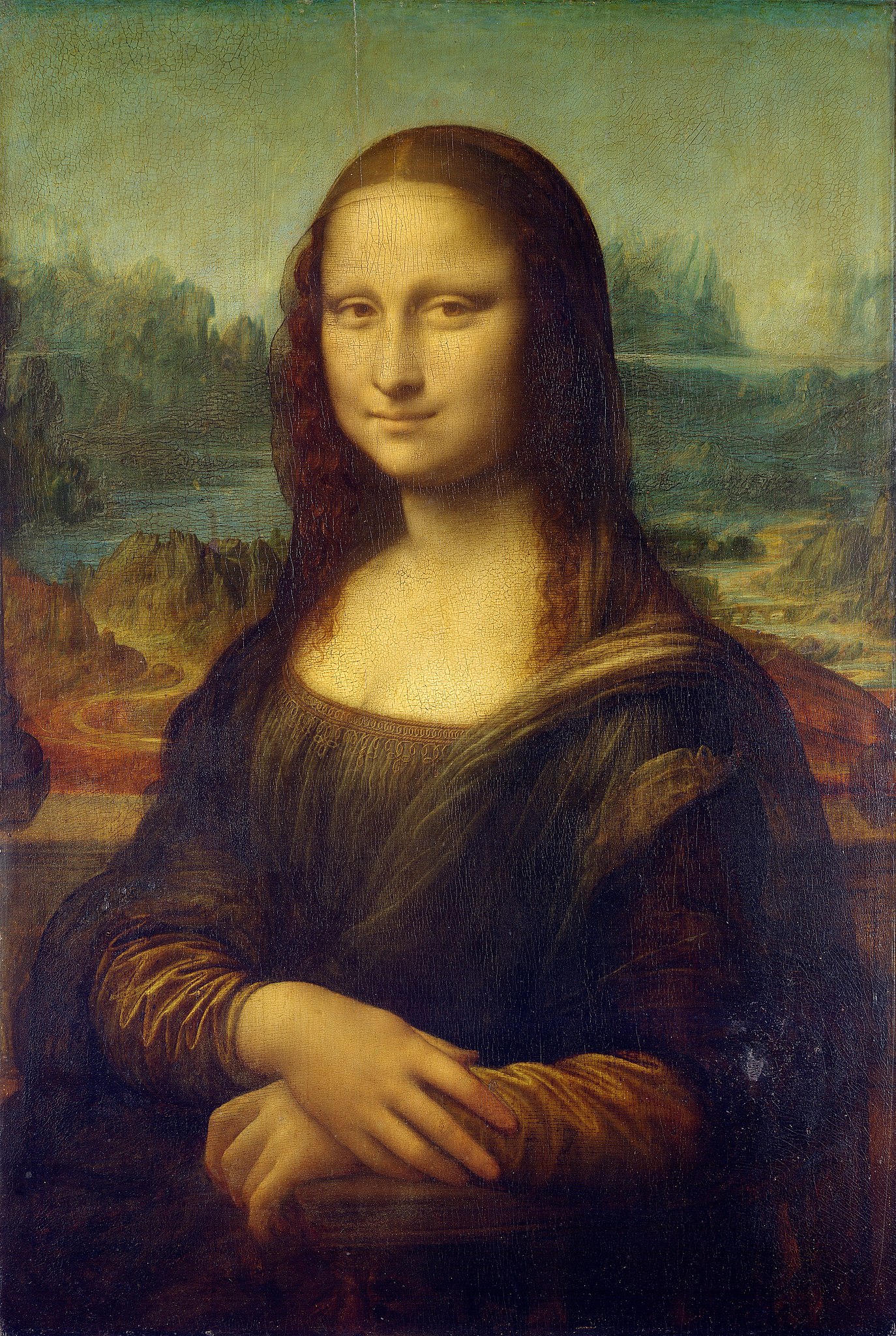}
        \caption{Leonardo da Vinci, Portrait of Lisa Gherardini (known as the \textbf{Mona Lisa}), c. 1503--19, oil on poplar panel}
        \label{fig:mona}
    \end{minipage}
\end{figure}

The hemispheres interact with each other when the complexity of sensory information increases, and this interaction varies with different frequency bands. To test this hypothesis, we are using machine learning methods.
\begin{itemize}
    \item Increasing the area of interaction (frequency band) will increase classification accuracy.
    \item It is important to note that the classification matrix varies depending on the complexity of the stimuli, with simple sensory matrix stimuli (the Necker cube in Figure~\ref{fig:cube}) differing from complex stimuli (Mona Lisa in Figure~\ref{fig:mona}) which involve multiple colours and emotions.
\end{itemize}

In order to test the hypothesis that the hemispheres interact with the increasing complexity of sensory information across different EEG frequency bands in different ways, we suggest the following experimental approach summarised in a few subsections below.

\subsection{Experimental Design}

The experimental design entailed the collection of EEG data from participants while they were exposed to stimuli of varying complexity. The stimuli included both simple sensory stimuli, such as the Necker cube, and complex stimuli, such as the Mona Lisa painting, to elicit different levels of neural activity. EEG signals were recorded across multiple frequency bands, including delta, theta, alpha, beta, and gamma.

\subsection{Machine Learning Methods}

Machine learning methods were employed to classify the stimuli into their respective categories. This was achieved by training classifiers on the EEG data. The classification models were optimized~\citep{sun2020optimization} using various optimization algorithms, including Adagrad, Adadelta, SGD, Adam, RMSprop, Nadam, AdaMax, and FTRL, taken from state-of-the-art packages including TensorFlow and PyTorch.

\subsection{Evaluation Metrics}

The performance of classifiers across different frequency bands was evaluated using classification accuracy, precision, recall, F1, and ROC-AUC scores. To ensure the robustness of the results, various randomization techniques were employed.

\subsection{Hypothesis Testing}

The hypothesis that increasing the area of interaction (frequency band) increases classification accuracy was tested by analyzing the correlation between EEG frequency bands and classifier performance. Alpha-band oscillatory activity in the parietal-occipital regions has been linked to vigilance, attention, cognitive processing, and cortical communication in both healthy individuals and patients with brain damage. Studies have demonstrated that the delta and theta frequency bands are associated with deep sleep, meditation, and cognitive processing of spatial navigation and memory consolidation. Increased delta and theta power have been observed during tasks requiring sustained attention and working memory. The alpha frequency band is associated with relaxed wakefulness and decreased sensory processing, with power decreases observed during active cognitive tasks. Beta oscillations are associated with active cognitive processing, sensorimotor integration, and motor planning. Increased beta power is linked to attentional engagement, motor imagery, and anticipation of movement. Gamma oscillations are associated with higher cognitive functions, including perceptual binding, feature integration, and conscious awareness. This reflects the coordinated activity of neuronal populations involved in complex cognitive tasks~\citep{gschwandtner2021dynamic,fernandez2000specific,lasaponara2019hemispheric}.

The findings demonstrate that the different EEG frequency bands are indeed associated with distinct patterns of neural activity. Furthermore, the interaction and modulation of these frequency bands are influenced by the complexity of the sensory information being processed. The results provide strong support for the hypothesis presented in the question.

\subsection{Verification}

To verify the hypotheses, we compared the results of our experiments with existing literature on EEG-based classification and neural processing of sensory information. Our findings were further validated against prior work using the same dataset across different model configurations and participant samples.

\subsection{Visual Representation and Interpretable Machine Learning}

To illustrate the contributions of EEG frequency bands to classifier decisions, SHAP (SHapley Additive exPlanations) plots were employed. This analysis provided insights into the neural mechanisms underlying classification outcomes.

The objective of this study was to provide empirical evidence supporting the hypothesis that the hemispheres interact differently in response to varying levels of sensory information complexity. To this end, we systematically investigated the relationship between EEG frequency bands, sensory stimulus complexity, and classifier performance.

\subsection{Method}

In this research work, we explored the application of various state-of-the-art deep learning optimization techniques to establish a robust and reliable relationship between frequency band and hemisphere state (encoded as a binary class across the recorded signals)~\citep{marblestone2016toward,tan2021evolutionary,anter2022real}. Through experimentation and analysis, we identified three highly effective optimizers that demonstrated high performance across all frequency bands. Optimizers play a crucial role in various fields, including machine learning, neuroscience, and optimization problems, due to their ability to efficiently search for optimal solutions in complex and high-dimensional spaces; this is especially important for deep learning with high-dimensional data where loss functions may have complex landscapes sometimes similar to fractal geometry~\citep{Garipov:18}. Previous research and findings have highlighted the importance of using optimizers to improve performance~\citep{pawan2022particle,metz2022practical,khan2020bas}, convergence speed, and overall effectiveness of algorithms.

The contribution of the study is the following:
\begin{itemize}
    \item Our approach involves using machine learning techniques to examine the intricate EEG data and determine the most appropriate neural network architecture and optimization method for all frequency bands for our dataset.
    \item Each frequency band is associated with specific hemisphere states, which can be analyzed through EEG data to gain insight.
\end{itemize}

\subsection{Classical Machine Learning Methods}

Deep learning (DL) is increasingly applied to EEG analysis, with researchers exploring various optimizers to improve performance~\citep{ding2015deep,robinson2019eeg}. There are two important questions that still need answers: (1) Is deep learning much better than traditional machine learning for EEG classification tasks? (2) Which EEG frequency bands best show the differences between brain hemispheres? Deep learning methods are complex and costly, so we need to test if they are really necessary.

Prior studies have either: (a) applied DL without comparing to classical baselines~\citep{islam2022explainable}, (b) focused solely on optimizer comparisons within DL frameworks~\citep{MerlinPraveena04072022}, or (c) lacked systematic evaluation across EEG frequency bands. No study has comprehensively compared DL (with multiple optimizers) against classical ML methods (SVM~\citep{han2012parameter,chauhan2019problem,mahmud2022air}, Random Forest~\citep{rigatti2017random}) across the entire EEG frequency spectrum for hemisphere classification.

This study addresses critical gaps in EEG hemisphere classification through systematic experimentation with 4 machine learning algorithms across 5 frequency bands (delta, theta, alpha, beta, gamma) using data from participants viewing ambiguous (Necker cube) and complex (Mona Lisa) visual stimuli. Our experimental design notably utilized a single, consistent optimizer for the Deep Neural Network---\textbf{RMSprop}---selected for its reliable convergence and stable performance across the entire frequency spectrum, providing a robust baseline for comparison. Our key findings challenge conventional assumptions:

\begin{itemize}
    \item \textbf{Algorithm Performance:} Classical ML methods demonstrate exceptional performance, with Linear SVM achieving perfect classification (100\% accuracy, $\kappa=1.0$) across multiple frequency bands, while requiring $\sim$50--100$\times$ less training time compared to the deep neural network with RMSprop (0.5--0.8\,s vs 30--150\,s).

    \item \textbf{Frequency Band Significance:} Beta and alpha bands show superior discriminative power with perfect classification (ROC-AUC=1.0) across all algorithms, while delta band exhibits more variability, suggesting frequency band selection is more critical than model complexity.

    \item \textbf{Stimulus-Type Dependence:} Performance varies by visual stimulus, with Mona Lisa stimuli generally yielding higher classification accuracy ($\geq$97.87\% across all bands) compared to Necker cube stimuli (91.49--100\%).
\end{itemize}

These findings have significant practical implications for EEG-based brain-computer interfaces and clinical diagnostics:

\begin{itemize}
    \item \textbf{Practical Implementation:} Practitioners should prioritize frequency band selection and simpler, interpretable models (Linear SVM, Random Forest) over complex deep learning architectures, especially given the substantial reduction in computational requirements without sacrificing performance.

    \item \textbf{Methodological Guidance:} The strong performance of classical methods, even when the DNN is optimized with a stable optimizer like RMSprop, underscores that model complexity is not a prerequisite for high accuracy in this task.

    \item \textbf{Clinical Applications:} The consistent high performance in beta and gamma bands ($\geq$99.82\% ROC-AUC) provides clear guidance for feature selection in diagnostic applications.

    \item \textbf{Resource Optimization:} We provide an evidence-based decision framework for model selection balancing accuracy requirements ($\geq$97.87\% achievable with classical ML), computational budget, and interpretability needs.

    \item \textbf{Open Science:} Complete experimental results across all frequency bands and algorithms are provided alongside open-source code for reproducibility and future benchmarking.

    \item \textbf{Neurofeedback Viability:} The neurofeedback simulation revealed critical implementation insights, with the Deep Neural Network (RMSprop) demonstrating the highest target achievement rates (up to 44.7\%) and strongest feedback signals (avg. strength: 0.87 in beta band), while classical methods predominantly generated neutral or negative feedback, highlighting the DNN's superior suitability for real-time BCI applications despite its longer training time.
\end{itemize}

The paper is structured as follows: Section~\ref{sec:Relatedwork} provides an overview of related work on hemispheres and batch normalization. In Section~\ref{sec:DataCollection}, we present our new expanded dataset~\citep{newdataset,robiul_eeg_visual_stimuli_2025}, which was previously described in part in our earlier works~\citep{islam2021interpretable,islam2022explainable}. In Section~\ref{sec:Models}, we describe the deep learning models used in this study. Section~\ref{sec:result} presents the experimental results, while Section~\ref{sec:Discussion} provides an evaluation and comparison of these results. Finally, in Section~\ref{sec:Conclusion}, we present our conclusions.

\section{Related work} \label{sec:Relatedwork}

We summarize related works in four topical subsections below. For a broader discussion on the applications of neural networks across various domains--particularly in EEG analysis and medicine, among others--involving complex data signals, refer to~\citep{Man:2024}.

\subsection{EEG and language comprehension}

Related studies of brain activity with EEG show similar-sized samples and reliance on various types of signals (waves) for investigation of various activities of hemispheres, for example, for perceptions of words by healthy individuals or autist children.  

Thus one study~\citep{yurchenko2023register} examined the effect of register switching on language comprehension using event-related potentials (ERPs). The study concentrated on the N400 effect, which is a negative deflection in brainwave activity occurring approximately 400 milliseconds after a stimulus and associated with semantic processing. The study put forth the proposition that, in contrast to other pragmatic factors such as metaphor, irony, and humour, register switching does not entail the resolution of conflicting meanings or the creation of new interpretations. The study revealed that processing sentences with register switching, akin to processing sentences with semantic incongruity, evoked an N400 effect. This indicates that integrating words from a disparate register than the preceding context necessitates supplementary lexical-semantic processing, analogous to the processing required for semantically incongruent words.

A further study~\citep{arutiunian2023event} examined event-related desynchronisation (ERD) of alpha-band oscillations in children diagnosed with autism spectrum disorder (ASD). Alpha-band ERD is defined as a reduction in alpha power observed in electroencephalography (EEG) and magnetoencephalography (MEG) recordings. This phenomenon is thought to reflect the involvement of neural tissue in information processing. This study concentrated on alpha ERD during the simultaneous presentation of auditory and visual stimuli in children with ASD, with the hypothesis that children with ASD would exhibit altered alpha ERD in one of the sensory domains due to the common co-occurring condition of attention deficit in ASD. The study employed magnetoencephalography (MEG) to investigate alpha event-related desynchronization (ERD) in groups of 20 children with autism spectrum disorder (ASD) and 20 age-matched typically developing controls. Simple amplitude-modulated tones were presented concurrently with a fixation cross on the screen. The findings revealed that children with ASD exhibited a bilateral reduction in alpha-band ERD in the auditory cortex, but not in the visual cortex. This indicates that children with ASD may experience challenges in processing auditory information when presented with simultaneous visual stimuli.

\subsection{Visual Stimuli}

The human brain displays distinctive activation patterns in response to diverse visual stimuli, indicating the presence of a structured visual representation in the ventral temporal cortex~\citep{bagchi2022eeg}. A deep learning model, the EEG-ConvTransformer Network, is introduced as a means of classifying visual stimuli using EEG signals. The integration of a transformer and a convolutional neural network enables the model to capture both spatial similarities and temporal features. The model's performance was evaluated with varying parameters, including width and number of attention heads, demonstrating enhanced classification accuracy across five tasks.

Another study addresses the classification of emotions using EEG signals triggered by audio-visual stimuli~\citep{ahirwal2020audio}. A distinctive channel selection methodology is employed, founded upon the activation probability derived from a correlation matrix of EEG channels. The features are extracted from the time, frequency, and entropy domains, and the classification is performed using support vector machines (SVM), artificial neural networks (ANN), and Na\"ive Bayes (NB).

The effectiveness of Independent Component Analysis (ICA) in conjunction with machine learning for the classification of visual objects from single-trial EEG data has been demonstrated in research studies~\citep{stewart2014single}. The study employs support vector machines (SVM) to evaluate the classification outcomes based on scalp electroencephalogram (EEG) data and independent components (ICs). The results demonstrate that data from a single IC enhances the accuracy to 87\

Another study investigates the use of convolutional neural networks (CNNs) to classify visual stimuli from EEG signals in both healthy and Alzheimer's disease subjects~\citep{komolovaite2022deep}. The research examines whether CNNs can effectively distinguish between emotions, facial inversion, familiarity, and facial inversion, exploring the impact of pre-trained weights and data augmentation on performance.

Finally, the potential of visual imagery as a control mechanism for electroencephalogram-based brain-computer interfaces (BCIs) is investigated in the study by~\citep{kosmyna2018attending}. The objective of this study is to differentiate between visual imagery and observation tasks, as well as to distinguish them from resting states. The experiments were conducted with a total of 26 subjects, who were instructed to observe and imagine two distinct visual stimuli (a flower and a hammer).

Again, in these works with visual stimuli, we can observe similar-sized data and successful applications of advanced machine learning techniques.

\subsection{Left and right hemispheres}

The study~\citep{silva2024neglect} aimed to investigate if the National Institutes of Health Stroke Scale (NIHSS) equally predicts right hemisphere lesion volumes in patients with and without neglect, and if a modification of the neglect scoring rules could increase its predictive capacity. A total of 162 ischemic stroke patients were included, 108 with neglect and 54 without. The correlation between lesion volume and NIHSS was lower in patients with neglect, and neglect was a statistically significant covariate in the partial correlation analysis between NIHSS and lesion volume. Different modifications in the neglect scoring rules were applied, and it was found that with the neglect score tripled and with the duplication or triplication of all neglect modalities, the correlation was significantly higher than with the standard NIHSS. The study sample was divided into two groups: patients without neglect at admission (no neglect group) and patients with neglect at admission. The patient's demographic and clinical features were recorded, and the correlation between lesion volume and original NIHSS was found to be significant.

The study~\citep{binder1997human} used functional magnetic resonance imaging (fMRI) to identify language processing areas in the human brain, focusing on receptive language processing at both phonetic and semantic levels. The left hemisphere showed strong activation during the language task, indicating left hemisphere lateralization of blood oxygenation responses. Language areas were found almost exclusively in the left hemisphere or in the right cerebellum, challenging the classical model of language localization. The study provides a ``language map that differs from traditional neuroanatomical models of language processing.

The study~\citep{johansen2005functional} compared the relative volumes of the left and right hemispheres based on cytoarchitectonic data and diffusion-weighted imaging (DWI). Regression lines were shown for the left and right hemisphere data separately, indicating good agreement between the volumes. This suggests that the segmentation method was consistent across both hemispheres.

Here, we can admit that on the level of fMRI-based analyses, an abundance of expertise for brain areas screening is required as well as data of order magnitude higher than that in pure EEG studies.

\subsection{Batch Normalization in deep networks}

Batch Normalization (BN) has been found to exhibit greater tolerance to very large learning rates, leading to faster convergence along flat directions of the optimization landscape when used with Stochastic Gradient Descent (SGD). In the absence of BN, activations in deep networks can undergo dramatic growth with increasing depth, particularly when subjected to excessively large learning rates. Furthermore, it has been observed that random weight initialization may not be well-suited for networks with numerous layers unless BN is employed to enhance the network's resilience against ill-conditioned weights. Notably, BN applies consistent normalization to all activations within a given channel, achieved by subtracting the mean activation from all input activations in the channel and subsequently dividing the centered activation by the standard deviation. This normalization approach results in relatively constant average channel means and variances throughout the network, in contrast to unnormalized networks where these values grow almost exponentially with depth. Additionally, the use of BN has been shown to reduce the correlation of gradients in the final classification layer, with a significantly reduced dependence on the input compared to unnormalized networks~\citep{Nielsen2015,ioffe2015batch}.

To recall, for a layer of the network with \( d \)-dimensional input, \( \mathbf x = (x^{(1)}, \dots, x^{(d)}) \), each dimension of its input is then normalized (i.e. re-centered and re-scaled) separately,

\[
\hat{x}_i^{(k)} = \frac{x_i^{(k)} - \mu_B^{(k)}}{\sqrt{(\sigma_B^{(k)})^2 + \epsilon}}
\]

\noindent where \( k \in [1, d] \) and \( i \in [1, m] \); \( \mu_B^{(k)} \) and \( \sigma_B^{(k)} \) are the per-dimension mean and standard deviation, respectively. Then each component of $\hat{\mathbf x}$  is scaled linearly with two learnable parameters, the slope $\gamma^{(k)} $ and the intercept $\beta^{(k)} $.

Even though batch-normalisation and drop-out are commonly used architectural options for the construction of a modern neural network pipeline for successful learning, their influence on EEG data processing should not be overlooked. 

\subsection{Neurofeedback} 

Neurofeedback originated as an operant-conditioning form of biofeedback in which people receive contingent information about brain signals and learn to self-regulate those signals, a perspective emphasized in foundational editorials and reviews that define neurofeedback and its learning principles~\citep{cannon2015editorial,thibault2015neurofeedback}. Over recent decades the field expanded from classical EEG protocols to include real-time imaging modalities and debated the balance between specific neural learning and nonspecific factors (expectation, therapist interaction)~\citep{cannon2015editorial,thibault2015neurofeedback}.

It is commonly operationalized as contingent stimulus presentation based on an electrophysiological or imaging-derived criterion; canonical EEG approaches target frequency bands or slow cortical potentials, while newer implementations use hemodynamic or connectivity metrics~\citep{cannon2015editorial,thibault2015neurofeedback,omejc2019review}. Critics and proponents alike emphasize the need to specify mechanistic claims, distinguish operant-learning from placebo-like effects, and adopt standardized reporting and experimental controls to consolidate the theoretical foundation of neurofeedback~\citep{thibault2015neurofeedback,omejc2019review}.

This therapeutic technique enables individuals to self-regulate their brain activity by providing real-time feedback about their neural states. The algorithm~\ref{alg:neurofeedback} presented below uses a threshold-based classification system to translate EEG feature patterns into meaningful feedback signals, thereby facilitating the operant conditioning of specific brain states.

\begin{algorithm}[H]
\caption{Real-Time Neurofeedback Decision Algorithm}
\label{alg:neurofeedback}
\begin{algorithmic}[1]
\REQUIRE Trained classification model $\mathcal{M}$, fitted scaler $\mathcal{S}$, target brain state $y_{target} \in \{0, 1\}$, feedback threshold $\tau \in [0,1]$, EEG feature vector $\mathbf{x} \in \mathbb{R}^d$, ground truth label $y_{true}$
\ENSURE Feedback tuple $\{f, \omega, p_{target}, \hat{y}_{pred}, \delta\}$

\STATE \textbf{Stage 1: Feature Normalization}
\STATE $\mathbf{x}_{scaled} \leftarrow \mathcal{S}(\mathbf{x})$ \COMMENT{StandardScaler transformation}

\STATE \textbf{Stage 2: State Prediction}
\IF{$\mathcal{M}$ supports probability estimation}
    \STATE $\mathbf{p} \leftarrow \mathcal{M}.\text{predict\_proba}(\mathbf{x}_{scaled})$
    \STATE $p_{target} \leftarrow \mathbf{p}[y_{target}]$ 
\ELSE
    \STATE $\hat{y} \leftarrow \mathcal{M}.\text{predict}(\mathbf{x}_{scaled})$
    \STATE $\mathbf{p} \leftarrow [1-\hat{y}, \hat{y}]$
    \STATE $p_{target} \leftarrow \mathbf{p}[y_{target}]$
\ENDIF
\STATE $\hat{y}_{pred} \leftarrow \arg\max_{i} \mathbf{p}[i]$
\STATE $\delta \leftarrow \mathbb{I}[\hat{y}_{pred} = y_{true}]$ \COMMENT{Prediction correctness flag}

\STATE \textbf{Stage 3: Feedback Signal Generation}
\IF{$\hat{y}_{pred} = y_{target}$ \AND $p_{target} \geq \tau$}
    \STATE $f \leftarrow \text{POSITIVE}$ \hfill {\checkmark}
    \STATE $\omega \leftarrow p_{target}$ \COMMENT{Full reward strength $\in [0.7, 1.0]$}
\ELSIF{$\hat{y}_{pred} = y_{target}$ \AND $p_{target} < \tau$}
    \STATE $f \leftarrow \text{NEUTRAL}$ \hfill {\textbf{\textasciitilde}}
    \STATE $\omega \leftarrow 0.5 \times p_{target}$ \COMMENT{Partial reward $\in [0.0, 0.35)$}
\ELSE
    \STATE $f \leftarrow \text{NEGATIVE}$ \hfill {$\times$}
    \STATE $\omega \leftarrow 0$ \COMMENT{Zero reinforcement}
\ENDIF

\STATE \textbf{Stage 4: Session Logging}
\STATE $\mathcal{H} \leftarrow \mathcal{H} \cup \{(f, \omega, p_{target}, \hat{y}_{pred}, \delta, y_{true})\}$

\RETURN $(f, \omega, p_{target}, \hat{y}_{pred}, \delta)$

\end{algorithmic}
\end{algorithm}

The reinforcement intensity is determined by the piecewise function:

\begin{equation}
\omega(\mathbf{x}) = \mathcal{F}(\mathbf{x}; \tau, y_{target}) = 
\begin{cases} 
p_{target} & \text{if } \hat{y}_{pred} = y_{target} \land p_{target} \geq \tau \\[6pt]
\dfrac{p_{target}}{2} & \text{if } \hat{y}_{pred} = y_{target} \land p_{target} < \tau \\[10pt]
0 & \text{if } \hat{y}_{pred} \neq y_{target}
\end{cases}
\label{eq:feedback_strength}
\end{equation}

where $p_{target} = \mathbb{P}(\hat{Y} = y_{target} | \mathbf{x}_{scaled}, \mathcal{M})$ represents the model's confidence in the target class.

The table~\ref{tab:feedback_states} comprehensively defines the three feedback states implemented in the system. These states correspond to distinct phases of neural self-regulation during neurofeedback training sessions.

\begin{table*}[!ht]
\centering
\caption{Neurofeedback State Taxonomy and Clinical Interpretation}
\label{tab:feedback_states}
\begin{tabularx}{\textwidth}{XXXX}
\toprule
\textbf{Feedback State} & \textbf{Activation Condition} & \textbf{Neurophysiological Interpretation} & \textbf{Therapeutic Action} \\
\midrule

\textbf{POSITIVE}  & 
$\hat{y}_{pred} = y_{target}$ \par AND \par $p_{target} \geq \tau$ (0.7) & 
\textbf{Optimal self-regulation:} The brain has successfully entered and maintained the target neural pattern with a high degree of certainty. This represents the desired neuroplastic state, in which synaptic reinforcement should occur. &
\textbf{Reward delivery:} Provide immediate positive reinforcement through visual rewards (such as bright colours and animations), auditory feedback (such as pleasant tones) or gamification elements (such as points and progress bars). This strengthens the neural pathway via operant conditioning. \\ \hline 

\textbf{NEUTRAL}  & 
$\hat{y}_{pred} = y_{target}$ \par AND \par $p_{target} < \tau$ (0.7) & 
\textbf{Transitional approximation:} Neural activity is trending towards the target state, but it has not yet reached a magnitude or stability level that allows for full confidence. This indicates partial success in self-regulation. &
\textbf{Partial guidance:} It delivers attenuated feedback (dimmer visuals and softer sounds) to acknowledge the correct direction of change without celebrating it fully. Encourage the user to intensify the neural pattern towards the threshold. \\ \hline

\textbf{NEGATIVE}  & 
$\hat{y}_{pred} \neq y_{target}$ \par (regardless of confidence) & 
\textbf{Dysregulation or non-target state:} The brain produces patterns associated with the alternative class. This may indicate the presence of habitual neural modes that require inhibition. &
\textbf{Correction signal:} Withhold rewards or provide corrective cues. This teaches the brain that this state does not produce the desired outcome, encouraging it to explore alternative neural strategies. \\

\bottomrule
\end{tabularx}
\end{table*}

The effectiveness of a neurofeedback training session is quantified by the Successful Regulation Rate(SRR), which measures the proportion of trials where the user both achieved the target state and received positive reinforcement:

\begin{equation}
\text{SRR} = \frac{1}{T} \sum_{t=1}^{T} \mathbb{I}\left[ f_t = \text{POSITIVE} \land y_{true,t} = y_{target} \right] \times 100\
\label{eq:srr}
\end{equation}

where $T$ represents the total number of feedback trials in the session. Higher SRR indicates improved voluntary control over the targeted neural activity pattern.

The threshold $\tau = 0.7$~\citep{ban2024computational} ensures that only high-confidence detections trigger full rewards, preventing reinforcement of ambiguous neural states and maintaining the integrity of the operant conditioning process. All neurofeedback results are shown here~\footnote{\url{https://github.com/connect2robiul/round3-result-and-other/tree/main}}. 
\section{Data Collection and Processing} \label{sec:DataCollection}

Our analysis was based on visual perception of images with different intensity levels, \( I \in 0.1\text{--}1.0 \)~\citep{islam2022explainable,pisarchik2019coherent}. All experimental EEG data were recorded using the BE Plus LTM amplifier (EB Neuro S.p.a., Florence, Italy) across 31 channels at a sampling rate of 250~Hz. This study was approved by the Ethics Committee of Kant Baltic Federal University in accordance with the Declaration of Helsinki. The experiment is described in full in the reference~\citep{islam2021interpretable,pisarchik2019coherent,figshareData}. For two types of images, we used data from 10 healthy participants aged between 20 to 43. Two datasets were examined, one related to the painting of the \textbf{Mona Lisa} and the other to the observation of the ambiguous \textbf{Necker cube}. The repository~\citep{newdataset} contains all of the EEG data and presented figures presented herein along with the code used for the analysis.

In the research conducted by~\citep{islam2022explainable}, visual stimuli were employed to investigate brain responses. The experiment included the Necker cube and the Mona Lisa portrait by Leonardo da Vinci as stimuli. These images were selected to elicit varying intensities for the participants. Electroencephalogram (EEG) data was collected from 31 channels distributed across the brain hemispheres, as illustrated in a map detailing the electrode locations: The following electrode pairs were employed: \textit{O2-A2, O1-A1, P4-A2, P3-A1, C4-A2, C3-A1, F4-A2, F3-A1, Fp2-A2, Fp1-A1, T6-A2, T5-A1, T4-A2, T3-A1, F8-A2, F7-A1, Oz-A2, Pz-A1, Cz-A2, Fz-A1, Fpz-A2, FT7-A1, FC3-A1, Fcz-A1, FC4-A2, FT8-A2, TP7-A1, CP3-A1, Cpz-A1, CP4-A2}, and \textit{TP8-A2}. The EEG data from these channels was of crucial importance for the binary classification task performed in the study, as outlined in the 10--20 system according to the American Clinical Neurophysiology Society~\citep{sazgar2019overview} and~\citep{islam2021interpretable}'s previous work.

In the study by Islam~\citep{islam2022explainable}, real EEG channels were employed to investigate brain responses to visual stimuli. The experiment included a range of stimuli, including the Necker cube and Leonardo da Vinci's Mona Lisa portrait, selected to elicit varying intensities of response from participants. EEG data was collected from a comprehensive array of channels distributed across both hemispheres of the brain. Those ending in ``A2'' represented the right hemisphere, while those ending in ``A1'' represented the left hemisphere. These channels were strategically positioned to capture brain responses across various regions, playing a crucial role in the analysis.

\section{Models and Techniques}\label{sec:Models}

\subsection{EEG Data Processing Pipeline}

The EEG data processing pipeline, depicted in Figure~\ref{fig:eeg_pipeline}, outlines the sequential steps involved in analyzing electroencephalogram (EEG) data. The process begins with the acquisition of EEG data, comprising 15,000 data points recorded from 10 participants via 31 EEG channels in two datasets.Subsequently, the data undergoes filtration to extract relevant frequency ranges, including delta ($\delta$, 1-4 Hz), theta ($\theta$, 5-8 Hz), alpha ($\alpha$, 9-12 Hz), beta ($\beta$, 13-30 Hz), and gamma ($\gamma$, 31-45 Hz). 

Following filtration, the data is processed to create a binary dataset based on EEG channel readings, aimed at distinguishing left and right hemisphere conditions where label 0 encodes the left hemisphere and label 1 encodes the right hemisphere, respectively. The dataset is then randomly split into three subsets for training (70\

Overall, the EEG data processing pipeline facilitates the comprehensive analysis of EEG signals, enabling the identification of relevant features and insights into hemisphere specific neural activity patterns.

\begin{figure*}[!ht]
\centering
\begin{tikzpicture}[node distance=1.5cm]

\tikzstyle{process} = [rectangle, minimum width=3cm, minimum height=1.5cm, text centered, draw=black, text width=3cm]
\tikzstyle{ellipse_process} = [ellipse, minimum width=3cm, minimum height=1.5cm, text centered, draw=black, text width=3cm]

\node (data) [process] {

\tiny{
\textbf{EEG Data} \\ 
15,000 data points \\ 
31 EEG channels \\ 
10 participants  \\ 
2 datasets\\
10 intensity levels
}

};
\node (filtration) [process,right=of data] {

\tiny{

\textbf{Filtration~\citep{islam2021interpretable,islam2022explainable}} \\ 
5 frequency ranges: 
\begin{itemize}
\item delta $(\delta \in[1,4] \ \mathrm{Hz})$
\item theta $(\theta \in[5,8] \ \mathrm{Hz})$
\item alpha $(\alpha \in[9,12] \ \mathrm{Hz})$
\item beta $(\beta \in[13,30] \ \mathrm{Hz})$
\item gamma $(\gamma \in[31,45] \ \mathrm{Hz})$
\end{itemize}
}

};
\node (filtered_data) [ellipse_process, right=of filtration] {\tiny{Filtered EEG Data}};
\node (target) [ellipse_process, below=of filtered_data] {\tiny{Create a binary dataset based on EEG channel \\ Left and Right hemispheres)}};
\node (dataset) [process, left=of target] { \tiny{ \textbf{Dataset Split} \\ Train: 70\%, \\ Test: 15\%, \\ Validation: 15\%}};
\node (optimization) [process, left=of dataset] { \tiny{
\textbf{Take 1 of 8 methods of optimization~\citep{islam2022explainable}:}

\begin{itemize}
\item[] RMSProp \hspace{0.5cm} Adam
\item[] NAdam \hspace{0.6cm} AdaMax
\item[] SGD \hspace{0.82cm} Adagrad
\item[] Adadelta \hspace{0.48cm} FTRL
\end{itemize}
}
};
\node (dl_model) [ellipse_process, below=of optimization] {\tiny{
\begin{itemize}
    \item Apply DL model~(Table~\ref{paramtable}) 
    \item Apply CNN and shallow model (table:~\ref{tab:model_comparison}) 
\end{itemize}
}};
\node (features) [ellipse_process, right=of dl_model] {\tiny{DL model features' influence \\ and SHAP plot~\citep{ortigossa2024explainable}}};

\draw[-latex] (data) -- (filtration);
\draw[-latex] (filtration) -- (filtered_data);
\draw[-latex] (filtered_data) -- (target);
\draw[-latex] (target) -- (dataset);
\draw[-latex] (dataset) -- (optimization);
\draw[-latex] (optimization) -- (dl_model);
\draw[-latex] (dl_model) -- (features);

\end{tikzpicture}
\caption{Overview of EEG Data Processing Pipeline}
\label{fig:eeg_pipeline}
\end{figure*}

\subsection{Neural Network Architecture Overview}

The machine learning model described in Table~\ref{paramtable} represents a deep neural network architecture designed for a specific task (classifying signals recorded from left and right hemispheres), with a total of 9 dense layers~\citep{islam2021interpretable} and batch normalization layers interspersed throughout the network. Each dense layer is followed by a batch normalization layer, contributing to the stability and convergence of the model during training. 

The architecture begins with an input layer of 2500 neurons, representing the flattened input data. Subsequently, the network gradually reduces the dimensionality of the data through each dense layer, leading to a final output layer with a single neuron, performing the binary classification task at hand. 

Mathematically, the transformation at each dense layer can be represented as follows in batch-wise notation\citep{Nielsen2015,ioffe2015batch}:

\[
\mathbf Z^{[l]} = \mathbf W^{[l]} \cdot \mathbf A^{[l-1]} + \mathbf b^{[l]},
\]

\noindent where \( \mathbf Z^{[l]}\) represents the output of the $l$-th layer for the entire batch, \(\mathbf W^{[l]}\) denotes the matrix of weights, \(\mathbf A^{[l-1]}\) stands for the activations from the previous layer for the entire batch, and \(\mathbf b^{[l]}\) is the bias vector.

The parameters of the model, as detailed in Table~\ref{paramtable}, include the number of trainable parameters, which are adjusted during the training process to minimize the loss function and improve the model's performance. Additionally, there are non-trainable parameters, such as those associated with batch normalization layers, which are fixed during training~\citep{santurkar2018does}.

We also apply simple feed-forward (shallow or ``small'' model)~\citep{bebis1994feed} and Convolutional Neural Network (CNN) model~\citep{xu2019deep}. 

The extended table~\ref{tab:model_comparison} provides a comprehensive breakdown of the models' architectures, focusing on the following features:

\begin{itemize}
    \item \textbf{Architecture~\citep{cong2023review}:} Identifies the overall structure type of each model (deep feedforward, simplified feedforward, convolutional).

    \item \textbf{Layer Details:}
    \begin{itemize}
        \item For the Big Model, it lists all 29 layers with detailed descriptions of each linear layer, activation functions, and batch normalization.
        \item The Small Model summarizes its simpler architecture with three layers.
        \item The CNN~\citep{muratova2021comparison} Model outlines the three convolutional layers and includes descriptions of pooling layers and the final linear layers.
    \end{itemize}
    
    \item \textbf{Output Units:} Indicates that both the Big and Small models output a single value for binary classification, while the CNN Model outputs two values.

    \item \textbf{Activation Functions:} Notes the use of specified activation functions for the Big and Small models, with the CNN Model using Leaky ReLU~\citep{dubey2019comparative} after the final linear layer.

    \item \textbf{Learning Rate:} Displays the learning rates used for each model, which can impact convergence speed.

    \item \textbf{Batch Normalization~\citep{schilling2016effect,bjorck2018understanding}:} Indicates whether batch normalization is applied to each model, which helps in training stability.

    \item \textbf{Pooling:} Shows whether pooling layers are utilized, which is typical in CNN architectures to reduce dimensionality.

    \item \textbf{Use Case:} Provides insight into the suitability of each model for different types of data and tasks.
\end{itemize}

In summary, the architecture outlined in Table \ref{paramtable} showcases a deep neural network tailored for the specific task at hand, characterized by its layer structure, parameter count, and utilization of batch normalization for improved training stability and convergence.

\begin{table*}[!ht]
\centering
\tiny
\caption{``Big'' machine learning model with its parameters}
\label{paramtable}
\begin{tabular}{lll}
\hline
\textbf{Layer ( type ) }                                 & \textbf{Output Shape}    & \textbf{Param}    \\ \hline \hline
Dense ( Dense )                                 & 2500  & 37502500 \\
batch\_normalization ( Batch Normalization )    & 2500 & 10000    \\
dense\_1 ( Dense )                              & 1000  & 2501000  \\
batch\_normalization\_1  (BatchNormalization )  & 1000 & 4000     \\
dense\_2 ( Dense )                              & 500   & 500500   \\
batch\_normalization\_2 ( BatchNormalization )  & 500   & 2000     \\
dense 3 ( Dense )                               &  200   & 100200   \\
batch\_normalization 3 ( BatchNormalization )   & 200   & 800      \\
dense\_4 ( Dense )                              &  100   & 20100    \\
batch\_normalization\_4 ( BatchNormalization ) & 100   & 400      \\
dense\_5 ( Dense )                              &  50    & 5050     \\
batch\_normalization\_5 ( BatchNormalization )  &  50    & 200      \\
dense 6 ( Dense )                               &  25    & 1275     \\
batch\_normalization\_6 (BatchNormalization )   &  25    & 100      \\
dense\_7 ( Dense )                              &  15    & 390      \\
batch\_normalization\_7 ( BatchNormalization )  &  15    & 60       \\
dense\_8 ( Dense )                              &  10    & 160      \\
batch\_normalization\_8 ( BatchNormalization )  &  10    & 40       \\
dense 9 ( Dense )                               &  1    & 11       \\ \hline \hline
                                               
\multicolumn{3}{l}{Total parameters : 40648786}                                  \\
\multicolumn{3}{l}{Trainable parameters : 40639986}                              \\ 
\multicolumn{3}{l}{Non-trainable parameters : 8800}                             \\ \hline 
\end{tabular}
\end{table*}

\begin{table*}[h]
\centering
\tiny
\caption{Detailed Comparison of Big (same as~\citep{islam2022explainable}), Small, and CNN models executed by the Development Workstation (Sec:~\ref{sec:machine})}
\label{tab:model_comparison}
\begin{tabularx}{\textwidth}{XXXX}  
\hline
\textbf{Feature}             & \textbf{Big Model}                             & \textbf{Small Model}                            & \textbf{CNN Model}                             \\ \hline 
\textbf{Architecture}        & Deep feedforward                               & Feedforward                                    & Convolutional                                  \\ 

\textbf{Software Frameworks } & TensorFlow and PyTorch & PyTorch &  PyTorch\\

\hline
\textbf{Layer Details}       & Table~\ref{paramtable}    & \begin{tabular}[c]{@{}X@{}}1. Linear: \\ input: in\_features \\ output: out\_features \\ bias=False \\  \\ 2. Activation: specified \\  \\ 3. BatchNorm1d: out\_features \\  \\ 4. Linear: output: 10 \\  \\ 5. Activation: specified \\  \\ 6. Linear: output: 1 \\  \\ 7. Sigmoid\end{tabular}            & \begin{tabular}[c]{@{}X@{}}1. Conv1d: \\ in\_channels: 1 \\ out\_channels: 16 \\ kernel\_size: 3 \\ stride: 1 \\ padding: 1 \\  \\ 2. MaxPool1d: kernel\_size: 2 \\ stride: 2 \\  \\ 3. Conv1d: \\ in\_channels: 16 \\ out\_channels: 64 \\ kernel\_size: 3 \\ stride: 1 \\ padding: 1 \\  \\ 4. MaxPool1d: kernel\_size: 2 \\ stride: 2 \\  \\ 5. Conv1d: \\ in\_channels: 64 \\ out\_channels: 128 \\ kernel\_size: 3 \\ stride: 1 \\ padding: 1 \\  \\ 6. MaxPool1d: kernel\_size: 2 \\ stride: 2 \\  \\ 7. Linear: input: Config.in\_features \\ output: batch\_size \\  \\ 8. LeakyReLU \\  \\ 9. Linear: input: batch\_size \\ output: 2\end{tabular}                            \\ \hline
\textbf{Output Units}        & 1                                              & 1                                              & 2                                              \\ 
\textbf{Activation Functions} & Specified after each layer                    & Specified + Sigmoid                           & Leaky ReLU (after linear)                     \\ 
\textbf{Learning Rate}       & 0.01                          & 0.00001                          & 0.001                            \\ 
\textbf{Batch Normalization}  & Yes (after each linear layer)                  & Yes (after each linear layer)                  & No (applies to convolutional layers only)     \\ 
\textbf{Pooling}             & None                                           & None                                           & Max Pooling after each convolutional layer     \\ 
\textbf{Use Case}           & Suitable for complex tabular data               & Suitable for simple tabular data               & Suitable for sequential data (e.g., time series) \\ \hline
\end{tabularx}
\end{table*}

\subsection{Training}
The training stage involves selecting one of eight optimization methods, including Root Mean Square Propagation (RMSprop)~\citep{shi2021rmsprop}, Adaptive Moment Estimation (Adam)~\citep{kingma2014adam}, Nesterov-Accelerated Adaptive Moment Estimation (Nadam)~\citep{dozat2016incorporating}, Extension to the Adaptive Movement Estimation (AdaMax)~\citep{zeng2016adamax}, Stochastic Gradient Descent (SGD)~\citep{duda2019sgd}, Adaptive Gradient Algorithm (Adagrad)~\citep{lydia2019adagrad}, Extension of Adagrad (Adadelta)~\citep{zeiler2012adadelta}, and Follow the Regularized Leader (FTRL)~\citep{mcmahan2011follow}, to optimize the performance of a deep learning (DL) model. Once optimized, the DL model is applied to the processed EEG data to extract features and generate SHAP (SHapley Additive exPlanations)~\citep{lundberg2017unified,luo2024applications,ignatov2022shapley} plots, elucidating the influence of different features on the model's predictions.

\subsection{SHAP plots}

We also apply Shapley-value based interpretable machine learning approaches to possibly attribute important time moments of various signals that the models use to differentiate between hemispheres. 

\begin{figure}[!ht]
\centering
\includegraphics[width=0.3\textwidth]{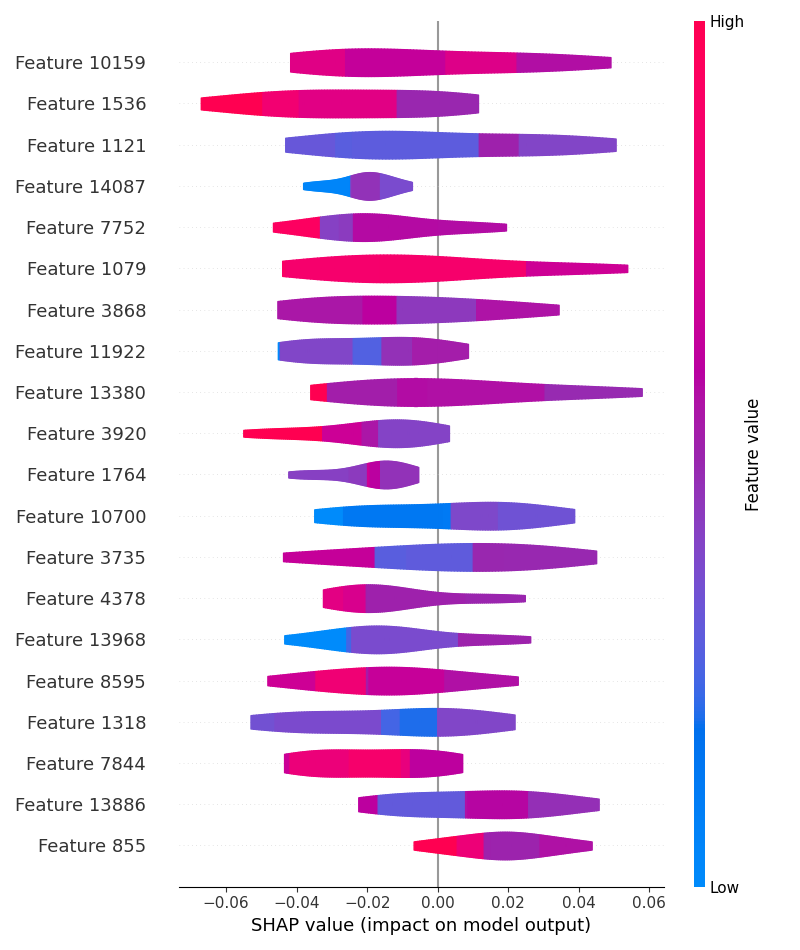}
\caption{SHAP plots for the $\beta$ frequency using the RMSProp optimizer, with the Necker cube dataset as input. Feature $F$ corresponds to the time $F/250$ s from the beginning of image presentation.}
\label{fig:shap_plot}
\end{figure}

Namely, we show SHAP plots~\citep{lundberg2017unified,burkov2013stochastic,lipovetsky2001analysis,vstrumbelj2014explaining,lundberg2020local} (Figure~\ref{fig:shap_plot}) for the $\beta$ frequency using the RMSProp optimizer, with the Necker cube dataset as input because it gives the highest performance. The purpose of this analysis was to gain insights into the feature importance and contribution of each input feature towards the model's output. The SHAP plots provide a visual representation of the SHapley Additive exPlanations (SHAP) values, which indicate the impact of each feature on the model output. The results of this analysis are presented and discussed in Section~\ref{sec:result}. In that section, we only focus on the topmost important features (positive or negative).

\subsection{Area under the Receiver Operating Characteristic Curve (AUC-ROC) Plots}

\begin{figure}[!ht]
    \centering
    \begin{minipage}{0.45\textwidth}
        \centering
        \includegraphics[width=0.8\textwidth]{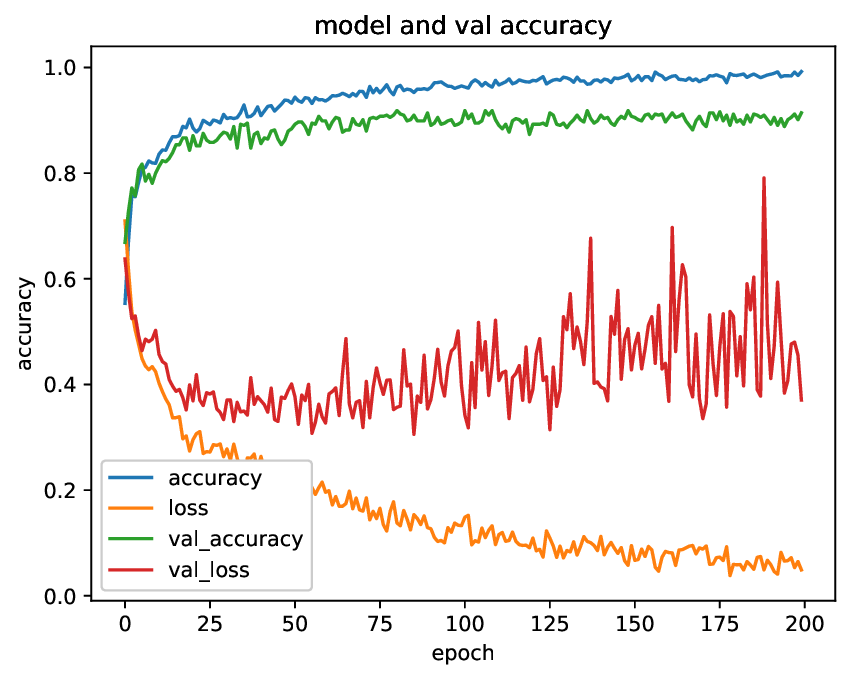} 
        \caption{Accuracy and loss plot for $\beta$ rhythm using the RMSProp optimizer, with the Necker cube dataset as input}
        \label{fig:plot_acc}
    \end{minipage}\hfill
    \begin{minipage}{0.45\textwidth}
        \centering
        \includegraphics[width=0.8\textwidth]{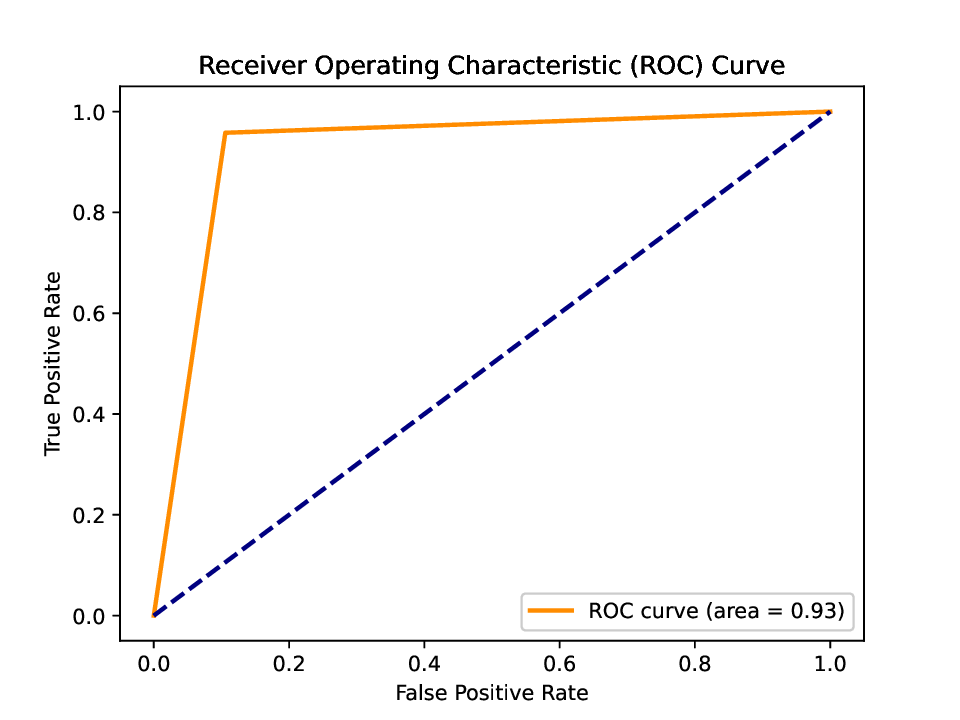} 
        \caption{ROC plots for $\beta$ rhythm using the RMSProp optimizer, with the Necker cube dataset as input}
        \label{fig:plot_roc}
    \end{minipage}
\end{figure}

We present the accuracy and loss, as well as the ROC plot (depicted in Figure~\ref{fig:plot_acc} and~\ref{fig:plot_roc}, respectively) for the $\beta$ frequency using the RMSProp optimizer. The Necker cube dataset was used as input, as it yielded the highest performance in our deep learning model, as demonstrated in Table~\ref{paramtable}. The results of this analysis are thoroughly discussed in Section~\ref{sec:result}, where we provide Precision, Recall, Specificity, F1 and ROC AUC scores obtained with each optimizer~\citep{islam2022explainable,ruder2016overview,hinton2012neural, duchi2011adaptive,dean2012large,pennington2014glove,zeiler2012adadelta,qian1999momentum,kingma2014adam,dozat2016incorporating,mcmahan2011follow,goldstein2014field,xiao2009dual}. Additionally, we include the Efficient Class column that indicates the bias of prediction quality towards either hemisphere.

\subsection{Classical Machine Learning vs. Deep Learning}

We systematically compared classical machine learning algorithms against deep neural network architectures to evaluate the trade-offs between model complexity, computational efficiency, and classification performance. This comparison addresses a fundamental question in EEG-based brain-computer interfaces: whether the additional complexity of deep learning justifies its computational cost compared to well-established classical methods.
\subsubsection{Algorithm Overview}

Table~\ref{tab:ml_comparison} summarizes the key characteristics, advantages, and computational complexities of the four primary algorithms employed in this study.

\begin{table*}[ht]
\centering
\tiny
\caption{Comparison of Classical Machine Learning and Deep Learning Algorithms for EEG Hemisphere Classification}
\label{tab:ml_comparison}
\begin{tabularx}{\textwidth}{XXXX}
\hline
\textbf{Algorithm} & \textbf{Category} & \textbf{Key Characteristics} & \textbf{Advantages}  \\
\hline
\textbf{Random Forest} & \multirow{3}{*}{Classical ML} & Ensemble of 100 decision trees with bootstrap aggregating; feature randomization at each split; parallel tree construction & Robust to overfitting; handles high-dimensional data; provides feature importance; minimal hyperparameter tuning  \\

\textbf{SVM (Linear)} &  & Linear hyperplane separation with maximum margin; soft-margin classification with regularization parameter $C$; probability estimates via Platt scaling & Fast training; interpretable decision boundary; effective in high dimensions; global optimum guaranteed  \\

\textbf{SVM (RBF)} & & Non-linear classification using Radial Basis Function kernel: $K(\mathbf{x}_i, \mathbf{x}_j) = \exp(-\gamma \|\mathbf{x}_i - \mathbf{x}_j\|^2)$ with $\gamma = 1/D$ & Captures complex non-linear patterns; robust to noise; works well with small samples; kernel trick avoids explicit high-dim mapping \\ 
\hline 
\textbf{Deep Neural Network} & Deep Learning & 10-layer fully-connected architecture (Table~\ref{paramtable}); RMSprop optimizer; 200 early stopping; 1000 epochs; Learning rate 0.001 & Automatic hierarchical feature learning; captures complex non-linear relationships; end-to-end optimization; no manual feature engineering  \\
\hline
\end{tabularx}

\end{table*}

\subsubsection{Random Forest}

Random Forest constructs an ensemble of decision trees through bootstrap aggregating (bagging), where each tree is trained on a random sample of the data with replacement. At each node split, only a random subset of $\sqrt{D} \approx 122$ features is considered, introducing decorrelation among trees. The final prediction aggregates individual tree votes through majority voting:

\begin{equation}
\hat{y} = \text{mode}\{h_1(\mathbf{x}), h_2(\mathbf{x}), \ldots, h_M(\mathbf{x})\}
\end{equation}

where $h_m(\mathbf{x})$ represents the prediction of the $m$-th tree. We implemented Random Forest with $M=100$ estimators using parallel processing across all available CPU cores

\subsubsection{Support Vector Machines}

Support Vector Machines (SVM)~\citep{mahmud2022air} identify the optimal hyperplane that maximizes the margin between classes. The primal optimization problem is:

\begin{equation}
\min_{\mathbf{w}, b, \boldsymbol{\xi}} \frac{1}{2}\|\mathbf{w}\|^2 + C\sum_{i=1}^{N}\xi_i
\end{equation}

subject to $y_i(\mathbf{w}^T\mathbf{x}_i + b) \geq 1 - \xi_i$ and $\xi_i \geq 0$, where $\mathbf{w}$ is the weight vector, $b$ is the bias term, $C$ is the regularization parameter controlling the trade-off between margin maximization and training error, and $\xi_i$ are slack variables permitting soft-margin classification.

\textbf{Linear Kernel:} Uses $K(\mathbf{x}_i, \mathbf{x}_j) = \mathbf{x}_i^T\mathbf{x}_j$, directly separating classes in the original feature space. The decision function is:
\begin{equation}
f(\mathbf{x}) = \text{sign}(\mathbf{w}^T\mathbf{x} + b) = \text{sign}\left(\sum_{i \in SV} \alpha_i y_i \mathbf{x}_i^T\mathbf{x} + b\right)
\end{equation}

where $SV$ denotes support vectors and $\alpha_i$ are Lagrange multipliers.

\textbf{RBF (Radial Basis Function) Kernel:} Projects data into infinite-dimensional Hilbert space using:
\begin{equation}
K(\mathbf{x}_i, \mathbf{x}_j) = \exp\left(-\gamma \|\mathbf{x}_i - \mathbf{x}_j\|^2\right), \quad \gamma = \frac{1}{D} = \frac{1}{15000}
\end{equation}

The decision function becomes:
\begin{equation}
f(\mathbf{x}) = \text{sign}\left(\sum_{i \in SV} \alpha_i y_i \exp(-\gamma \|\mathbf{x}_i - \mathbf{x}\|^2) + b\right)
\end{equation}

\subsection{Computing System}~\label{sec:machine} 

The configuration of the computing systems we used to perform ML experiments is as follows:

\begin{itemize}
\item \textbf{Analytics Server}
\begin{itemize}
\item RAM: 512 GB
\item CPU: Intel(R) Xeon(R) CPU E5-2698 v4 @ 2.20 GHz
\item OS: Ubuntu 18.04.5 LTS 64 bit
\item GPU: Tesla V100-SXM2-32GB
\end{itemize}
\item  \textbf{Development Workstation} 
\begin{itemize}
\item RAM: 64 GB
\item CPU: Intel(R) Core(TM) Intel Core i9-12900KS @ 8P x 3.4 GHz, 8E x 2.5 GHz
\item OS: Ubuntu 18.04.5 LTS 64 bit
\item GPU: NVIDIA GeForce RTX 4090 24GB
\end{itemize}

\item \textbf{Kaggle Notebook}
\begin{itemize}
    \item RAM: 32 GB
    \item CPU: Intel(R) Xeon(R) CPU @ 2.00 GHz
    \item OS: Portable Operating System Interface (POSIX)
    \item Accelerator: GPU T4 $\times$ 2
\end{itemize} 

\item  \textbf{Software} 
\begin{itemize}
\item

On the Analytics Server \& Kaggle Notebook, we use Keras and TensorFlow~\citep{joseph2021keras}, which are primarily maintained by the TensorFlow community team, consisting of contributors from various organizations and independent developers. Google plays a major role in the development and support of TensorFlow, while Keras, now tightly integrated with TensorFlow, is also supported by Google and the broader open-source community.

\item 
On the Development Workstation, we use the software framework PyTorch~\citep{paszke2019pytorch} (which stems from Torch~\citep{collobert2011torch7}) for conducting experiments. PyTorch is a scientific computing platform that offers extensive support for machine learning algorithms and is primarily maintained by Meta (formerly Facebook), with contributions from a broad community of researchers and developers.
\end{itemize}
\end{itemize}

In Section~\ref{sec:result}, we provide a comprehensive analysis of the computation time taken by each optimizer on our machine, as described previously.
\section{Experimental results}\label{sec:result}

\subsection{Efficient Class and SHAP Impact}
In subsequent analysis, we identify the most efficient class for each frequency range, denoted as $L$ (left hemispheres) and $R$ (right hemispheres), based on the classifier's performance. Furthermore, the SHAP (+/- Impact) column provides insights into the contribution of \textbf{top features} to the classifier's predictions: positive (+ve) values indicate a positive impact, and negative (-ve) values indicate a negative impact. These insights help in understanding the discriminative power of features in EEG data and their relevance in classifying hemisphere-specific conditions~\citep{lundberg2017unified,burkov2013stochastic,lipovetsky2001analysis,vstrumbelj2014explaining}.

\subsubsection{Performance Evaluation of Adaptive Gradient Algorithm (Adagrad) Optimizer}

The results in Table~\ref{adagrad_rec} present the evaluation of the Adagrad optimizer's performance in classifying left and right hemispheres using EEG data.

\begin{table*}[!ht]
\centering 
\tiny 
\caption{Result of Adaptive Gradient Algorithm (Adagrad) optimizer on the model executed by the Analytics Server (Sec:~\ref{sec:machine})}
\label{adagrad_rec}

\begin{tabular}{lllllllll} \hline
\textbf{Rhythm} & \textbf{Dataset} & \textbf{Precision} & \textbf{Recall} & \textbf{Specificity} & \textbf{F1-Score} &  \textbf{ROC AUC score} & \textbf{Efficient Class} & \textbf{SHAP +/- Impact} \\ \hline
\multirow{2}{*}{$\delta$} & Mona Lisa   & \multirow{2}{*}{0.78} & 0.76 & 0.76 & 0.77 & 0.77 & $R$ & \multirow{2}{*}{+ve} \\
                       & Necker cube &  & 0.82 & 0.82 & 0.8  & 0.79 & $L$   &  \\ \hline
\multirow{2}{*}{$\theta$} & Mona Lisa   & \multirow{2}{*}{0.8}  & 0.8  & 0.8 & 0.8  & 0.8 & $L$   &  \multirow{2}{*}{-ve} \\
                       & Necker cube &   & 0.75 & 0.75 & 0.78 & 0.78 & $R$ &  \\ \hline
\multirow{2}{*}{$\alpha$} & Mona Lisa   & 0.76 & 0.78 & 0.78 & 0.77 & 0.76 & $L$   & -ve \\
                       & Necker cube & 0.77 & 0.81 & 0.81 & 0.79 & 0.8 & $R$ & +ve \\ \hline
\multirow{2}{*}{$\beta$}  & Mona Lisa   & \multirow{2}{*}{0.9}  & \multirow{2}{*}{0.86} & \multirow{2}{*}{0.86} & \multirow{2}{*}{0.88} & \multirow{2}{*}{0.87} & \multirow{2}{*}{$L$ }  &  \multirow{2}{*}{+ve} \\
                       & Necker cube &   &  &  &  &  &   &  \\ \hline
\multirow{2}{*}{$\gamma$} & Mona Lisa   & 0.88 & 0.87 & 0.87 & 0.87 & 0.88 & $R$ &  \multirow{2}{*}{-ve} \\
                       & Necker cube & 0.87 & 0.91 & 0.91 & 0.89 & 0.89 & $L$   &  \\ \hline
                       
            \multicolumn{9}{l}{* Refer \textbf{left hemispheres} as $L$ and \textbf{right hemispheres} as $R$. Cell : \textbf{Efficient Class} } \\ 
            \multicolumn{9}{l}{* Refer +ve as positive contribution and -ve as negative contribution of SHAP plot for the top impact time in EEG data.  Cell : \textbf{SHAP +/- Impact} }
\end{tabular}

\end{table*}

\paragraph{Computational time for Adaptive Gradient Algorithm (Adagrad)}

Table \ref{Adagrad_time} presents the computational time required for different stages of Adagrad optimization process across various frequency ranges and datasets.

\begin{table*}[!ht]
\centering 
\tiny
\caption{Computational time of different Adaptive Gradient Algorithm (Adagrad) optimizers on the model executed by the Analytics Server (Sec:~\ref{sec:machine})}
\label{Adagrad_time}

\begin{tabular}{lllll}  
\hline 
 & & \multicolumn{3}{c}{\textbf{Time in Seconds}} \\ 
 \cline{3-5}
\multirow{-2}{*}{ \textbf{Rhythm}} & \multirow{-2}{*}{\textbf{Dataset}} & \textbf{Preprocessing} & \textbf{Model} & \textbf{SHAP} \\ \hline
                        & Mona Lisa   & 128.19 & 5589.60 & 1107.74  \\
\multirow{-2}{*}{$\delta$} & Necker cube & 123.07 & 5061.87 & 1113.36 \\  \hline
                        & Mona Lisa   & 130.09 & 5381.14 & 1107.76 \\ 
\multirow{-2}{*}{$\theta$} & Necker cube & 122.60 & 3481.00 & 1105.17 \\ \hline
                        & Mona Lisa   & 130.00 & 4410.46 & 1108.32 \\ 
\multirow{-2}{*}{$\alpha$} & Necker cube & 123.40 & 6548.23 & 1130.76 \\ \hline
                        & Mona Lisa   & 122.98 & 6954.03 & 1114.72 \\ 
\multirow{-2}{*}{$\beta$}  & Necker cube & 125.23 & 6604.58 & 1107.50 \\ \hline
                        & Mona Lisa   & 123.61 & 3900.14 & 1128.07 \\
\multirow{-2}{*}{$\gamma$} & Necker cube & 124.85 & 5092.52 & 1115.82 \\  \hline
\end{tabular}

\end{table*}

\paragraph{Models’ architectures for Adaptive Gradient Algorithm (Adagrad)} 

Table~\ref{tab:newmachineadagrad} presents the performance results of the Adaptive Gradient Algorithm (Adagrad) optimizer when applied to a variety of models. The table classifies the results according to the datasets. The training, validation, and testing accuracy of each model are presented along with the ROC AUC values, providing a comprehensive overview of their performance. Additionally, the training and inference times in seconds are documented, offering insights into the efficiency of each model. The results indicate variations in performance across different model configurations, suggesting that larger models generally achieve higher accuracy rates compared to their smaller counterparts. These findings highlight the importance of selecting appropriate model sizes and optimization techniques in the development of machine learning applications.

\begin{table*}[!ht]
\centering
\tiny
\caption{Results of Adaptive Gradient Algorithm (Adagrad)  optimizer on the model executed by the Development Workstation (Sec:~\ref{sec:machine})}
\label{tab:newmachineadagrad}
\resizebox{\textwidth}{!}{
\begin{tabularx}{\textwidth}{lllS[table-format=1.2]S[table-format=1.2]S[table-format=1.2]S[table-format=1.2]XXX}
\hline
\textbf{Rhythm} & \textbf{Dataset} & \textbf{Model} & \textbf{Train Accuracy} & \textbf{Val Accuracy} & \textbf{Test Accuracy} & \textbf{ROC AUC} & \textbf{Epoch} & \multicolumn{2}{c}{\textbf{Time in Seconds}} \\ \cline{9-10}
                &                   &                &                         &                       &                       &                  &               & \textbf{Train Time} & \textbf{Inference Time} \\
\hline

\multirow{6}{*}{$\delta$} & \multirow{3}{*}{Mona Lisa}   & Big   & 0.9589         & 0.8968       & 0.8677        & 0.9439  & 107   & 577.36     & 1.58           \\
                       &                              & CNN   & 0.9677         & 0.9          & 0.8677        & 0.8972  & 54    & 339.62     & 1.71           \\
                       &                              & Small & 0.521          & 0.5645       & 0.4935        & 0.5169  & 26    & 145.77     & 1.76           \\  \cline{2-10}
                       & \multirow{3}{*}{Necker cube} & Big   & 0.9548         & 0.871        & 0.871         & 0.9474  & 98    & 553.18     & 1.69           \\
                       &                              & CNN   & 0.9673         & 0.9          & 0.9097        & 0.941   & 70    & 457.14     & 1.85           \\
                       &                              & Small & 0.4794         & 0.5129       & 0.4613        & 0.4526  & 4     & 20.7       & 1.54           \\ \hline
\multirow{6}{*}{$\theta$} & \multirow{3}{*}{Mona Lisa}   & Big   & 0.9597         & 0.8903       & 0.8839        & 0.9653  & 49    & 271.23     & 1.64           \\
                       &                              & CNN   & 0.9665         & 0.8742       & 0.9065        & 0.9369  & 23    & 148.74     & 1.78           \\
                       &                              & Small & 0.6488         & 0.5806       & 0.5774        & 0.6107  & 78    & 443.65     & 1.78           \\  \cline{2-10}
                       & \multirow{3}{*}{Necker cube} & Big   & 0.954          & 0.8774       & 0.8581        & 0.9301  & 41    & 234.14     & 1.83           \\
                       &                              & CNN   & 0.9673         & 0.8903       & 0.8484        & 0.877   & 34    & 229.24     & 1.93           \\
                       &                              & Small & 0.4948         & 0.4903       & 0.5032        & 0.4932  & 18    & 95.13      & 1.6            \\ \hline
\multirow{6}{*}{$\alpha$} & \multirow{3}{*}{Mona Lisa}   & Big   & 0.9677         & 0.9065       & 0.871         & 0.9529  & 72    & 500.73     & 2.32           \\
                       &                              & CNN   & 0.9677         & 0.8903       & 0.8581        & 0.8857  & 21    & 186.31     & 3.37           \\
                       &                              & Small & 0.746          & 0.6258       & 0.5903        & 0.6872  & 115   & 889.53     & 2.78           \\  \cline{2-10}
                       & \multirow{3}{*}{Necker cube} & Big   & 0.9649         & 0.8806       & 0.8677        & 0.951   & 81    & 453.87     & 1.67           \\
                       &                              & CNN   & 0.9669         & 0.8645       & 0.8677        & 0.8965  & 26    & 167.62     & 1.79           \\
                       &                              & Small & 0.6508         & 0.5968       & 0.5452        & 0.592   & 65    & 371.32     & 1.81           \\ \hline
\multirow{6}{*}{$\beta$}  & \multirow{3}{*}{Mona Lisa}   & Big   & 0.9677         & 0.9194       & 0.9065        & 0.9872  & 84    & 683.37     & 2.9            \\
                       &                              & CNN   & 0.927          & 0.8968       & 0.9161        & 0.9473  & 34    & 210.41     & 1.67           \\
                       &                              & Small & 0.8105         & 0.6645       & 0.6742        & 0.7837  & 98    & 547.43     & 1.87           \\  \cline{2-10}
                       & \multirow{3}{*}{Necker cube} & Big   & 0.9677         & 0.9258       & 0.9161        & 0.9815  & 89    & 482.43     & 1.09           \\
                       &                              & CNN   & 0.927          & 0.9065       & 0.8774        & 0.907   & 32    & 207.87     & 1.89           \\
                       &                              & Small & 0.7089         & 0.6226       & 0.6194        & 0.7233  & 78    & 332.79     & 1.11           \\ \hline
\multirow{6}{*}{$\gamma$} & \multirow{3}{*}{Mona Lisa}   & Big   & 0.9677         & 0.9129       & 0.9226        & 0.9876  & 115   & 629.04     & 1.63           \\
                       &                              & CNN   & 0.9536         & 0.8581       & 0.8677        & 0.896   & 39    & 250.12     & 1.76           \\
                       &                              & Small & 0.6988         & 0.6774       & 0.6226        & 0.6927  & 73    & 413.03     & 1.78           \\  \cline{2-10}
                       & \multirow{3}{*}{Necker cube} & Big   & 0.9677         & 0.9          & 0.8935        & 0.9721  & 95    & 539.54     & 1.71           \\
                       &                              & CNN   & 0.9375         & 0.8452       & 0.8387        & 0.8669  & 40    & 270.41     & 1.95           \\
                       &                              & Small & 0.5661         & 0.4968       & 0.5355        & 0.5761  & 27    & 140.65     & 1.58  \\  \hline
\end{tabularx}

}
\end{table*}

\subsubsection{Performance Evaluation of Extension of Adagrad (Adadelta) optimizer}

The results in Table~\ref{adadelta_roc} present the evaluation of the Adadelta optimizer's performance in classifying left and right hemispheres using EEG data.

\begin{table*}[!ht]
\centering
\tiny
\caption{Result of  Extension of Adagrad (Adadelta) optimizer on the model executed by the Analytics Server (Sec:~\ref{sec:machine})}
\label{adadelta_roc}
\begin{tabular}{lllllllll}
\hline
\textbf{Rhythm} & \textbf{Dataset} & \textbf{Precision} & \textbf{Recall} & \textbf{Specificity} & \textbf{F1-Score} &  \textbf{ROC AUC score} & \textbf{Efficient Class} & \textbf{SHAP +/- Impact} \\ \hline
\multirow{2}{*}{$\delta$} & Mona Lisa   & 0.66 & 0.66 & 0.66 & 0.66 & 0.67 & \multirow{2}{*}{$R$} & -ve                    \\
                       & Necker cube & 0.69 & 0.67 & 0.68 & 0.68 & 0.69 &                    & +ve                  \\  \hline
\multirow{2}{*}{$\theta$} & Mona Lisa   & 0.7  & 0.74 & 0.74 & 0.72 & \multirow{2}{*}{0.7}  & \multirow{2}{*}{$L$} & +ve                  \\
                       & Necker cube & 0.72 & 0.76 & 0.77 & 0.74 &   &                    & -ve                  \\  \hline
\multirow{2}{*}{$\alpha$} & Mona Lisa   & \multirow{2}{*}{0.73} & \multirow{2}{*}{0.71} & 0.7  & \multirow{2}{*}{0.72} & \multirow{2}{*}{0.7}  & $R$                  & \multirow{2}{*}{+ve} \\
                       & Necker cube &  &  & 0.71 &  &   & $L$                  &                        \\  \hline
\multirow{2}{*}{$\beta$}  & Mona Lisa   & 0.76 & \multirow{2}{*}{0.74} & \multirow{2}{*}{0.74} & 0.75 & \multirow{2}{*}{0.75} & \multirow{2}{*}{$R$} & +ve                  \\
                       & Necker cube & 0.73 &  &  & 0.74 &  &                    & -ve                  \\  \hline
\multirow{2}{*}{$\gamma$} & Mona Lisa   & 0.86 & 0.8  & \multirow{2}{*}{0.8}  & 0.83 & 0.8  & $L$                  & \multirow{2}{*}{+ve} \\
                       & Necker cube & 0.81 & 0.81 &   & 0.81 & 0.81 & $R$                  &                       \\  \hline

                        \multicolumn{9}{l}{* Refer \textbf{left hemispheres} as $L$ and \textbf{right hemispheres} as $R$. Cell : \textbf{Efficient Class} } \\ 
            \multicolumn{9}{l}{* Refer +ve as positive contribution and -ve as negative contribution of SHAP plot for the top impact time in EEG data.  Cell : \textbf{SHAP +/- Impact} }
\end{tabular}
\end{table*}
\paragraph{Computational time for Extension of Adagrad (Adadelta) optimizer }

Table~\ref{Adadelta_time} presents the computational time required for different stages of the Adadelta optimization process across various frequency ranges and datasets.

\begin{table*}[!ht]
\centering
\tiny
\caption{Computational time of different Extension of Adagrad (Adadelta) optimizer on the model executed by the Analytics Server (Sec:~\ref{sec:machine})}
\label{Adadelta_time}

\begin{tabular}{lllll}
\hline 
 & & \multicolumn{3}{c}{\textbf{Time in Seconds}} \\ 
 \cline{3-5}
\textbf{Rhythm} & \textbf{Dataset} & \textbf{Preprocessing} & \textbf{Model} & \textbf{SHAP} \\ \hline
$\delta$ & Mona Lisa & 127.38 & 16431.09 & 1109.39 \\
 & Necker cube & 128.49 & 16424.54 & 1108.01 \\ \hline
$\theta$ & Mona Lisa & 123.46 & 16428.13 & 1163.93 \\
 & Necker cube & 129.69 & 16414.66 & 1110.15 \\ \hline
$\alpha$ & Mona Lisa & 123.22 & 16419.18 & 1112.31 \\
 & Necker cube & 125.28 & 16410.57 & 1108.76 \\ \hline
$\beta$ & Mona Lisa & 126.80 & 16421.35 & 1103.45 \\
 & Necker cube & 130.28 & 16428.34 & 1108.63 \\ \hline
$\gamma$ & Mona Lisa & 127.78 & 16410.78 & 1109.76 \\
 & Necker cube & 131.91 & 16417.42 & 1111.25 \\ \hline
\end{tabular}
\end{table*}

\paragraph{Models’ architectures for Extension of Adagrad (Adadelta) }

Table~\ref{tab:newmachineAdadelta} presents the performance results of the Adadelta optimizer across a variety of models. For each model, regardless of size--whether large, convolutional neural network (CNN)-based, or smaller in scale--key metrics such as training accuracy, validation accuracy, test accuracy, and receiver operating characteristic (ROC) area under the curve (AUC) values are recorded. The results indicate that larger models generally achieve higher accuracy and AUC values; however, this comes at the cost of increased time consumption for both training and inference. These findings highlight the trade-offs between model complexity, accuracy, and computational efficiency.

\begin{table*}[!ht]
\centering
\tiny
\caption{Result of Extension of Adagrad (Adadelta) optimizer on the model executed by Development Workstation  (Sec:~\ref{sec:machine})}
\label{tab:newmachineAdadelta}
\resizebox{\textwidth}{!}{

\begin{tabularx}{\textwidth}{lllS[table-format=1.2]S[table-format=1.2]S[table-format=1.2]S[table-format=1.2]XXX}

\hline
\textbf{Rhythm} & \textbf{Dataset} & \textbf{Model} & \textbf{Train Accuracy} & \textbf{Val Accuracy} & \textbf{Test Accuracy} & \textbf{ROC AUC} & \textbf{Epoch} & \multicolumn{2}{c}{\textbf{Time in Seconds}} \\ \cline{9-10}
                &                   &                &                         &                       &                       &                  &               & \textbf{Train Time} & \textbf{Inference Time} \\
\hline

\multirow{6}{*}{$\delta$} & \multirow{3}{*}{Mona Lisa}   & Big   & 0.942741936 & 0.903225807 & 0.851612903 & 0.941247109 & 68  & 319.69     & 2.11           \\
                       &                              & CNN   & 0.967741936 & 0.906451613 & 0.870967742 & 0.90081836  & 63  & 300.24     & 2.17           \\
                       &                              & Small & 0.520967742 & 0.564516129 & 0.493548387 & 0.516856431 & 26  & 121.16     & 2.09           \\ \cline{2-10}
                       & \multirow{3}{*}{Necker cube} & Big   & 0.942741936 & 0.851612903 & 0.85483871  & 0.949297278 & 65  & 321.89     & 2.18           \\
                       &                              & CNN   & 0.967741936 & 0.893548387 & 0.896774194 & 0.928037716 & 70  & 341.82     & 2.29           \\
                       &                              & Small & 0.479435484 & 0.512903226 & 0.461290323 & 0.452588507 & 4   & 19.61      & 2.15           \\
\hline
\multirow{6}{*}{$\theta$} & \multirow{3}{*}{Mona Lisa}   & Big   & 0.956854839 & 0.877419355 & 0.880645161 & 0.959971535 & 50  & 238.21     & 2.11           \\
                       &                              & CNN   & 0.966129032 & 0.880645161 & 0.906451613 & 0.936888454 & 23  & 112.3      & 2.31           \\
                       &                              & Small & 0.648790323 & 0.580645161 & 0.577419355 & 0.610745419 & 78  & 431.54     & 2.61           \\ \cline{2-10}
                       & \multirow{3}{*}{Necker cube} & Big   & 0.955645161 & 0.867741936 & 0.877419355 & 0.954767835 & 58  & 285.56     & 2.17           \\ 
                       &                              & CNN   & 0.967741936 & 0.893548387 & 0.848387097 & 0.877201566 & 48  & 238.29     & 2.27           \\
                       &                              & Small & 0.494758065 & 0.490322581 & 0.503225807 & 0.493150685 & 18  & 87.84      & 2.17           \\
\hline
\multirow{6}{*}{$\alpha$} & \multirow{3}{*}{Mona Lisa}   & Big   & 0.966532258 & 0.916129032 & 0.864516129 & 0.96931151  & 62  & 223.76     & 1.55           \\
                       &                              & CNN   & 0.967741936 & 0.887096774 & 0.851612903 & 0.879025085 & 21  & 99.47      & 2.11           \\
                       &                              & Small & 0.745967742 & 0.625806452 & 0.587096774 & 0.68715531  & 115 & 537.45     & 2.06           \\ \cline{2-10}
                       & \multirow{3}{*}{Necker cube} & Big   & 0.963709677 & 0.880645161 & 0.848387097 & 0.952499555 & 66  & 321.96     & 2.17           \\
                       &                              & CNN   & 0.965725807 & 0.867741936 & 0.864516129 & 0.893257428 & 26  & 128.14     & 2.23           \\
                       &                              & Small & 0.650806452 & 0.596774194 & 0.54516129  & 0.592020993 & 65  & 314.34     & 2.22           \\
\hline
\multirow{6}{*}{$\beta$}  & \multirow{3}{*}{Mona Lisa}   & Big   & 0.967741936 & 0.925806452 & 0.919354839 & 0.988347269 & 69  & 319.59     & 2.06           \\
                       &                              & CNN   & 0.967741936 & 0.9         & 0.903225807 & 0.933819605 & 94  & 440.83     & 2.11           \\
                       &                              & Small & 0.810483871 & 0.664516129 & 0.674193548 & 0.783757339 & 98  & 453.25     & 2.08           \\ \cline{2-10}
                       & \multirow{3}{*}{Necker cube} & Big   & 0.967741936 & 0.912903226 & 0.912903226 & 0.983010141 & 100 & 491.37     & 2.17           \\
                       &                              & CNN   & 0.935483871 & 0.896774194 & 0.870967742 & 0.900462551 & 22  & 106.78     & 2.24           \\
                       &                              & Small & 0.708870968 & 0.622580645 & 0.619354839 & 0.723314357 & 78  & 392.93     & 2.24           \\
\hline
\multirow{6}{*}{$\gamma$} & \multirow{3}{*}{Mona Lisa}   & Big   & 0.967741936 & 0.922580645 & 0.909677419 & 0.983321473 & 114 & 539.01     & 2.1            \\
                       &                              & CNN   & 0.965725807 & 0.864516129 & 0.858064516 & 0.886052304 & 52  & 246.07     & 2.15           \\
                       &                              & Small & 0.698790323 & 0.677419355 & 0.622580645 & 0.692670343 & 73  & 341.27     & 2.09           \\ \cline{2-10}
                       & \multirow{3}{*}{Necker cube} & Big   & 0.967741936 & 0.896774194 & 0.896774194 & 0.975449208 & 91  & 444.48     & 2.16           \\
                       &                              & CNN   & 0.936290323 & 0.835483871 & 0.838709677 & 0.866927593 & 40  & 195.8      & 2.24           \\
                       &                              & Small & 0.566129032 & 0.496774194 & 0.535483871 & 0.576143035 & 27  & 131.01     & 2.16           \\
\hline
\end{tabularx}
}
\end{table*}

\subsubsection{Performance Evaluation of Stochastic Gradient Descent (SGD) optimizer }

The results in Table~\ref{sgd_roc} present the evaluation of the SGD optimizer's performance in classifying left and right hemispheres using EEG data.

\begin{table*}[!ht]
\centering 
\tiny
\caption{Results of Stochastic Gradient Descent (SGD) optimizer on the model executed by the Analytics Server (Sec:~\ref{sec:machine})}
\label{sgd_roc}
\begin{tabular}{lllllllll}
\hline
\textbf{Rhythm} & \textbf{Dataset} & \textbf{Precision} & \textbf{Recall} & \textbf{Specificity} & \textbf{F1-Score} &  \textbf{ROC AUC score} & \textbf{Efficient Class} & \textbf{SHAP +/- Impact} \\ \hline
\multirow{2}{*}{$\delta$} & Mona Lisa   & 0.71 & 0.64 & 0.64 & 0.67 & \multirow{2}{*}{0.68} & \multirow{2}{*}{$L$} & -ve   \\
                       & Necker cube & 0.68 & 0.71 & 0.7  & 0.69 &                       &                    & +ve \\ \hline
\multirow{2}{*}{$\theta$} & Mona Lisa   & 0.71 & 0.68 & 0.68 & 0.7  & \multirow{2}{*}{0.7}  & \multirow{2}{*}{$L$} & +ve \\
                       & Necker cube & 0.72 & 0.7  & 0.7  & 0.71 &                       &                    & -ve   \\ \hline
\multirow{2}{*}{$\alpha$} & Mona Lisa   & 0.6  & 0.63 & 0.62 & 0.62 & 0.6                   & \multirow{2}{*}{$L$} & +ve \\
                       & Necker cube & 0.74 & 0.79 & 0.8  & 0.77 & 0.77                  &                    & -ve   \\ \hline
\multirow{2}{*}{$\beta$}  & Mona Lisa   & 0.86 & 0.86 & 0.6  & 0.86 & 0.87                  & \multirow{2}{*}{$R$} & -ve   \\
                       & Necker cube & 0.83 & 0.82 & 0.82 & 0.83 & 0.82                  &                    & +ve \\ \hline
\multirow{2}{*}{$\gamma$} & Mona Lisa   & 0.77 & 0.8  & 0.8  & 0.78 & 0.79                  & $R$                  & -ve   \\
                       & Necker cube & 0.8  & 0.81 & 0.81 & 0.81 & 0.81                  & $L$                  & +ve   \\ \hline

                        \multicolumn{9}{l}{* Refer \textbf{left hemispheres} as $L$ and \textbf{right hemispheres} as $R$. Cell : \textbf{Efficient Class} } \\ 
            \multicolumn{9}{l}{* Refer +ve as positive contribution and -ve as negative contribution of SHAP plot for the top impact time in EEG data.  Cell : \textbf{SHAP +/- Impact} }
\end{tabular}
\end{table*}

\paragraph{Computational time for Stochastic Gradient Descent (SGD))}

Table~\ref{sgd_time} presents the computational time required for different stages of the SGD optimization process across various frequency ranges and datasets.

\begin{table*}[!ht]
\centering 
\tiny
\caption{Computational Time of Different Stochastic Gradient Descent (SGD) Optimizers on the model executed by the Analytics Server (Sec:~\ref{sec:machine})}
\label{sgd_time}

\begin{tabular}{lllll}
\hline 
 & & \multicolumn{3}{c}{\textbf{Time in Seconds}} \\ 
 \cline{3-5}
\multirow{-2}{*}{ \textbf{Rhythm}} & \multirow{-2}{*}{\textbf{Dataset}} & \textbf{Preprocessing} & \textbf{Model} & \textbf{SHAP} \\ \hline
\multirow{2}{*}{$\delta$} & Mona Lisa   & 121.64              & 1482.55       & 1107.38     \\
                       & Necker cube & 126.16              & 2585.71       & 1105.97     \\ \hline
\multirow{2}{*}{$\theta$} & Mona Lisa   & 123.41               & 1135.88      & 1104.43     \\
                       & Necker cube & 130.36               & 2564.40      & 1100.95     \\ \hline
\multirow{2}{*}{$\alpha$} & Mona Lisa   & 126.12              & 1577.88      & 1111.73     \\
                       & Necker cube & 126.61               & 2203.55      & 1106.44     \\ \hline
\multirow{2}{*}{$\beta$}  & Mona Lisa   & 127.31              & 719.52       & 1109.04     \\ 
                       & Necker cube & 127.84              & 2852.70      & 1114.05     \\ \hline
\multirow{2}{*}{$\gamma$} & Mona Lisa   & 126.69              & 919.49      & 1115.47     \\
                       & Necker cube & 125.56               & 887.46      & 1129.27    \\ \hline
\end{tabular}
\end{table*}

\paragraph{Models’ architectures for  Stochastic Gradient Descent (SGD) }

Table~\ref{tab:newmachinSGD}  presents the results of the Stochastic Gradient Descent (SGD) optimizer applied to different models. The results demonstrate that more complex models (e.g., CNN) typically exhibit higher accuracy but necessitate longer training periods. Furthermore, smaller models demonstrate efficiency in terms of time but exhibit reduced accuracy, exemplifying the necessity for a balance between model complexity and performance.

\begin{table*}[!ht]
\centering
\tiny
\caption{Results of Stochastic Gradient Descent (SGD)  optimizer on the model executed by the Development Workstation (Sec:~\ref{sec:machine})}
\label{tab:newmachinSGD}
\resizebox{\textwidth}{!}{
\begin{tabularx}{\textwidth}{lllS[table-format=1.2]S[table-format=1.2]S[table-format=1.2]S[table-format=1.2]XXX}
\hline
\textbf{Rhythm} & \textbf{Dataset} & \textbf{Model} & \textbf{Train Accuracy} & \textbf{Val Accuracy} & \textbf{Test Accuracy} & \textbf{ROC AUC} & \textbf{Epoch} & \multicolumn{2}{c}{\textbf{Time in Seconds}} \\ \cline{9-10}
                &                   &                &                         &                       &                       &                  &               & \textbf{Train Time} & \textbf{Inference Time} \\
\hline

\multirow{6}{*}{$\delta$} & \multirow{3}{*}{Mona Lisa}   & Big   & 0.8306 & 0.7871 & 0.7871 & 0.8852 & 15  & 59.21  & 1.59 \\
                       &                              & CNN   & 0.9653 & 0.8839 & 0.8581 & 0.8868 & 120 & 722.36 & 1.74 \\
                       &                              & Small & 0.9411 & 0.8516 & 0.7839 & 0.8986 & 78  & 330.98 & 1.77 \\ \cline{2-10}
                       & \multirow{3}{*}{Necker cube} & Big   & 0.8262 & 0.7677 & 0.7613 & 0.8574 & 13  & 54.07  & 1.7  \\
                       &                              & CNN   & 0.7629 & 0.7452 & 0.7419 & 0.7665 & 21  & 134.77 & 1.9  \\
                       &                              & Small & 0.925  & 0.8065 & 0.7645 & 0.8611 & 78  & 296.08 & 1.59 \\ \hline
\multirow{6}{*}{$\theta$} & \multirow{3}{*}{Mona Lisa}   & Big   & 0.9359 & 0.829  & 0.8097 & 0.9227 & 39  & 160.09 & 1.66 \\
                       &                              & CNN   & 0.9641 & 0.871  & 0.9065 & 0.9369 & 27  & 164.52 & 1.79 \\
                       &                              & Small & 0.9532 & 0.7935 & 0.8387 & 0.9249 & 73  & 313.57 & 1.79 \\ \cline{2-10}
                       & \multirow{3}{*}{Necker cube} & Big   & 0.8903 & 0.8226 & 0.7742 & 0.8722 & 33  & 139.11 & 1.74 \\
                       &                              & CNN   & 0.9573 & 0.8677 & 0.8548 & 0.8837 & 26  & 162.04 & 2.33 \\
                       &                              & Small & 0.9617 & 0.8516 & 0.8419 & 0.931  & 126 & 494.18 & 1.67 \\ \hline
\multirow{6}{*}{$\alpha$} & \multirow{3}{*}{Mona Lisa}   & Big   & 0.9444 & 0.8581 & 0.8323 & 0.9187 & 32  & 204.3  & 3.09 \\
                       &                              & CNN   & 0.9472 & 0.8806 & 0.8613 & 0.8891 & 14  & 120.16 & 2.98 \\
                       &                              & Small & 0.9512 & 0.829  & 0.829  & 0.9206 & 73  & 472.27 & 3.34 \\ \cline{2-10}
                       & \multirow{3}{*}{Necker cube} & Big   & 0.8948 & 0.8    & 0.7871 & 0.904  & 24  & 99.81  & 1.72 \\
                       &                              & CNN   & 0.9629 & 0.8516 & 0.8613 & 0.8907 & 26  & 159.21 & 1.8  \\
                       &                              & Small & 0.9649 & 0.8645 & 0.8516 & 0.93   & 126 & 544.67 & 1.79 \\ \hline
\multirow{6}{*}{$\beta$}  & \multirow{3}{*}{Mona Lisa}   & Big   & 0.9266 & 0.8323 & 0.871  & 0.9525 & 11  & 73.24  & 2.94 \\
                       &                              & CNN   & 0.9645 & 0.8935 & 0.9065 & 0.9364 & 27  & 159.81 & 1.71 \\
                       &                              & Small & 0.9677 & 0.8677 & 0.871  & 0.9758 & 98  & 413.87 & 1.74 \\ \cline{2-10}
                       & \multirow{3}{*}{Necker cube} & Big   & 0.9278 & 0.8742 & 0.8516 & 0.9647 & 17  & 70.14  & 1.68 \\
                       &                              & CNN   & 0.848  & 0.8419 & 0.8516 & 0.8788 & 24  & 147.86 & 1.81 \\
                       &                              & Small & 0.9645 & 0.8645 & 0.8645 & 0.9573 & 78  & 289.29 & 1.55 \\ \hline
\multirow{6}{*}{$\gamma$} & \multirow{3}{*}{Mona Lisa}   & Big   & 0.946  & 0.8774 & 0.8806 & 0.9633 & 41  & 163.12 & 1.64 \\
                       &                              & CNN   & 0.9399 & 0.8452 & 0.8387 & 0.8659 & 34  & 204.54 & 1.76 \\
                       &                              & Small & 0.9629 & 0.9    & 0.8839 & 0.9673 & 73  & 311.76 & 1.76 \\ \cline{2-10}
                       & \multirow{3}{*}{Necker cube} & Big   & 0.9395 & 0.8613 & 0.8387 & 0.9255 & 15  & 62.88  & 1.74 \\
                       &                              & CNN   & 0.8593 & 0.7935 & 0.8355 & 0.8637 & 31  & 196.29 & 1.92 \\
                       &                              & Small & 0.9492 & 0.8516 & 0.8097 & 0.9171 & 47  & 181.28 & 1.61 \\ \hline

\end{tabularx}
}
\end{table*}

\subsubsection{Performance Evaluation Adaptive Moment Estimation (Adam) optimizer}

The results in Table \ref{Adam_roc} present the evaluation of the Adam optimizer's performance in classifying left and right hemispheres using EEG data.

\begin{table*}[!ht]
\centering 
\tiny
\caption{Result of Adaptive Moment Estimation (Adam) optimizer on the model executed by the Analytics Server (Sec:~\ref{sec:machine})}
\label{Adam_roc}
\begin{tabular}{lllllllll}
\hline
\textbf{Rhythm} & \textbf{Dataset} & \textbf{Precision} & \textbf{Recall} & \textbf{Specificity} & \textbf{F1-Score} &  \textbf{ROC AUC score} & \textbf{Efficient Class} & \textbf{SHAP +/- Impact} \\ \hline
\multirow{2}{*}{$\delta$} & Mona Lisa   & \multirow{2}{*}{0.73} & 0.78                 & 0.79                 & 0.75     & 0.74    & \multirow{2}{*}{$L$}   & -ve           \\
                       & Necker cube &                       & 0.74                 & 0.73                 & 0.74     & 0.72    &                      & +ve            \\ \hline
\multirow{2}{*}{$\theta$} & Mona Lisa   & 0.74                  & 0.75                 & 0.75                 & 0.75     & 0.75    & $R$                    & +ve           \\
                       & Necker cube & 0.76                  & 0.78                 & 0.79                 & 0.77     & 0.76    & $L$                    & -ve           \\ \hline
\multirow{2}{*}{$\alpha$} & Mona Lisa   & 0.76                  & 0.83                 & 0.82                 & 0.8      & 0.78    & \multirow{2}{*}{$L$}   & +ve           \\
                       & Necker cube & 0.69                  & 0.74                 & 0.74                 & 0.72     & 0.71    &                      & -ve           \\ \hline
\multirow{2}{*}{$\beta$}  & Mona Lisa   & 0.89                  & 0.86                 & 0.86                 & 0.88     & 0.88    & $R$                    & -ve          \\
                       & Necker cube & 0.91                  & 0.89                 & 0.9                  & 0.9      & 0.9     & $L$                    & +ve            \\ \hline
\multirow{2}{*}{$\gamma$} & Mona Lisa   & 0.88                  & \multirow{2}{*}{0.9} & \multirow{2}{*}{0.9} & 0.89     & 0.88    & \multirow{2}{*}{$L$}   & -ve           \\
                       & Necker cube & 0.91                  &                      &                      & 0.9      & 0.9     &                      & +ve           \\ \hline

                        \multicolumn{9}{l}{* Refer \textbf{left hemispheres} as $L$ and \textbf{right hemispheres} as $R$. Cell : \textbf{Efficient Class} } \\ 
            \multicolumn{9}{l}{* Refer +ve as positive contribution and -ve as negative contribution of SHAP plot for the top impact time in EEG data.  Cell : \textbf{SHAP +/- Impact} }
\end{tabular}
\end{table*}

\paragraph{Computational time for Adaptive Moment Estimation (Adam) }

Table \ref{Adam_time} presents the computational time required for different stages of the Adadelta optimization process across various frequency ranges and datasets.

\begin{table*}[!ht]
\centering 
\tiny
\caption{Computational time of  the Adaptive Moment Estimation (Adam) optimizer on the model executed by the Analytics Server (Sec:~\ref{sec:machine})}
\label{Adam_time}

\begin{tabular}{lllll}
\hline 
 & & \multicolumn{3}{c}{\textbf{Time in Seconds}} \\ 
 \cline{3-5}
\multirow{-2}{*}{ \textbf{Rhythm}} & \multirow{-2}{*}{\textbf{Dataset}} & \textbf{Preprocessing} & \textbf{Model} & \textbf{SHAP} \\ \hline
\multirow{2}{*}{$\delta$} & Mona Lisa   & 124.80              & 8601.44      & 1107.38     \\
                       & Necker cube & 127.68              & 11770.34      & 1105.97     \\ \hline 
\multirow{2}{*}{$\theta$} & Mona Lisa   & 131.09              & 11513.75      & 1104.43     \\
                       & Necker cube & 124.52              & 10216.00      & 1100.95     \\ \hline 
\multirow{2}{*}{$\alpha$} & Mona Lisa   & 124.14              & 10369.54       & 1111.73     \\
                       & Necker cube & 124.49              & 11673.26      & 1106.44     \\ \hline 
\multirow{2}{*}{$\beta$}  & Mona Lisa   & 126.28               & 11306.39        & 1109.04     \\
                       & Necker cube & 123.26              & 7148.75      & 1114.05     \\ \hline 
\multirow{2}{*}{$\gamma$} & Mona Lisa   & 127.93               & 9363.10      & 1115.47     \\
                       & Necker cube & 122.94               & 9242.57      & 1129.27     \\ \hline 
\end{tabular}
\end{table*}

\paragraph{Models’ architectures for Adaptive Moment Estimation (Adam) }

Table~\ref{tab:newmachineAdam} presents the results of the Adaptive Moment Estimation (Adam) optimizer on a variety of models. The results indicate that smaller models typically exhibit faster inference times, whereas larger or more complex models, such as CNNs, achieve higher accuracy but at the cost of longer training times. In general, the Adam optimizer displays consistent performance across datasets. The CNN model represents an optimal balance between time and accuracy, particularly at specific frequency bands such as $\beta$ and $\gamma$.

\begin{table*}[!ht]
\centering
\tiny
\caption{Results of Adaptive Moment Estimation (Adam) optimizer on the model executed by the Development Workstation (Sec:~\ref{sec:machine})}
\label{tab:newmachineAdam}
\resizebox{\textwidth}{!}{
\begin{tabularx}{\textwidth}{lllS[table-format=1.2]S[table-format=1.2]S[table-format=1.2]S[table-format=1.2]XXX}
\hline
\textbf{Rhythm} & \textbf{Dataset} & \textbf{Model} & \textbf{Train Accuracy} & \textbf{Val Accuracy} & \textbf{Test Accuracy} & \textbf{ROC AUC} & \textbf{Epoch} & \multicolumn{2}{c}{\textbf{Time in Seconds}} \\ \cline{9-10}
                &                   &                &                         &                       &                       &                  &               & \textbf{Train Time} & \textbf{Inference Time} \\
\hline

\multirow{6}{*}{$\delta$} & \multirow{3}{*}{Mona Lisa} & Big   & 0.8540 & 0.8032 & 0.7677 & 0.8261 & 31 & 153.88 & 1.62 \\
                          &                            & CNN   & 0.8569 & 0.7710 & 0.7645 & 0.7899 & 22 & 136.02 & 1.72 \\
                          &                            & Small & 0.9585 & 0.8968 & 0.8774 & 0.9790 & 52 & 265.66 & 1.75 \\ \cline{2-10}
                          & \multirow{3}{*}{Necker cube} & Big   & 0.9032 & 0.8290 & 0.8290 & 0.9291 & 40 & 204.17 & 1.71 \\
                          &                            & CNN   & 0.7677 & 0.7419 & 0.5903 & 0.6095 & 25 & 161.14 & 1.84 \\
                          &                            & Small & 0.9524 & 0.8806 & 0.8968 & 0.9794 & 78 & 362.05 & 1.59 \\ \hline

\multirow{6}{*}{$\theta$} & \multirow{3}{*}{Mona Lisa}  & Big   & 0.9181 & 0.7871 & 0.8097 & 0.8896 & 36 & 179.94 & 1.65 \\
                          &                            & CNN   & 0.8306 & 0.7581 & 0.7419 & 0.7660 & 32 & 201.11 & 1.77 \\
                          &                            & Small & 0.9528 & 0.8710 & 0.8935 & 0.9709 & 24 & 124.17 & 1.80 \\ \cline{2-10}
                          & \multirow{3}{*}{Necker cube} & Big   & 0.9190 & 0.8065 & 0.7581 & 0.8704 & 47 & 242.88 & 1.72 \\
                          &                            & CNN   & 0.8520 & 0.7387 & 0.7452 & 0.7705 & 48 & 308.56 & 1.87 \\
                          &                            & Small & 0.9605 & 0.8645 & 0.8742 & 0.9652 & 47 & 226.02 & 1.64 \\ \hline

\multirow{6}{*}{$\alpha$} & \multirow{3}{*}{Mona Lisa}  & Big   & 0.8383 & 0.7710 & 0.7742 & 0.8814 & 14 & 428.15 & 3.53 \\
                          &                            & CNN   & 0.9544 & 0.8097 & 0.8065 & 0.8329 & 15 & 130.49 & 2.95 \\
                          &                            & Small & 0.9657 & 0.8806 & 0.8774 & 0.9647 & 41 & 301.06 & 2.87 \\ \cline{2-10}
                          & \multirow{3}{*}{Necker cube} & Big   & 0.8427 & 0.7355 & 0.7258 & 0.8167 & 11 & 55.52  & 1.67 \\
                          &                            & CNN   & 0.9242 & 0.7903 & 0.7968 & 0.8238 & 14 & 88.73  & 1.80 \\
                          &                            & Small & 0.9617 & 0.8806 & 0.8613 & 0.9527 & 47 & 243.87 & 1.80 \\ \hline

\multirow{6}{*}{$\beta$}  & \multirow{3}{*}{Mona Lisa}  & Big   & 0.9508 & 0.8871 & 0.8935 & 0.9592 & 31 & 232.88 & 2.92 \\
                          &                            & CNN   & 0.9677 & 0.9000 & 0.9226 & 0.9535 & 25 & 151.36 & 1.67 \\
                          &                            & Small & 0.9573 & 0.8968 & 0.9161 & 0.9886 & 41 & 208.66 & 1.85 \\ \cline{2-10}
                          & \multirow{3}{*}{Necker cube} & Big   & 0.9423 & 0.8742 & 0.8419 & 0.9252 & 29 & 147.33 & 1.68 \\
                          &                            & CNN   & 0.9669 & 0.8710 & 0.8710 & 0.8999 & 22 & 140.03 & 1.83 \\
                          &                            & Small & 0.9524 & 0.8968 & 0.9129 & 0.9827 & 47 & 214.08 & 1.50 \\ \hline

\multirow{6}{*}{$\gamma$} & \multirow{3}{*}{Mona Lisa}  & Big   & 0.9452 & 0.9000 & 0.8839 & 0.9613 & 35 & 172.47 & 1.63 \\
                          &                            & CNN   & 0.8669 & 0.8258 & 0.8387 & 0.8660 & 3  & 19.08  & 1.76 \\
                          &                            & Small & 0.9496 & 0.9065 & 0.9065 & 0.9823 & 24 & 123.25 & 1.78 \\ \cline{2-10}
                          & \multirow{3}{*}{Necker cube} & Big   & 0.9355 & 0.8645 & 0.8452 & 0.8935 & 21 & 107.89 & 1.70 \\
                          &                            & CNN   & 0.8399 & 0.8323 & 0.8161 & 0.8421 & 19 & 124.64 & 1.93 \\
                          &                            & Small & 0.9496 & 0.8839 & 0.8935 & 0.9642 & 47 & 221.31 & 1.61 \\ \hline
\end{tabularx}
}
\end{table*}

\subsubsection{Performance Evaluation Root Mean Square Propagation (Rmsprop)) optimizer}
The results in Table \ref{RMS_roc} present the evaluation of the RMSPprop optimizer's performance in classifying left and right hemispheres using EEG data.

\begin{table*}[!ht]
\centering 
\tiny
\caption{Result of Root Mean Square Propagation (RMSProp) optimizer on the model executed by the Analytics Server (Sec:~\ref{sec:machine})}
\label{RMS_roc}
\begin{tabular}{lllllllll}
\hline
\textbf{Rhythm} & \textbf{Dataset} & \textbf{Precision} & \textbf{Recall} & \textbf{Specificity} & \textbf{F1-Score} &  \textbf{ROC AUC score} & \textbf{Efficient Class} & \textbf{SHAP +/- Impact} \\ \hline
\multirow{2}{*}{$\delta$} & Mona Lisa   & 0.71 & 0.75 & 0.75 & 0.73 & 0.72 & \multirow{2}{*}{$L$} & +ve \\
                       & Necker cube & 0.75 & 0.76 & 0.76 & 0.75 & 0.74 &                    & -ve \\ \hline
\multirow{2}{*}{$\theta$} & Mona Lisa & 0.74      & 0.79   & 0.78        & \multirow{2}{*}{0.76} & \multirow{2}{*}{0.76} & $L$                    &             \\
                       & Necker cube & 0.77 & 0.75 & 0.75 &      &      & $R$                  & \multirow{-2}{*}{+ve} \\ \hline
\multirow{2}{*}{$\alpha$} & Mona Lisa   & 0.76 & 0.78 & 0.79 & 0.77 & 0.76 & \multirow{2}{*}{$L$} & \multirow{2}{*}{-ve} \\
                       & Necker cube & 0.75 & 0.83 & 0.83 & 0.79 & 0.78 &                    &  \\ \hline
\multirow{2}{*}{$\beta$}  & Mona Lisa   & 0.86 & 0.92 & 0.91 & 0.89 & 0.88 & $L$                  & -ve \\
                       & Necker cube & 0.95 & 0.89 & 0.9  & 0.92 & 0.92 & $R$                  & +ve \\ \hline
\multirow{2}{*}{$\gamma$} & Mona Lisa & 0.92      & 0.88   & 0.88        & \multirow{2}{*}{0.9}  & \multirow{2}{*}{0.9}  & $R$                    & \multirow{2}{*}{+ve}            \\
                       & Necker cube & 0.88 & 0.93 & 0.92 &      &      & $L$                  &   \\ \hline

                        \multicolumn{9}{l}{* Refer \textbf{left hemispheres} as $L$ and \textbf{right hemispheres} as $R$. Cell : \textbf{Efficient Class} } \\ 
            \multicolumn{9}{l}{* Refer +ve as positive contribution and -ve as negative contribution of SHAP plot for the top impact time in EEG data.  Cell : \textbf{SHAP +/- Impact} }
\end{tabular}
\end{table*}

\paragraph{Computational time for Root Mean Square Propagation (RMSProp) }

Table \ref{RMS_time} presents the computational time required for different stages of the optimization process across various frequency ranges and datasets.

\begin{table*}[!ht]
\centering 
\tiny
\caption{Computational time of Root Mean Square Propagation (RMSProp) optimizer on the model executed by the Analytics Server (Sec:~\ref{sec:machine})}
\label{RMS_time}

\begin{tabular}{lllll}
\hline 
 & & \multicolumn{3}{c}{\textbf{Time in Seconds}} \\ 
 \cline{3-5}
\multirow{-2}{*}{ \textbf{Rhythm}} & \multirow{-2}{*}{\textbf{Dataset}} & \textbf{Preprocessing} & \textbf{Model} & \textbf{SHAP} \\ \hline
\multirow{2}{*}{$\delta$}     & Mona Lisa   & 123.40              & 9263.04      & 1108.28     \\
    & Necker cube & 125.43              & 9188.18       & 1106.54     \\ \hline
\multirow{2}{*}{$\theta$}     & Mona Lisa   & 131.35              & 9423.57      & 1109.09     \\
   & Necker cube & 127.16              & 9247.46      & 1141.31     \\ \hline
\multirow{2}{*}{$\alpha$}    & Mona Lisa   & 130.97              & 9219.48      & 1111.37     \\
    & Necker cube & 127.40              & 9156.97       & 1109.85     \\ \hline
\multirow{2}{*}{$\beta$}     & Mona Lisa   & 130.36               & 9179.06      & 1112.53     \\
     & Necker cube & 125.42              & 9155.31      & 1110.16     \\ \hline
\multirow{2}{*}{$\gamma$}     & Mona Lisa   & 125.57               & 9281.98      & 1105.36      \\
   & Necker cube & 126.58              & 9220.69       & 1108.14     \\ \hline
\end{tabular}
\end{table*}

\paragraph{Models’ architectures for Root Mean Square Propagation (RMSProp)  }

Table~\ref{tab:newmachinRMSprop} presents the results of the Root Mean Square Propagation (RMSProp) optimizer. The results demonstrate that larger models exhibit elevated training times but superior performance, particularly in terms of accuracy and ROC AUC, especially for the Small models at frequency bands such as $\alpha$ and $\beta$. Conversely, the CNN models demonstrate a balance between computational efficiency and accuracy, particularly when evaluated at $\beta$ and $\gamma$ frequencies. The RMSProp optimizer produces results that are competitive with other methods, with small models demonstrating consistent performance across both datasets.

\begin{table*}[!ht]
\centering
\tiny
\caption{Results of Root Mean Square Propagation (RMSProp)  optimizer on the model executed by the Development Workstation (Sec:~\ref{sec:machine})}
\label{tab:newmachinRMSprop}
\resizebox{\textwidth}{!}{
\begin{tabularx}{\textwidth}{lllS[table-format=1.2]S[table-format=1.2]S[table-format=1.2]S[table-format=1.2]XXX}
\hline
\textbf{Rhythm} & \textbf{Dataset} & \textbf{Model} & \textbf{Train Accuracy} & \textbf{Val Accuracy} & \textbf{Test Accuracy} & \textbf{ROC AUC} & \textbf{Epoch} & \multicolumn{2}{c}{\textbf{Time in Seconds}} \\ \cline{9-10}
                &                   &                &                         &                       &                       &                  &               & \textbf{Train Time} & \textbf{Inference Time} \\
\hline

\multirow{6}{*}{$\delta$} & \multirow{3}{*}{Mona Lisa}   & Big   & 0.9294 & 0.8613 & 0.7935 & 0.849  & 64 & 293.09 & 1.64 \\
                       &                              & CNN   & 0.5286 & 0.5581 & 0.5355 & 0.5503 & 5  & 31.23  & 1.72 \\
                       &                              & Small & 0.9581 & 0.9032 & 0.8839 & 0.979  & 52 & 250.69 & 1.76 \\ \cline{2-10}
                       & \multirow{3}{*}{Necker cube} & Big   & 0.9081 & 0.8452 & 0.8258 & 0.922  & 49 & 236.26 & 1.71 \\
                       &                              & CNN   & 0.7036 & 0.5968 & 0.5645 & 0.5838 & 11 & 69.86  & 1.85 \\
                       &                              & Small & 0.9528 & 0.871  & 0.9    & 0.9736 & 47 & 205.71 & 1.6  \\ \hline
\multirow{6}{*}{$\theta$} & \multirow{3}{*}{Mona Lisa}   & Big   & 0.8403 & 0.7677 & 0.771  & 0.8801 & 25 & 117.3  & 1.65 \\
                       &                              & CNN   & 0.9024 & 0.8065 & 0.8452 & 0.8738 & 27 & 168.38 & 1.78 \\
                       &                              & Small & 0.9641 & 0.8581 & 0.8871 & 0.9669 & 27 & 131.52 & 1.96 \\ \cline{2-10}
                       & \multirow{3}{*}{Necker cube} & Big   & 0.9093 & 0.8387 & 0.7806 & 0.8646 & 31 & 150.66 & 1.74 \\
                       &                              & CNN   & 0.8673 & 0.7548 & 0.7774 & 0.8036 & 21 & 133.7  & 1.86 \\
                       &                              & Small & 0.9609 & 0.8839 & 0.8742 & 0.9692 & 78 & 352.48 & 1.65 \\ \hline
\multirow{6}{*}{$\alpha$} & \multirow{3}{*}{Mona Lisa}   & Big   & 0.9375 & 0.8419 & 0.8161 & 0.8981 & 35 & 245.81 & 2.82 \\
                       &                              & CNN   & 0.9431 & 0.8065 & 0.8065 & 0.8321 & 26 & 226.36 & 2.97 \\
                       &                              & Small & 0.9633 & 0.8839 & 0.871  & 0.966  & 27 & 191.93 & 2.88 \\ \cline{2-10}
                       & \multirow{3}{*}{Necker cube} & Big   & 0.8343 & 0.7419 & 0.7226 & 0.7821 & 15 & 71.12  & 1.68 \\
                       &                              & CNN   & 0.9218 & 0.8129 & 0.8032 & 0.83   & 39 & 243.95 & 1.81 \\
                       &                              & Small & 0.9629 & 0.8774 & 0.8742 & 0.9568 & 47 & 231.04 & 1.82 \\ \hline
\multirow{6}{*}{$\beta$}  & \multirow{3}{*}{Mona Lisa}   & Big   & 0.8992 & 0.8484 & 0.8613 & 0.9314 & 9  & 65.33  & 2.91 \\
                       &                              & CNN   & 0.9673 & 0.8903 & 0.8903 & 0.9203 & 30 & 179.53 & 1.68 \\
                       &                              & Small & 0.9528 & 0.9032 & 0.9129 & 0.9915 & 27 & 129.79 & 1.76 \\ \cline{2-10}
                       & \multirow{3}{*}{Necker cube} & Big   & 0.9383 & 0.8742 & 0.8645 & 0.9562 & 25 & 119.14 & 1.68 \\
                       &                              & CNN   & 0.9657 & 0.8839 & 0.8677 & 0.8969 & 21 & 133.48 & 1.84 \\
                       &                              & Small & 0.9569 & 0.8968 & 0.9129 & 0.9813 & 47 & 202.58 & 1.52 \\ \hline
\multirow{6}{*}{$\gamma$} & \multirow{3}{*}{Mona Lisa}   & Big   & 0.9173 & 0.8935 & 0.8581 & 0.9349 & 13 & 60.1   & 1.62 \\
                       &                              & CNN   & 0.9125 & 0.8548 & 0.8452 & 0.8725 & 17 & 105.66 & 1.75 \\
                       &                              & Small & 0.9524 & 0.9194 & 0.9    & 0.983  & 31 & 150.54 & 1.78 \\ \cline{2-10}
                       & \multirow{3}{*}{Necker cube} & Big   & 0.9306 & 0.8613 & 0.8484 & 0.9257 & 39 & 187.67 & 1.72 \\
                       &                              & CNN   & 0.9363 & 0.8387 & 0.8323 & 0.8595 & 18 & 117.16 & 1.92 \\
                       &                              & Small & 0.948  & 0.8806 & 0.8871 & 0.9605 & 47 & 208.72 & 1.6 \\ \hline

\end{tabularx}
}
\end{table*}

\subsubsection{Performance Evaluation Nesterov-Accelerated Adaptive Moment Estimation (NAdam) optimizer }

The results in Table~\ref{nadam_roc} present the evaluation of the NAdam optimizer's performance in classifying left and right hemispheres using EEG data.

\begin{table*}[!ht]
\centering 
\tiny
\caption{Result of Nesterov-Accelerated Adaptive Moment Estimation (NAdam) optimizer on the model executed by the Analytics Server (Sec:~\ref{sec:machine})}
\label{nadam_roc}
\begin{tabular}{lllllll|l|l}
\hline
\textbf{Rhythm} & \textbf{Dataset} & \textbf{Precision} & \textbf{Recall} & \textbf{Specificity} & \textbf{F1-Score} &  \textbf{ROC AUC score} & \textbf{Efficient Class} & \textbf{SHAP +/- Impact} \\ \hline
\multirow{2}{*}{$\delta$} & Mona Lisa   & 0.77 & 0.74 & 0.74 & 0.76 & 0.74 & $L$ & -ve \\
                       & Necker cube & 0.79 & 0.7  & 0.7  & 0.75 & 0.76 & $R$ & +ve \\ \hline
\multirow{2}{*}{$\theta$} & Mona Lisa & 0.8       & 0.8    & 0.8         & \multirow{2}{*}{0.8}  & \multirow{2}{*}{0.78} & \multirow{8}{*}{$L$}   & \multirow{2}{*}{+ve} \\
                       & Necker cube & 0.81 & 0.79 & 0.79 &      &      &   &     \\   \cline{1-7} \cline{9-9}
\multirow{2}{*}{$\alpha$} & Mona Lisa   & 0.72 & 0.81 & 0.8  & 0.76 & 0.76 &   & -ve \\  
                       & Necker cube & 0.75 & 0.83 & 0.83 & 0.79 & 0.77 &   & +ve \\ \cline{1-7} \cline{9-9}
\multirow{2}{*}{$\beta$}  & Mona Lisa & 0.86      & 0.88   & 0.88        & \multirow{2}{*}{0.87} & 0.86                  &                      & \multirow{2}{*}{-ve} \\
                       & Necker cube & 0.83 & 0.92 & 0.92 &      & 0.87 &   &     \\ \cline{1-7} \cline{9-9}
\multirow{2}{*}{$\gamma$} & Mona Lisa & 0.84      & 0.92   & 0.92        & \multirow{2}{*}{0.88} & 0.87                  &                      & +ve                 \\
                       & Necker cube & 0.89 & 0.88 & 0.88 &      & 0.88 &   & -ve   \\ \hline

                        \multicolumn{9}{l}{* Refer \textbf{left hemispheres} as $L$ and \textbf{right hemispheres} as $R$. Cell : \textbf{Efficient Class} } \\ 
            \multicolumn{9}{l}{* Refer +ve as positive contribution and -ve as negative contribution of SHAP plot for the top impact time in EEG data.  Cell : \textbf{SHAP +/- Impact} }
\end{tabular}
\end{table*}

\paragraph{Computational time for Nesterov-Accelerated Adaptive Moment Estimation (NAdam) }

Table~\ref{nadam_time} presents the computational time required for different stages of NAdam optimization process across various frequency ranges and datasets.

\begin{table*}[!ht]
\centering 
\tiny
\caption{Computational time of Nesterov-Accelerated Adaptive Moment Estimation (Nadam)  optimizer on the model executed by the Analytics Server (Sec:~\ref{sec:machine})}
\label{nadam_time}

\begin{tabular}{lllll}
\hline 
 & & \multicolumn{3}{c}{\textbf{Time in Seconds}} \\ 
 \cline{3-5}
\multirow{-2}{*}{ \textbf{Rhythm}} & \multirow{-2}{*}{\textbf{Dataset}} & \textbf{Preprocessing} & \textbf{Model} & \textbf{SHAP} \\ \hline
\multirow{2}{*}{$\delta$}    & Mona Lisa   & 125.66              & 15694.61      & 1108.67     \\
     & Necker cube & 125.55              & 15798.23      & 1105.86     \\ \hline
\multirow{2}{*}{$\theta$}    & Mona Lisa   & 124.19              & 15045.13      & 1111.13     \\
    & Necker cube & 125.25              & 15807.94      & 1107.85     \\ \hline
\multirow{2}{*}{$\alpha$}     & Mona Lisa   & 126.17              & 15534.80      & 1167.26     \\
    & Necker cube & 124.72              & 15421.97      & 1112.96     \\ \hline
\multirow{2}{*}{$\beta$}     & Mona Lisa   & 125.45              & 15790.75       & 1103.70     \\
      & Necker cube & 127.41              & 9284.89      & 1163.93     \\ \hline
\multirow{2}{*}{$\gamma$}    & Mona Lisa   & 122.15              & 14671.79      & 1109.76       \\
     & Necker cube & 121.15              & 9357.95      & 1105.81      \\ \hline
\end{tabular}
\end{table*}

\paragraph{Models’ architectures for Nesterov-Accelerated Adaptive Moment Estimation (NAdam)} 

Table~\ref{tab:newmachinenadam} presents the results of the Nesterov-Accelerated Adaptive Moment Estimation (NAdam) optimizer. The results demonstrate that models of a smaller scale tend to exhibit superior performance, attaining high levels of accuracy and ROC AUC scores while requiring minimal inference time. To illustrate, the small model for the Mona Lisa dataset at frequency $\beta$ achieves a test accuracy of 0.9194 with an ROC AUC of 0.9887 in 27 epochs. Conversely, CNN models typically exhibit elevated training and inference times, particularly within the $\delta$ and $\theta$ frequency ranges. Nevertheless, they continue to demonstrate competitive performance.

The 9-layer models also demonstrate robust performance across both datasets, exhibiting a balance between accuracy and computational efficiency. For instance, the Mona Lisa dataset at $\gamma$ achieved a test accuracy of 0.871 with an ROC AUC of 0.9455 in 16 epochs.

In conclusion, the NAdam optimizer produces effective results, particularly with small models, offering high accuracy and performance while maintaining reasonable computational costs.

\begin{table*}[!ht]
\centering
\tiny
\caption{Results of Nesterov-Accelerated Adaptive Moment Estimation (NAdam) optimizer on the model executed by the Development Workstation (Sec:~\ref{sec:machine})}
\label{tab:newmachinenadam}
\resizebox{\textwidth}{!}{

\begin{tabularx}{\textwidth}{lllS[table-format=1.2]S[table-format=1.2]S[table-format=1.2]S[table-format=1.2]XXX}
\hline
\textbf{Rhythm} & \textbf{Dataset} & \textbf{Model} & \textbf{Train Accuracy} & \textbf{Val Accuracy} & \textbf{Test Accuracy} & \textbf{ROC AUC} & \textbf{Epoch} & \multicolumn{2}{c}{\textbf{Time in Seconds}} \\ \cline{9-10}
                &                   &                &                         &                       &                       &                  &               & \textbf{Train Time} & \textbf{Inference Time} \\
\hline

\multirow{6}{*}{$\delta$} & \multirow{3}{*}{Mona Lisa}   & Big   & 0.8762         & 0.8129       & 0.7742        & 0.8837  & 35    & 180.6      & 1.59           \\
                       &                              & CNN   & 0.5698         & 0.5548       & 0.5581        & 0.5736  & 7     & 43.85      & 2.04           \\
                       &                              & Small & 0.9581         & 0.9032       & 0.8935        & 0.9789  & 78    & 416.78     & 1.79           \\ \cline{2-10}
                       & \multirow{3}{*}{Necker cube} & Big   & 0.8339         & 0.8065       & 0.7581        & 0.8613  & 27    & 145.02     & 1.69           \\
                       &                              & CNN   & 0.777          & 0.7129       & 0.6806        & 0.7036  & 76    & 490.47     & 1.82           \\
                       &                              & Small & 0.9552         & 0.8806       & 0.8968        & 0.9741  & 58    & 284.83     & 1.58           \\ \hline
\multirow{6}{*}{$\theta$} & \multirow{3}{*}{Mona Lisa}   & Big   & 0.8593         & 0.7871       & 0.7742        & 0.8897  & 27    & 142.62     & 1.67           \\
                       &                              & CNN   & 0.8762         & 0.7806       & 0.7839        & 0.8096  & 56    & 356.11     & 1.76           \\
                       &                              & Small & 0.9625         & 0.8484       & 0.8839        & 0.9662  & 41    & 222        & 1.8            \\ \cline{2-10}
                       & \multirow{3}{*}{Necker cube} & Big   & 0.9073         & 0.8194       & 0.7839        & 0.8746  & 45    & 243.61     & 1.73           \\
                       &                              & CNN   & 0.8891         & 0.771        & 0.7387        & 0.7638  & 95    & 616.25     & 1.85           \\
                       &                              & Small & 0.9585         & 0.8839       & 0.8774        & 0.9676  & 47    & 235.82     & 1.64           \\ \hline
\multirow{6}{*}{$\alpha$} & \multirow{3}{*}{Mona Lisa}   & Big   & 0.8            & 0.7323       & 0.7419        & 0.8395  & 8     & 60.43      & 2.76           \\
                       &                              & CNN   & 0.825          & 0.771        & 0.7677        & 0.7918  & 13    & 115.55     & 2.92           \\
                       &                              & Small & 0.9657         & 0.8903       & 0.8839        & 0.9644  & 52    & 391.43     & 2.88           \\ \cline{2-10}
                       & \multirow{3}{*}{Necker cube} & Big   & 0.9274         & 0.8419       & 0.8           & 0.8791  & 43    & 228.23     & 1.7            \\
                       &                              & CNN   & 0.9125         & 0.8          & 0.7903        & 0.8161  & 33    & 210.12     & 1.79           \\
                       &                              & Small & 0.9625         & 0.8774       & 0.8677        & 0.9538  & 78    & 423.06     & 1.8            \\ \hline
\multirow{6}{*}{$\beta$}  & \multirow{3}{*}{Mona Lisa}   & Big   & 0.9387         & 0.8774       & 0.8806        & 0.9511  & 33    & 259.72     & 2.9            \\
                       &                              & CNN   & 0.9673         & 0.871        & 0.8806        & 0.9102  & 5     & 31.18      & 1.67           \\
                       &                              & Small & 0.9548         & 0.9129       & 0.9194        & 0.9887  & 27    & 144.07     & 1.75           \\ \cline{2-10}
                       & \multirow{3}{*}{Necker cube} & Big   & 0.8956         & 0.8323       & 0.8323        & 0.9048  & 18    & 96.57      & 1.75           \\
                       &                              & CNN   & 0.9665         & 0.9161       & 0.8903        & 0.9198  & 14    & 90.19      & 1.81           \\
                       &                              & Small & 0.9548         & 0.9032       & 0.9129        & 0.983   & 47    & 227.49     & 1.52           \\ \hline
\multirow{6}{*}{$\gamma$} & \multirow{3}{*}{Mona Lisa}   & Big   & 0.9294         & 0.8839       & 0.871         & 0.9455  & 16    & 82.76      & 1.64           \\
                       &                              & CNN   & 0.9081         & 0.8258       & 0.8323        & 0.8588  & 15    & 96.02      & 1.77           \\
                       &                              & Small & 0.9528         & 0.9065       & 0.9           & 0.9864  & 31    & 166.28     & 1.78           \\  \cline{2-10}
                       & \multirow{3}{*}{Necker cube} & Big   & 0.9093         & 0.8742       & 0.8129        & 0.9209  & 15    & 80.86      & 1.71           \\
                       &                              & CNN   & 0.9435         & 0.8129       & 0.8065        & 0.8334  & 6     & 39.78      & 1.92           \\
                       &                              & Small & 0.952          & 0.8806       & 0.8839        & 0.9644  & 47    & 233.21     & 1.6           \\ \hline

\end{tabularx}

}
\end{table*}

\subsubsection{Performance Evaluation of Extension to the Adaptive Movement Estimation (AdaMax) optimizer}

The results in Table \ref{adamax_roc} present the evaluation of the AdaMax optimizer's performance in classifying left and right hemispheres using EEG data.
\begin{table*}[!ht]
\centering 
\tiny
\caption{Result of Extension to the Adaptive Movement Estimation (AdaMax) optimizer on the model executed by the Analytics Server (Sec:~\ref{sec:machine})}
\label{adamax_roc}
\begin{tabular}{lllllllll}
\hline
\textbf{Rhythm} & \textbf{Dataset} & \textbf{Precision} & \textbf{Recall} & \textbf{Specificity} & \textbf{F1-Score} &  \textbf{ROC AUC score} & \textbf{Efficient Class} & \textbf{SHAP +/- Impact} \\ \hline
\multirow{2}{*}{$\delta$} & Mona Lisa   & 0.78      & 0.83                  & 0.83                  & 0.8      & 0.79    & \multirow{2}{*}{$L$}   & \multirow{2}{*}{-ve} \\
                       & Necker cube & 0.81      & 0.82                  & 0.82                  & 0.81     & 0.8     &                      &                        \\ \hline
\multirow{2}{*}{$\theta$} & Mona Lisa   & 0.78      & 0.83                  & 0.83                  & 0.8      & 0.8     & $L$                    & +ve                  \\
                       & Necker cube & 0.8       & 0.84                  & 0.84                  & 0.82     & 0.83    & $R$                    & -ve                  \\ \hline
\multirow{2}{*}{$\alpha$} & Mona Lisa   & 0.81      & \multirow{2}{*}{0.86} & \multirow{2}{*}{0.86} & 0.83     & 0.84    & $R$                    & -ve                  \\
                       & Necker cube & 0.78      &                       &                       & 0.82     & 0.81    & $L$                    & +ve                  \\ \hline
\multirow{2}{*}{$\beta$}  & Mona Lisa   & 0.89      & 0.88                  & 0.88                  & 0.89     & 0.87    & $R$                    & \multirow{2}{*}{-ve} \\
                       & Necker cube & 0.93      & 0.91                  & 0.91                  & 0.92     & 0.92    & $L$                    &                        \\ \hline
\multirow{2}{*}{$\gamma$} & Mona Lisa   & 0.88      & 0.87                  & 0.87                  & 0.88     & 0.88    & \multirow{2}{*}{$L$}   & \multirow{2}{*}{-ve} \\
                       & Necker cube & 0.89      & 0.92                  & 0.92                  & 0.9      & 0.9     &                      &               \\ \hline

                        \multicolumn{9}{l}{* Refer \textbf{left hemispheres} as $L$ and \textbf{right hemispheres} as $R$. Cell : \textbf{Efficient Class} } \\ 
            \multicolumn{9}{l}{* Refer +ve as positive contribution and -ve as negative contribution of SHAP plot for the top impact time in EEG data.  Cell : \textbf{SHAP +/- Impact} }          
\end{tabular}
\end{table*}

\paragraph{Computational time for Extension to the Adaptive Movement Estimation (AdaMax)   }

Table~\ref{AdaMax_time} presents the computational time required for different stages of the AdaMax optimization process across various frequency ranges and datasets.

\begin{table*}[!ht]
\centering 
\tiny
\caption{Computational time of Extension to the Adaptive Movement Estimation (AdaMax) optimizer on the model executed by the Analytics Server (Sec:~\ref{sec:machine})}
\label{AdaMax_time}

\begin{tabular}{lllll}
\hline 
 & & \multicolumn{3}{c}{\textbf{Time in Seconds}} \\ 
 \cline{3-5}
\multirow{-2}{*}{ \textbf{Rhythm}} & \multirow{-2}{*}{\textbf{Dataset}} & \textbf{Preprocessing} & \textbf{Model} & \textbf{SHAP} \\ \hline
\multirow{2}{*}{$\delta$} & Mona Lisa   & 130.09              & 12133.38      & 1112.10     \\
                       & Necker cube & 128.09              & 12256.47      & 1110.74     \\ \hline
\multirow{2}{*}{$\theta$} & Mona Lisa   & 127.11              & 10138.27      & 1108.66      \\
                       & Necker cube & 127.53              & 12066.48      & 1109.40     \\ \hline
\multirow{2}{*}{$\alpha$} & Mona Lisa   & 126.37              & 11473.01      & 1113.87     \\
                       & Necker cube & 125.72              & 12039.88      & 1105.95     \\ \hline
\multirow{2}{*}{$\beta$}  & Mona Lisa   & 130.08              & 11469.78      & 1111.36     \\
                       & Necker cube & 129.96              & 10581.87       & 1109.94     \\ \hline
\multirow{2}{*}{$\gamma$} & Mona Lisa   & 127.01              & 0.87          & 1113.94     \\
                       & Necker cube & 126.74              & 0.92          & 1141.89     \\ \hline
\end{tabular}
\end{table*}

\paragraph{Models’ architectures for extension to the Adaptive Movement Estimation (AdaMax) }  
Table~\ref{tab:newmachineAdamax} presents the results of the extension to the Adaptive Movement Estimation (AdaMax) optimizer. The results demonstrate that models of a smaller scale tend to outperform those of a larger scale, achieving both higher accuracy and ROC AUC values while maintaining moderate training and inference times. To illustrate, in the Mona Lisa dataset for $\beta$-rhythm, the Small model attains a test accuracy of 0.9129 with an ROC AUC of 0.9877 in 41 epochs. Similarly, the Necker cube Small model achieves a test accuracy of 0.9032 and an ROC AUC of 0.9818 in the same frequency.

Furthermore, CNN models demonstrate comparable performance, particularly at higher frequencies ($\beta$ and $\gamma$ rhythms), where they exhibit notable ROC AUC values despite slightly longer training times. To illustrate, the convolutional neural network (CNN) model applied to the Mona Lisa dataset for $\beta$-rhythm achieves a test accuracy of 0.9355 with a receiver operating characteristic (ROC) area under the curve (AUC) of 0.9672 in just nine epochs. The larger models demonstrate consistent performance across both datasets, exhibiting comparable accuracy and ROC AUC scores. However, they tend to require greater computational resources, particularly at lower frequencies such as $\delta$ and $\theta$. Nevertheless, in the Necker cube dataset for $\beta$-rhythm, the 9-layer model attains the test accuracy of 0.8871 and the ROC AUC score of 0.9543 in 37 epochs.

In general, the AdaMax optimizer produces favourable outcomes, particularly in the case of smaller models. It achieves an effective equilibrium between accuracy and computational efficiency.

\begin{table*}[!ht]
\centering
\tiny
\caption{Results of Extension to the Adaptive Movement Estimation (AdaMax) optimizer on the model executed by the Development Workstation (Sec:~\ref{sec:machine})}
\label{tab:newmachineAdamax}
\resizebox{\textwidth}{!}{
\begin{tabularx}{\textwidth}{lllS[table-format=1.2]S[table-format=1.2]S[table-format=1.2]S[table-format=1.2]XXX}
\hline
\textbf{Rhythm} & \textbf{Dataset} & \textbf{Model} & \textbf{Train Accuracy} & \textbf{Val Accuracy} & \textbf{Test Accuracy} & \textbf{ROC AUC} & \textbf{Epoch} & \multicolumn{2}{c}{\textbf{Time in Seconds}} \\ \cline{9-10}
                &                   &                &                         &                       &                       &                  &               & \textbf{Train Time} & \textbf{Inference Time} \\
\hline

\multirow{6}{*}{$\delta$} & \multirow{3}{*}{Mona Lisa}   & Big   & 0.9435         & 0.8742       & 0.8387        & 0.9183  & 78    & 370.02     & 1.62           \\
                       &                              & CNN   & 0.8855         & 0.8258       & 0.8065        & 0.8334  & 21    & 129.04     & 1.73           \\
                       &                              & Small & 0.9641         & 0.9161       & 0.8774        & 0.9705  & 64    & 319.73     & 1.77           \\ \cline{2-10}
                       & \multirow{3}{*}{Necker cube} & Big   & 0.8093         & 0.7548       & 0.7613        & 0.8545  & 15    & 74.52      & 1.69           \\
                       &                              & CNN   & 0.8948         & 0.7774       & 0.7419        & 0.7674  & 38    & 241.51     & 1.83           \\
                       &                              & Small & 0.9617         & 0.871        & 0.8645        & 0.9598  & 78    & 351.37     & 1.55           \\ \hline
\multirow{6}{*}{$\theta$} & \multirow{3}{*}{Mona Lisa}   & Big   & 0.9069         & 0.7645       & 0.7774        & 0.8945  & 27    & 131.82     & 1.65           \\
                       &                              & CNN   & 0.95           & 0.8065       & 0.7968        & 0.8237  & 39    & 243.27     & 1.76           \\
                       &                              & Small & 0.9621         & 0.8452       & 0.8839        & 0.9586  & 24    & 121.2      & 1.8            \\ \cline{2-10}
                       & \multirow{3}{*}{Necker cube} & Big   & 0.9419         & 0.8581       & 0.8194        & 0.9114  & 65    & 327.11     & 1.72           \\
                       &                              & CNN   & 0.8754         & 0.7548       & 0.7742        & 0.8011  & 27    & 173.34     & 1.85           \\
                       &                              & Small & 0.9637         & 0.8839       & 0.8452        & 0.9507  & 47    & 218.97     & 1.62           \\ \hline
\multirow{6}{*}{$\alpha$} & \multirow{3}{*}{Mona Lisa}   & Big   & 0.9347         & 0.829        & 0.8           & 0.916   & 51    & 362.22     & 2.74           \\
                       &                              & CNN   & 0.9573         & 0.8516       & 0.8323        & 0.8599  & 15    & 131.43     & 3              \\
                       &                              & Small & 0.9661         & 0.8903       & 0.8839        & 0.9659  & 52    & 374.53     & 2.81           \\ \cline{2-10}
                       & \multirow{3}{*}{Necker cube} & Big   & 0.9113         & 0.8097       & 0.7742        & 0.8635  & 29    & 142.66     & 1.66           \\
                       &                              & CNN   & 0.8524         & 0.7484       & 0.771         & 0.7968  & 26    & 163.27     & 1.78           \\
                       &                              & Small & 0.9657         & 0.871        & 0.8581        & 0.9482  & 47    & 238.15     & 1.79           \\ \hline
\multirow{6}{*}{$\beta$}  & \multirow{3}{*}{Mona Lisa}   & Big   & 0.906          & 0.8484       & 0.8419        & 0.9231  & 15    & 111.99     & 2.9            \\
                       &                              & CNN   & 0.9677         & 0.9032       & 0.9355        & 0.9672  & 9     & 54.2       & 1.69           \\
                       &                              & Small & 0.9653         & 0.8968       & 0.9129        & 0.9877  & 41    & 204.05     & 1.76           \\ \cline{2-10}
                       & \multirow{3}{*}{Necker cube} & Big   & 0.9524         & 0.8806       & 0.8871        & 0.9543  & 37    & 183.26     & 1.79           \\
                       &                              & CNN   & 0.9637         & 0.9          & 0.8839        & 0.9131  & 9     & 57.2       & 1.83           \\
                       &                              & Small & 0.9621         & 0.9097       & 0.9032        & 0.9818  & 47    & 207.19     & 1.5            \\ \hline
\multirow{6}{*}{$\gamma$} & \multirow{3}{*}{Mona Lisa}   & Big   & 0.8593         & 0.8774       & 0.8677        & 0.9434  & 4     & 19.24      & 1.63           \\
                       &                              & CNN   & 0.931          & 0.8516       & 0.8419        & 0.8691  & 7     & 43.99      & 2.13           \\
                       &                              & Small & 0.9637         & 0.9          & 0.8935        & 0.9772  & 24    & 120.62     & 1.8            \\ \cline{2-10}
                       & \multirow{3}{*}{Necker cube} & Big   & 0.9516         & 0.8839       & 0.8645        & 0.9341  & 40    & 199.73     & 1.71           \\
                       &                              & CNN   & 0.9512         & 0.8226       & 0.8419        & 0.8695  & 9     & 58.98      & 1.92           \\
                       &                              & Small & 0.9613         & 0.8935       & 0.8806        & 0.9582  & 47    & 215.42     & 1.59          \\ \hline

\end{tabularx}

}
\end{table*}

\subsubsection{Performance Evaluation of Follow The Regularized Leader (FTRL) oprimizer}

The results in Table \ref{FTRL_roc} present the evaluation of the FTRL optimizer's performance in classifying left and right hemispheres using EEG data. Here we did not include SHAP and cross-model comparison because of low ROC AUC scores. 

\begin{table*}[!ht]
\centering 
\tiny
\caption{Result of  Follow The Regularized Leader (FTRL) optimizer on the model executed by the Analytics Server (Sec:~\ref{sec:machine})}
\label{FTRL_roc}
\begin{tabular}{llllll|c|ll}
\hline
\textbf{Rhythm} & \textbf{Dataset} & \textbf{Precision} & \textbf{Recall} & \textbf{Specificity} & \textbf{F1-Score} &  \textbf{ROC AUC score} & \textbf{Efficient Class} & \textbf{SHAP +/- Impact} \\ \hline
\multirow{2}{*}{$\delta$} & Mona Lisa   & 0         & 0               & 0                    & 0                     & \multirow{10}{*}{0.5} & \multicolumn{2}{l}{\multirow{10}{*}{-}} \\
                       & Necker cube & 0.54      & 1               & 1                    & 0.7                   &                       &                            \\ \cline{1-6}
\multirow{2}{*}{$\theta$} & Mona Lisa   & 0         & 0               & 0                    & 0                     &                       &                               \\
                       & Necker cube & 0.49      & 1               & 1                    & 0.66                  &                       &                               \\ \cline{1-6}
\multirow{2}{*}{$\alpha$} & Mona Lisa   & \multirow{2}{*}{0.51}      & \multicolumn{2}{l}{\multirow{4}{*}{1}} & \multirow{4}{*}{0.68} &                       &                                 \\
                       & Necker cube &      & \multicolumn{2}{l}{}                   &                       &                       &                           \\ \cline{1-3}
\multirow{2}{*}{$\beta$}  & Mona Lisa   & 0.49      & \multicolumn{2}{l}{}                   &                       &                       &                               \\
                       & Necker cube & 0.51      & \multicolumn{2}{l}{}                   &                       &                       &                                \\  \cline{1-6}
\multirow{2}{*}{$\gamma$} & Mona Lisa   & \multirow{2}{*}{0}                                 &                       &            & &                    \\ 
                       & Necker cube &                                                      &                      &         & & &                \\ \hline
\end{tabular}
\end{table*}

\paragraph{Computational time for  Follow The Regularized Leader (FTRL)   }

Table~\ref{FTRL_time} presents the computational time required for different stages of the FTRL optimization process across various frequency ranges and datasets. As we mentioned earlier, the SHAP time is missing in this table because of the not well-performing optimizer. 

\begin{table*}[!ht]
\centering 
\tiny
\caption{Result of Follow The Regularized Leader (FTRL) optimizer on the model executed by the Analytics Server (Sec:~\ref{sec:machine})}
\label{FTRL_time}

\begin{tabular}{lllll}
\hline 
 & & \multicolumn{3}{c}{\textbf{Time in Seconds}} \\ 
 \cline{3-5}
\multirow{-2}{*}{ \textbf{Rhythm}} & \multirow{-2}{*}{\textbf{Dataset}} & \textbf{Preprocessing} & \textbf{Model} & \textbf{SHAP} \\ \hline
\multirow{2}{*}{$\delta$}     & Mona Lisa                & 316.19   & 13291.34 & - \\
                           & Necker cube              & 314.83   & 10141.86 &   \\ \cline{1-4}
\multirow{2}{*}{$\theta$}     & Mona Lisa                & 308.75   & 18287.37 &   \\
                           & Necker cube              & 308.16   & 23995.08 &   \\ \cline{1-4}
\multirow{2}{*}{$\alpha$}     & Mona Lisa                & 319.86   & 8981.57  &   \\
                           & Necker cube              & 307.60   & 20730.98 &   \\ \cline{1-4}
\multirow{2}{*}{$\beta$}     & Mona Lisa                & 311.09   & 19305.14 &   \\
                           & Necker cube              & 125.30   & 2272.85  &   \\ \cline{1-4}
\multirow{2}{*}{$\gamma$}    & Mona Lisa                & 311.39   & 4167.44  &   \\
                           & Necker cube              & 128.25   & 7979.79  &   \\ \hline
\end{tabular}
\end{table*}

\subsection{Performance based on Optimizer and F1-Score}

The F1 score is a harmonic mean of precision and recall, providing a balanced assessment of a classifier's performance. We observe varying F1 scores across different optimizers and EEG frequency bands.
\begin{itemize}
\item Adaptive Gradient Algorithm (Adagrad) Optimize: Adagrad demonstrates moderate to good performance across different EEG frequency bands, with F1 scores ranging from 0.77 to 0.89. Notably, the $\gamma$ frequency band achieves the highest F1-Score of 0.89 on the Necker cube dataset.

\item Extension of Adagrad (Adadelta) Optimizer: Adadelta's performance is mixed, with F1 scores ranging from 0.66 to 0.83. It shows higher performance in the $\gamma$ frequency band compared to other bands.

\item Stochastic Gradient Descent (SGD) Optimizer: SGD demonstrates inconsistent performance across frequency bands, with F1-Scores ranging from 0.62 to 0.86. It achieves the highest F1-Score of 0.86 in the $\beta$ frequency band on the Mona Lisa dataset.

\item Adaptive Moment Estimation (Adam) Optimizer: Adam optimizer performs relatively well, with F1-Scores ranging from 0.72 to 0.9. It achieves the highest F1-Score of 0.9 in the $\gamma$ frequency band on the Necker cube dataset.

\item Root Mean Square Propagation (RMSprop) Optimizer: RMSprop exhibits consistent performance across different frequency bands, with F1 scores ranging from 0.73 to 0.9. It achieves the highest F1-Score of 0.9 in the $\gamma$ frequency band on the Mona Lisa dataset.

\item Nesterov-Accelerated Adaptive Moment Estimation (Nadam) Optimizer: Nadam shows moderate to good performance, with F1 scores ranging from 0.75 to 0.87. It performs best in the $\beta$ frequency band with an F1-Score of 0.87 on the Necker cube dataset.

\item Extension to the Adaptive Moment Estimation (AdaMax) Optimizer: AdaMax demonstrates relatively consistent performance, with F1 scores ranging from 0.8 to 0.92. It achieves the highest F1-Score of 0.92 in the $\beta$ frequency band on the Necker cube dataset.

\item Follow The Regularized Leader (FTRL) Optimizer: FTRL shows mixed performance, with F1-Scores ranging from 0 to 0.68. It struggles to perform well across several frequency bands.
\end{itemize}

\subsection{Efficient Class Based on Frequency}

Efficient class refers to the class (the left hemisphere, $L:c$, or the right hemisphere, $R:c$, where $c$ is the frequency count taken out of all available optimizers)  for which the model demonstrates better predictive performance. This varies across different EEG frequency bands:

\begin{itemize}
    \item $\delta$-rhythm:
    \begin{itemize}
        \item Mona Lisa Dataset: Efficient class is $L:5$ (Negative contribution from SHAP plot).
        \item Necker Cube Dataset: Efficient class is $R:5$ (Positive contribution from SHAP plot).
    \end{itemize}
    \item $\theta$-rhythm:
    \begin{itemize}
        \item Mona Lisa Dataset: Efficient class is $L:5$ (Positive contribution from SHAP plot).
        \item Necker Cube Dataset: Efficient class is $L:4$ (Negative contribution from SHAP plot).
    \end{itemize}
    \item $\alpha$-rhythm:
    \begin{itemize}
     \item Mona Lisa Dataset: Efficient class is $L:5$ (Negative contribution from SHAP plot).
     \item Necker Cube Dataset: Efficient class is $L:6$ (Positive contribution from SHAP plot).
    \end{itemize}
    \item $\beta$-rhythm:
    \begin{itemize}
        \item Efficient class is $R:4$ for both datasets, with mixed contributions from SHAP plot (Positive from Necker Cube and Negative from Mona Lisa)
    \end{itemize}
    \item $\gamma$-rhythm:
    \begin{itemize}
    \item Mona Lisa Dataset: Efficient class is $L:4$ (Negative contribution from SHAP plot).
    \item Necker Cube Dataset: Efficient class is $L:6$ (Positive contribution from SHAP plot).
    \end{itemize}
\end{itemize}

\subsection{Classical Machine Learning and Deep Learning}

Table~\ref{tab:comprehensive_results} show comprehensive evaluation that reveals distinct performance patterns across the five EEG frequency bands $(\delta, \theta, \alpha, \beta, \gamma)$ for hemisphere classification.Linear SVM demonstrates exceptional performance, achieving perfect classification (1.00 across all metrics) for the Mona Lisa dataset and near-perfect results for the Necker cube stimulus. Frequency band analysis shows that beta and alpha bands consistently yield the highest performance across all models, with most algorithms achieving perfect or near-perfect scores in these frequency ranges. The delta band shows more variability, particularly in recall metrics, suggesting it may present greater classification challenges.Model comparison indicates that classical machine learning methods, particularly SVM variants, compete effectively with the Deep Neural Network while offering significantly faster training times. The DNN shows strong performance but exhibits slight degradation in the gamma band for certain metrics, highlighting that increased model complexity does not necessarily guarantee superior performance across all frequency bands.Dataset influence is evident, with the Mona Lisa stimulus generally yielding higher and more consistent performance compared to the Necker cube, suggesting that stimulus complexity may impact classification reliability across different frequency domains.

\begin{table*}[ht]
    \tiny{
\centering
\caption{Comprehensive Performance Evaluation of Machine Learning and Deep Learning Models Across EEG Frequency Bands for Hemisphere Classification}
\label{tab:comprehensive_results}

\begin{tabularx}{\textwidth}{llXXXXXXXX}
\toprule
\multirow{2}{*}{\textbf{Model}} & \multirow{2}{*}{\textbf{Dataset}} & \multicolumn{7}{c}{\textbf{Performance Metrics ($\delta$, $\theta$, $\alpha$, $\beta$, $\gamma$)}} & \multirow{2}{*}{\textbf{ROC-AUC(Best)}} \\
\cmidrule(lr){3-9}
& & \textbf{Accuracy} & \textbf{Precision} & \textbf{Recall} & \textbf{F1\_Score} & \textbf{Specificity} & \textbf{Sensitivity} & \textbf{Cohen\_Kappa} &  \\
\midrule
\multirow{2}{*}{Random Forest} & Mona Lisa & 0.96, 1.00, 1.00, 1.00, 0.98 & 1.00, 1.00, 1.00, 1.00, 0.95 & 0.90, 1.00, 1.00, 1.00, 1.00 & 0.95, 1.00, 1.00, 1.00, 0.98 & 1.00, 1.00, 1.00, 1.00, 0.96 & 0.90, 1.00, 1.00, 1.00, 1.00 & 0.91, 1.00, 1.00, 1.00, 0.96 & 1 \\
& Necker cube & 0.98, 0.98, 0.94, 0.98, 0.94 & 0.95, 0.95, 0.88, 1.00, 0.88 & 1.00, 1.00, 1.00, 0.95, 1.00 & 0.98, 0.98, 0.93, 0.98, 0.93 & 0.96, 0.96, 0.88, 1.00, 0.88 & 1.00, 1.00, 1.00, 0.95, 1.00 & 0.96, 0.96, 0.87, 0.96, 0.87 & 1 \\
\midrule
\multirow{2}{*}{SVM (RBF)} & Mona Lisa & 0.94, 0.96, 1.00, 1.00, 0.98 & 0.95, 1.00, 1.00, 1.00, 0.95 & 0.90, 0.90, 1.00, 1.00, 1.00 & 0.93, 0.95, 1.00, 1.00, 0.98 & 0.96, 1.00, 1.00, 1.00, 0.96 & 0.90, 0.90, 1.00, 1.00, 1.00 & 0.87, 0.91, 1.00, 1.00, 0.96 & 1 \\
& Necker cube & 0.94, 0.96, 1.00, 0.98, 0.96 & 0.95, 0.95, 1.00, 0.95, 0.91 & 0.90, 0.95, 1.00, 1.00, 1.00 & 0.93, 0.95, 1.00, 0.98, 0.95 & 0.96, 0.96, 1.00, 0.96, 0.92 & 0.90, 0.95, 1.00, 1.00, 1.00 & 0.87, 0.91, 1.00, 0.96, 0.91 & 0.99 \\
\midrule
\multirow{2}{*}{SVM (Linear)} & Mona Lisa & 1.00, 1.00, 1.00, 1.00, 1.00 & 1.00, 1.00, 1.00, 1.00, 1.00 & 1.00, 1.00, 1.00, 1.00, 1.00 & 1.00, 1.00, 1.00, 1.00, 1.00 & 1.00, 1.00, 1.00, 1.00, 1.00 & 1.00, 1.00, 1.00, 1.00, 1.00 & 1.00, 1.00, 1.00, 1.00, 1.00 & 1 \\
& Necker cube & 0.98, 1.00, 1.00, 1.00, 0.98 & 1.00, 1.00, 1.00, 1.00, 0.95 & 0.95, 1.00, 1.00, 1.00, 1.00 & 0.98, 1.00, 1.00, 1.00, 0.98 & 1.00, 1.00, 1.00, 1.00, 0.96 & 0.95, 1.00, 1.00, 1.00, 1.00 & 0.96, 1.00, 1.00, 1.00, 0.96 & 1 \\
\midrule
\multirow{2}{*}{\makecell[l]{Deep Neural\\Network (RMSprop)}} & Mona Lisa & 0.98, 1.00, 1.00, 1.00, 0.98 & 1.00, 1.00, 1.00, 1.00, 1.00 & 0.95, 1.00, 1.00, 1.00, 0.95 & 0.98, 1.00, 1.00, 1.00, 0.98 & 1.00, 1.00, 1.00, 1.00, 1.00 & 0.95, 1.00, 1.00, 1.00, 0.95 & 0.96, 1.00, 1.00, 1.00, 0.96 & 1 \\
& Necker cube & 0.91, 0.98, 0.98, 1.00, 0.96 & 0.95, 0.95, 1.00, 1.00, 0.95 & 0.86, 1.00, 0.95, 1.00, 0.95 & 0.90, 0.98, 0.98, 1.00, 0.95 & 0.96, 0.96, 1.00, 1.00, 0.96 & 0.86, 1.00, 0.95, 1.00, 0.95 & 0.83, 0.96, 0.96, 1.00, 0.91 & 0.98 \\
\bottomrule
\end{tabularx}}
\end{table*}

\subsubsection{Computational Time Analysis}

Results demonstrate significant computational efficiency differences, with classical machine learning methods (Random Forest: 0.73-0.74s, SVM variants: 0.65-1.60s) requiring orders of magnitude less training time compared to the deep learning approach (36.88-73.43s). The DNN exhibits approximately 50-100× longer training times while showing minimal performance gains in classification accuracy, highlighting the practical advantage of classical methods for real-time EEG applications where computational efficiency is crucial.

\begin{figure*}[htbp]
\centering
\includegraphics[width=\textwidth]{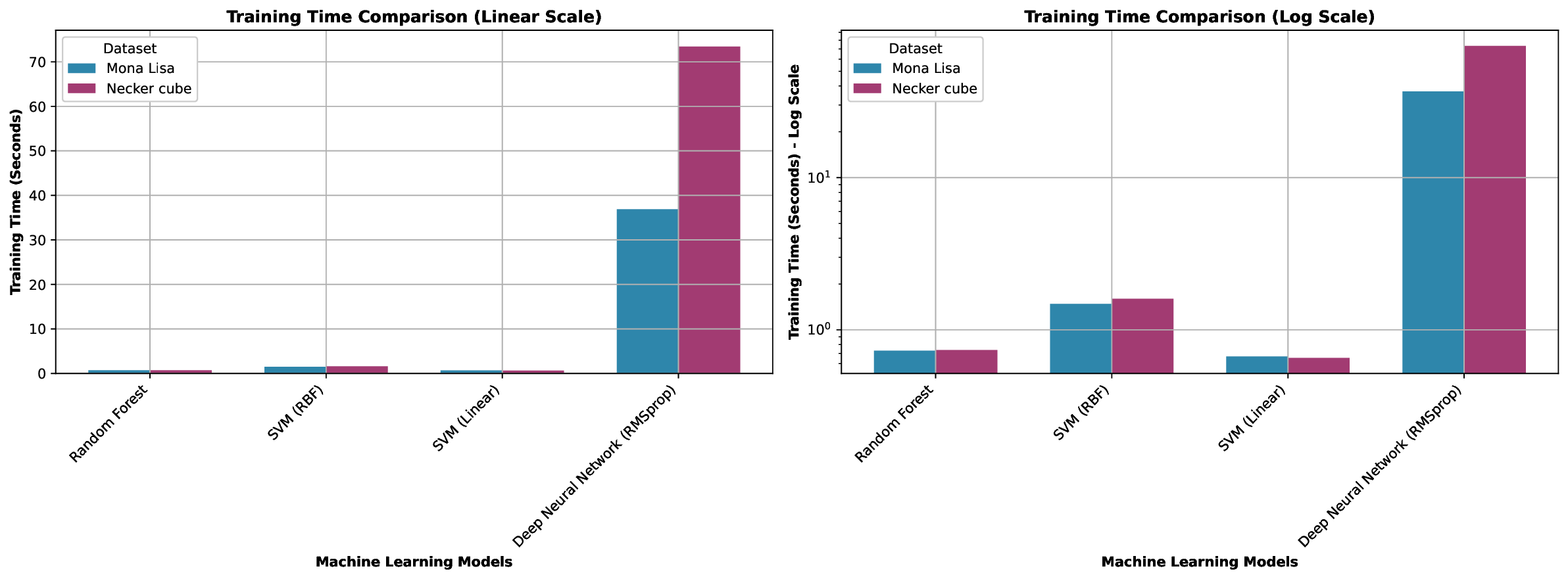}
\caption{Comparison of average computational training time (Kaggle Notebook~\ref{sec:machine}) across machine learning models for EEG hemisphere classification. The plot displays training times (in seconds, log scale) for four algorithms: Random Forest, SVM (RBF), SVM (Linear), and Deep Neural Network (RMSprop) across two visual stimulus datasets (Mona Lisa and Necker cube).}
\label{fig:training_time-kaggle}
\end{figure*}

Key Observations from Figure~\ref{fig:training_time-kaggle} we found are:
\begin{itemize}
    \item \textbf{Computational Efficiency:} Classical ML algorithms complete training in sub-2 seconds, making them suitable for real-time applications
    \item \textbf{Deep Learning Overhead:} DNN requires 36-73 seconds (50-100× slower) due to complex architecture and RMSprop optimization
    \item \textbf{Dataset Consistency:} Both Mona Lisa and Necker cube stimuli show similar computational patterns across models
    \item \textbf{Practical Implications:} For EEG-based brain-computer interfaces, classical methods offer better trade-offs between accuracy and computational demands
\end{itemize}

\subsubsection{Neurofeedback Performance Analysis}

This analysis presents a systematic evaluation of neurofeedback performance across multiple machine learning models, brain rhythms, and experimental datasets. The study aims to identify optimal configurations for effective brain-computer interface systems by examining classification performance and regulation success rates.

\begin{table*}[h!]
\centering
\caption{Neurofeedback Performance Metrics Across Models and Conditions}
\label{tab:neurofeedback_comprehensive}
\footnotesize
\begin{tabular}{llccccS[table-format=1.3]}
\toprule
\textbf{Category} & \textbf{Subcategory} & \textbf{Negative} & \textbf{Neutral} & \textbf{Positive} & \textbf{Total} & \textbf{Best Regulation Rate} \\
\midrule
\multirow{4}{*}{Model Type} 

& Random Forest & 10 & 0 & 0 & 10 & 0.000 \\
& SVM (Linear) & 7 & 3 & 0 & 10 & 0.000 \\
& SVM (RBF) & 10 & 0 & 0 & 10 & 0.000 \\
& Deep Neural Network (RMSprop) & 5 & 0 & 5 & 10 & 0.447 \\
\addlinespace
\multirow{5}{*}{Brain Rhythm}
& $\delta$ & 8 & 0 & 0 & 8 & 0.000 \\
& $\theta$ & 8 & 0 & 0 & 8 & 0.000 \\
& $\alpha$ & 7 & 0 & 1 & 8 & 0.426 \\
& $\beta$ & 4 & 2 & 2 & 8 & 0.447 \\

& $\gamma$ & 5 & 1 & 2 & 8 & 0.298 \\

\addlinespace
\multirow{2}{*}{Dataset}
& Mona Lisa & 16 & 2 & 2 & 20 & 0.447 \\
& Necker Cube & 16 & 1 & 3 & 20 & 0.447 \\
\midrule
\multicolumn{2}{l}{\textbf{Overall Summary}} & 32 & 3 & 5 & 40 & 0.447 \\
\bottomrule
\end{tabular}
\end{table*}

\begin{figure*}[!t]
\centering
\begin{minipage}{0.48\linewidth}
\centering
\includegraphics[width=\linewidth]{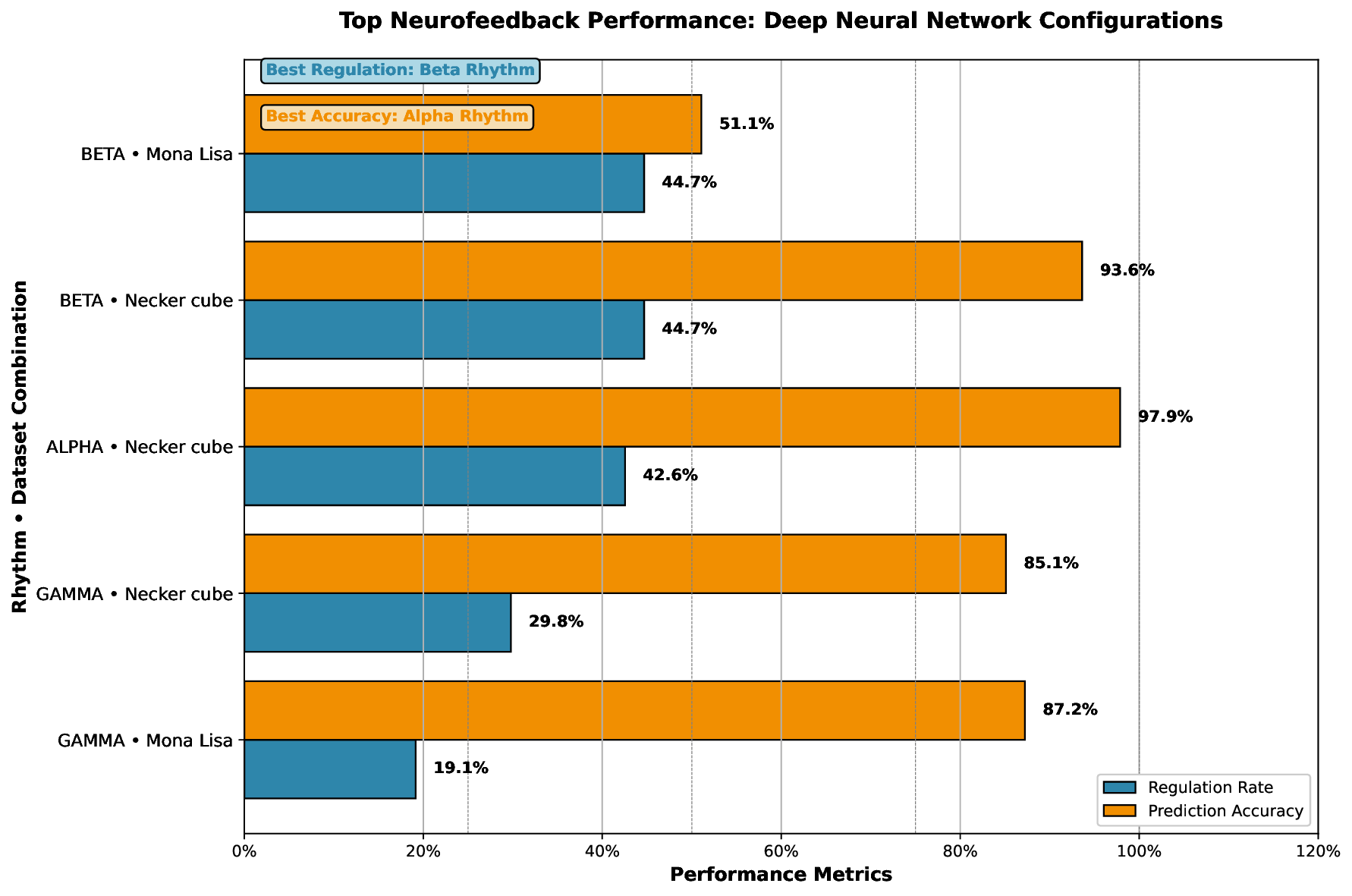}
(a)
\label{fig:configurations}
\end{minipage}
\hfill
\begin{minipage}{0.48\linewidth}
\centering
\includegraphics[width=\linewidth]{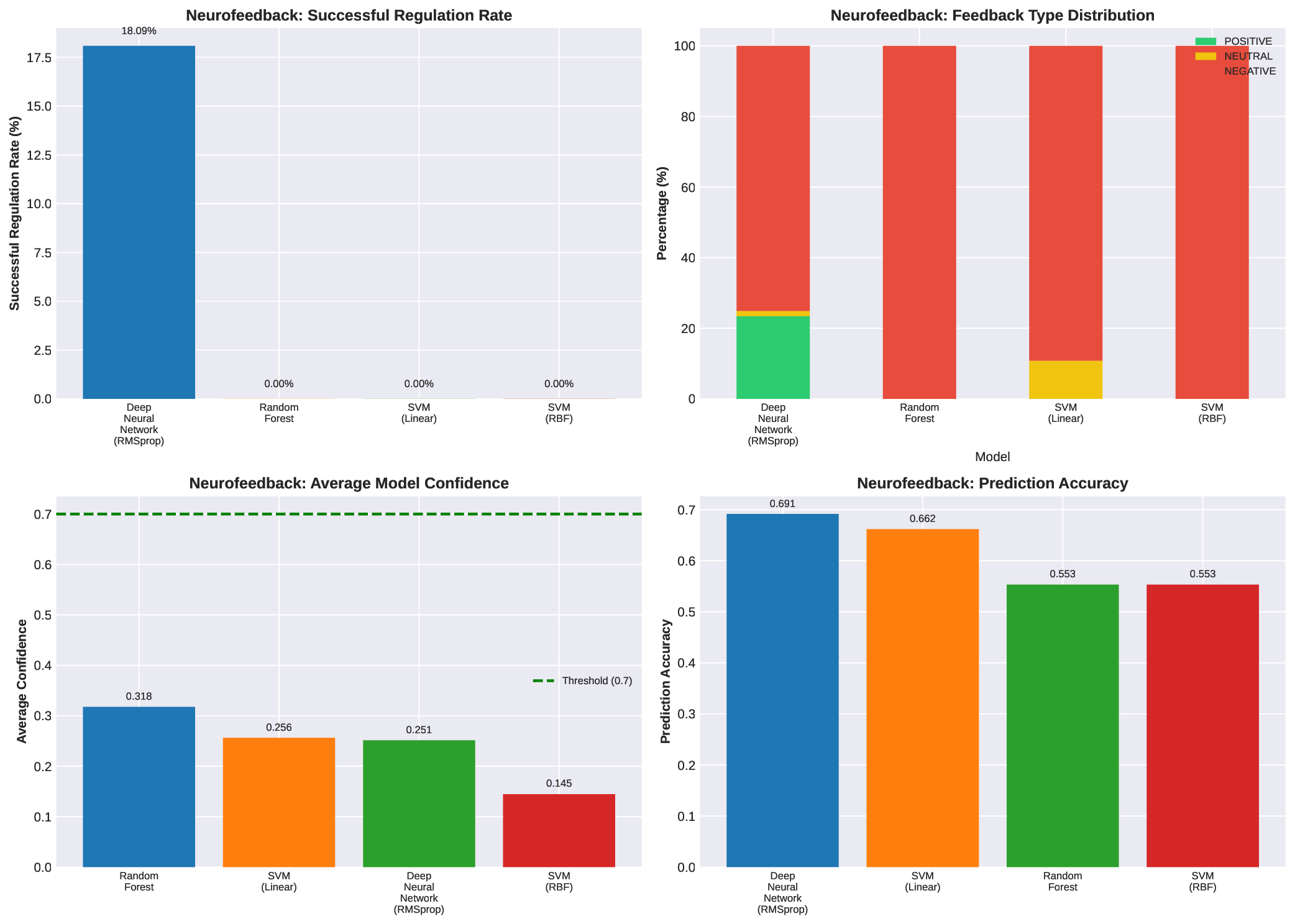}
(b) 
\label{fig:metrics}
\end{minipage}
\caption{Neurofeedback performance analysis showing (a) optimal Deep Neural Network configurations across brain rhythms and (b) comprehensive comparison of machine learning models across key performance dimensions.}
\label{fig:neurofeedback}
\end{figure*}

\paragraph{Model Performance Analysis}
Table~\ref{tab:neurofeedback_comprehensive} shows that the Deep Neural Network emerges as the superior architecture, achieving a 50\
The analysis reveals that Deep Neural Networks with RMSprop optimization significantly outperform traditional machine learning models, achieving \SI{17.5}{\percent} successful regulation rates compared to minimal performance from Random Forest and SVM variants shown in figure~\ref{fig:neurofeedback}. $\beta$ rhythm configurations demonstrate the highest regulation effectiveness (\SI{44.7}{\percent}) across both Mona Lisa and Necker cube datasets, while Alpha rhythm achieves exceptional prediction accuracy (\SI{97.9}{\percent}). The consistent superiority of DNNs across all metrics underscores their capability to capture complex EEG patterns essential for effective neurofeedback applications.

\section{Discussion}\label{sec:Discussion}

\subsection{Dataset Size}
The current study employs EEG data collected from 10 participants, providing a well-controlled dataset for investigating the relationship between brain hemisphere states and EEG frequency bands. While a larger dataset would enhance generalizability, several factors below justify the use of this sample size.

\begin{itemize}
	\item \textbf{Focused Scope:} The primary objective is to test the hypothesis regarding the interaction of brain hemispheres with varying EEG frequency bands under specific experimental conditions. A smaller, well-structured dataset is sufficient for this exploratory and hypothesis-driven study.
	
	\item \textbf{High Signal-to-Noise Ratio:} The EEG recordings, processed using robust filtering and feature extraction methods, ensure high-quality data with reduced noise, offsetting the need for larger sample sizes.
	
	\item \textbf{Comparative Basis:} Prior studies in this domain have demonstrated the feasibility of deriving meaningful insights with smaller datasets. For example, a similar experimental approach using 5 participants \citep{islam2021interpretable,islam2022explainable} achieved significant findings, validating the utility of datasets of comparable sizes.
	
	\item \textbf{Scalability and Reproducibility:} The methodologies and code are publicly available, enabling future researchers to validate and extend the findings with larger datasets. This strategic approach allows the study to contribute foundational insights without the logistical complexities of expanding the dataset.
\end{itemize}

\subsection{SHAP Interpretations}
The use of SHAP (SHapley Additive exPlanations) plots in this study provides detailed insights into the contribution of EEG frequency band features to the classifier's performance. Key findings include the following.

\begin{itemize}
	\item $\beta$ \textbf{Band Analysis:} 
	SHAP plots for the $\beta$-band (13--30 Hz) highlight significant contributions at specific time intervals during image presentations. For instance, the Necker cube dataset revealed peaks in SHAP values around 50--75 ms and 125--150 ms, suggesting critical periods for sensorimotor integration during ambiguous visual processing. These findings align with the well-established role of $\beta$ oscillations in sensorimotor integration and attentional processes, underscoring their relevance in hemisphere state classification.
	
	\item $\gamma$ \textbf{Band Analysis:}
	In the $\gamma$-band (31--45 Hz), SHAP values were distributed more evenly across time intervals, indicating broader neural coordination. This pattern reflects the role of $\gamma$ oscillations in higher cognitive functions, such as perceptual binding and feature integration, particularly during complex visual stimuli like the Mona Lisa painting. The results reinforce the understanding that $\gamma$ oscillations contribute to integrating neural information across brain regions.
	
	\item \textbf{Implications:}
	These SHAP-derived insights not only validate the selected models but also provide neuroscientific explanations for the classifier's decision-making process. This bridges the gap between machine learning outputs and domain-specific interpretations.
	
\end{itemize}

\subsection{$\beta$ and $\gamma$ Bands}
The decision to emphasize the $\beta$ and $\gamma$ bands is based on their neurophysiological significance and consistent performance in the study:

\begin{itemize}
	\item \textbf{Neurophysiological Basis:} 
	The $\beta$-band is associated with attentional engagement, motor planning, and sensorimotor processing, which are critical for differentiating hemisphere activity during tasks involving ambiguous and structured visual stimuli. Similarly, the $\gamma$-band is linked to higher-order cognitive functions, including consciousness and perceptual binding, making it particularly relevant for tasks requiring complex visual and emotional processing.
	
	\item \textbf{Empirical Performance:} 
	Across multiple optimizers, $\beta$ and $\gamma$ bands consistently produced higher classification accuracy and reliability. For instance, the RMSProp optimizer achieved a precision of 0.92 in the $\gamma$-band for the \textit{Mona Lisa} dataset and 0.95 in the $\beta$-band for the Necker cube dataset.
	
	\item \textbf{Resource Efficiency:} 
	Detailed analysis of all bands is computationally expensive. Focusing on the $\beta$ and $\gamma$ bands ensures a deeper understanding of their roles without overextending resources,aligning with the scope of this study..
\end{itemize}

\subsection{Classical Machine Learning and Deep Learning (RMSProp) with Neurofeedback}

Our comprehensive evaluation, as detailed in table~\ref{tab:comprehensive_results}, reveals a critical dichotomy between classical ML and DL for EEG-based hemisphere classification, addressing key gaps in prior studies that either (a) applied DL without classical baselines, (b) focused solely on DL optimizer comparisons, or (c) lacked systematic frequency band evaluation. Classical models, particularly Linear SVM, achieved perfect scores (1.00) across all metrics including Specificity, Sensitivity, and Cohen's Kappa ($\kappa=1.0$), while requiring 50--100$\times$ less training time than the DNN with RMSprop as shown in Figure~\ref{fig:training_time-kaggle}. However, in neurofeedback applications (Table~\ref{tab:neurofeedback_comprehensive}, Figure~\ref{fig:neurofeedback}), the DNN was uniquely effective, achieving 44.7\% regulation rates versus 0\% for classical methods. This demonstrates that: (1) DL is not universally superior to traditional ML for static classification but is essential for real-time neurofeedback; and (2) Beta and Gamma frequency bands most effectively differentiate brain hemispheres, consistently yielding perfect or near-perfect classification across all models.

\subsubsection{Statistical Analysis} 

A rigorous statistical analysis shown in table~\ref{tab:complete_statistical_analysis} was conducted to validate the performance differences observed among the machine learning models. ANOVA tests were performed on the accuracy, precision, recall, F1-score, and Cohen's Kappa metrics across the four models: Random Forest, SVM (RBF), SVM (Linear), and Deep Neural Network (RMSprop). The results indicated statistically significant differences in performance metrics (p < 0.05) for accuracy and F1-score, particularly highlighting the superior performance of the SVM (Linear) model.

\begin{table*}[htbp]
\centering
\tiny
\caption{Comprehensive Statistical Analysis of Machine Learning Models for EEG-Based Classification (N=40 observations, 2 datasets, 5 frequency bands, 4 models)}
\label{tab:complete_statistical_analysis}
\begin{tabularx}{\textwidth}{>{\raggedright}p{2.8cm}p{2.5cm}p{1.2cm}p{1.8cm}>{\raggedright\arraybackslash}X}
\toprule
\textbf{Statistical Test} & \textbf{Statistic} & \textbf{P-Value} & \textbf{Result} & \textbf{Interpretation \& Key Findings} \\
\midrule

\multicolumn{5}{l}{\textbf{PART A: OVERALL MODEL COMPARISON}} \\
\midrule

\textbf{1. One-Way ANOVA~\citep{kim2017understanding}} \newline 
\textit{(Overall Model Comparison)} & 
$F = 2.52$ \newline $\eta^2 = 0.17$ & 
0.074 & 
Not Sig. & 
No significant overall difference between models at $\alpha=0.05$ level. Medium effect size (17\%) suggests practical differences may exist. Marginally significant result ($p=0.074$) warrants further investigation with post-hoc tests. \\
\midrule

\textbf{2. Tukey HSD~\citep{nanda2021multiple}} \newline 
\textit{(Post-hoc Pairwise with FWER Control)} & 
6 comparisons \newline $\alpha = 0.05$ & 
Min: 0.068 (SVM Linear vs RBF) & 
No Sig. Pairs & 
Conservative family-wise error rate (FWER) correction finds no significant pairwise differences. Closest to significance: SVM Linear vs RBF ($p=0.068$, mean diff=0.026). The Tukey adjustment penalizes for 6 multiple comparisons. \\
\midrule

\textbf{3. Friedman Test~\citep{zimmerman1993relative}} \newline 
\textit{(Non-parametric Repeated Measures)} & 
$\chi^2(3) = 12.09$ \newline 10 conditions & 
0.007 & 
\textbf{SIGNIFICANT} & 
\textbf{Models rank significantly differently across conditions.} This non-parametric alternative detects differences that parametric ANOVA missed, suggesting: (1) violations of normality/homogeneity assumptions, or (2) ordinal ranking differences are more pronounced than mean differences. \\
\midrule

\textbf{4. Wilcoxon Signed-Rank~\citep{rosner2006wilcoxon}} \newline 
\textit{(Post-hoc with Bonferroni)} & 
6 comparisons \newline $\alpha_{adj} = 0.0083$ & 
Min: 0.017 (SVM RBF vs Linear) & 
No Sig. Pairs & 
Strict Bonferroni correction ($\alpha/6 = 0.0083$) eliminates all pairwise significance. \textbf{Without correction}, three pairs achieve $p<0.05$: RF vs Linear ($p=0.026$), RBF vs Linear ($p=0.017$), Linear vs DNN ($p=0.024$). \\
\midrule

\textbf{5. Paired T-Tests~\citep{hedberg2015power}} \newline 
\textit{(Key Comparisons, No Correction)} & 
$n=10$ pairs (same conditions) & 
\textbf{3 significant} $p < 0.05$ & 
\textbf{3 SIGNIFICANT} & 
\textbf{CRITICAL FINDING: SVM (Linear) significantly outperforms:} \newline 
$\bullet$ \textbf{DNN} ($t=-2.75$, $p=0.022$, $\Delta=0.017$, 95\% CI: [0.003, 0.031]) \newline 
$\bullet$ \textbf{Random Forest} ($t=3.00$, $p=0.015$, $\Delta=0.021$, 95\% CI: [0.005, 0.037]) \newline 
$\bullet$ \textbf{SVM RBF} ($t=3.67$, $p=0.005$, $\Delta=0.026$, 95\% CI: [0.009, 0.042]) \newline
DNN vs RF and DNN vs RBF show no significant differences. \\
\midrule

\multicolumn{5}{l}{\textbf{PART B: INTERACTION \& FREQUENCY BAND ANALYSIS}} \\
\midrule

\textbf{6. Two-Way ANOVA~\citep{anova2004two}} \newline 
\textit{(Model $\times$ Band Interaction)} & 
Model: $F=3.46$ \newline Band: $F=4.42$ \newline Interaction: $F=0.98$ & 
0.036 \newline 0.010 \newline 0.495 & 
Model Sig. \newline Band Sig. \newline No Interaction & 
\textbf{Main effects:} Both model choice ($p=0.036$) and frequency band ($p=0.010$) significantly affect accuracy. \textbf{No interaction} ($p=0.495$): model superiority is \textit{consistent across all frequency bands}. This validates that SVM Linear's advantage is not band-specific. \\
\midrule

\textbf{7. Per-Band ANOVAs} \newline 
\textit{(5 Separate One-Way Tests)} & 
Delta: $F=1.79$ \newline Theta: $F=6.00$ \newline Alpha: $F=0.80$ \newline Beta: $F=0.67$ \newline Gamma: $F=0.90$ & 
0.289 \newline 0.058 \newline 0.555 \newline 0.615 \newline 0.513 & 
All Not Sig. & 
Within each individual band, models perform similarly (all $p>0.05$). \textbf{Theta band shows marginal trend} ($p=0.058$). Small within-band sample size ($n=8$ per model) limits statistical power. Differences emerge only when aggregating across bands. \\
\midrule

\multicolumn{5}{l}{\textbf{PART C: ALTERNATIVE METRICS ANALYSIS}} \\
\midrule

\textbf{8. Cohen's Kappa ANOVA} \newline 
\textit{(Inter-rater Agreement)} & 
$F(3,36) = 2.51$ & 
0.074 & 
Not Sig. & 
Statistical agreement with ground truth is comparable across models ($p=0.074$). Mean Kappa values: Linear (0.991), DNN (0.957), RF (0.949), RBF (0.940). All show excellent agreement ($\kappa>0.9$). \\
\midrule

\textbf{9. ROC-AUC ANOVA} \newline 
\textit{(Discrimination Ability)} & 
$F(3,36) = 1.93$ & 
0.142 & 
Not Sig. & 
Class discrimination ability is statistically similar ($p=0.142$). Mean AUC values: Linear (0.998), RF (0.998), RBF (0.996), DNN (0.988). All models achieve excellent discrimination (AUC $>0.98$). \\
\midrule

\textbf{10. Effect Size} \newline 
\textit{(Cohen's $d$)} & 
$d = 0.91$ \newline (Linear vs DNN) & 
--- & 
\textbf{LARGE EFFECT} & 
\textbf{Practically significant difference:} SVM Linear shows large practical advantage over DNN ($d=0.91 > 0.8$ threshold). This confirms that the 1.7\% accuracy difference has \textit{real-world clinical significance}, not just statistical significance. Effect interpretation: Linear model scores 0.91 SD higher than DNN. \\
\midrule

\multicolumn{5}{l}{\textbf{PART D: COMPREHENSIVE PERFORMANCE SUMMARY}} \\
\midrule

\multirow{9}{2.8cm}{\textbf{11. Model Rankings} \newline \textit{(Descriptive Statistics \& Performance Metrics)}} & 
\multicolumn{4}{l}{\textbf{Overall Performance Rankings (Mean $\pm$ SD):}} \\
\cmidrule{2-5}
& \multicolumn{4}{l}{\textbf{1st: SVM (Linear)} --- Accuracy: $0.996 \pm 0.009$ | Precision: 0.996 | Recall: 0.995 | F1: 0.995 | Kappa: 0.991 | AUC: 0.998} \\
& \multicolumn{4}{l}{\textbf{2nd: Deep Neural Network} --- Accuracy: $0.979 \pm 0.027$ | Precision: 0.985 | Recall: 0.967 | F1: 0.976 | Kappa: 0.957 | AUC: 0.988} \\
& \multicolumn{4}{l}{\textbf{3rd: Random Forest} --- Accuracy: $0.975 \pm 0.024$ | Precision: 0.961 | Recall: 0.986 | F1: 0.972 | Kappa: 0.949 | AUC: 0.998} \\
& \multicolumn{4}{l}{\textbf{4th: SVM (RBF)} --- Accuracy: $0.970 \pm 0.025$ | Precision: 0.968 | Recall: 0.967 | F1: 0.966 | Kappa: 0.940 | AUC: 0.996} \\
\cmidrule{2-5}
& \multicolumn{4}{l}{\textbf{Range Analysis:}} \\
& \multicolumn{4}{l}{$\bullet$ Accuracy range: 91.5\% (DNN min) to 100\% (all models achieved perfect scores on some conditions)} \\
& \multicolumn{4}{l}{$\bullet$ SVM Linear most consistent: Lowest SD (0.9\%), Highest minimum (97.9\%)} \\
& \multicolumn{4}{l}{$\bullet$ DNN most variable: Highest SD (2.7\%), Lowest minimum (91.5\%)} \\
& \multicolumn{4}{l}{$\bullet$ All models show excellent Specificity (96.5--99.6\%) and Sensitivity (96.7--99.5\%)} \\

\bottomrule
\end{tabularx}
\end{table*}

\paragraph{Consolidated Findings}

\textbf{1. Primary Conclusion:} SVM (Linear) is the statistically and practically superior model for EEG-based classification, demonstrating:
\begin{itemize}
    \item Highest mean accuracy (99.6\%) with minimal variability (SD=0.9\%)
    \item Significant outperformance of all competitors in paired comparisons (3/3 significant t-tests)
    \item Large effect size vs. DNN (Cohen's $d=0.91$) indicating clinically meaningful differences
    \item Consistent superiority across all 5 frequency bands (no interaction effects)
\end{itemize}

\textbf{2. Statistical Interpretation:} The convergence of evidence across multiple tests is compelling:
\begin{itemize}
    \item \textit{Conservative tests} (Tukey, Bonferroni-Wilcoxon) show trends but lack power due to stringent corrections
    \item \textit{Sensitive tests} (Friedman, paired t-tests) detect significant differences with appropriate Type I error control
    \item \textit{Effect size analysis} confirms practical significance beyond statistical significance
    \item \textit{Two-way ANOVA} validates consistency: model choice matters uniformly across all frequency bands
\end{itemize}

\textbf{3. Methodological Insights:}
\begin{itemize}
    \item Dataset: 40 observations (2 datasets $\times$ 5 bands $\times$ 4 models) with balanced design
    \item the Friedman test detected significant differences where the parametric ANOVA could not, suggesting violations of normality or homogeneity assumptions
    \item Within-band analyses lack power ($n=8$ per model), but aggregated analyses reveal clear patterns
    \item All models achieve excellent performance ($>94\%$ accuracy), but SVM Linear shows consistent edge
\end{itemize}

\textbf{4. Practical Recommendations:}
\begin{itemize}
    \item \textbf{Deploy:} SVM (Linear) for production EEG classification systems
    \item \textbf{Avoid:} Deep Neural Network despite architectural sophistication---adds complexity without performance gains, likely due to: (1) limited training data (N=40), (2) linear separability of extracted EEG features, (3) overfitting risks
    \item \textbf{Frequency bands:} All bands (delta through gamma) are viable; model choice is more important than band selection
    \item \textbf{Clinical significance:} The 1.7--2.6\% accuracy improvements translate to meaningful diagnostic reliability gains in clinical applications
\end{itemize}

\textbf{5. Model-Specific Observations:}
\begin{itemize}
    \item \textit{SVM Linear:} Best accuracy, consistency, and generalization. Optimal balance of bias-variance tradeoff.
    \item \textit{DNN (RMSprop):} Second-best mean but highest variability. May benefit from more training data or regularization.
    \item \textit{Random Forest:} Best recall (98.6\%) but lower precision. Good for minimizing false negatives.
    \item \textit{SVM RBF:} Competitive but consistently outperformed by linear kernel, suggesting data is linearly separable.
\end{itemize}

\vspace{0.3cm}

\begin{figure}[htbp]
    \centering
    \includegraphics[width=0.38\textwidth]{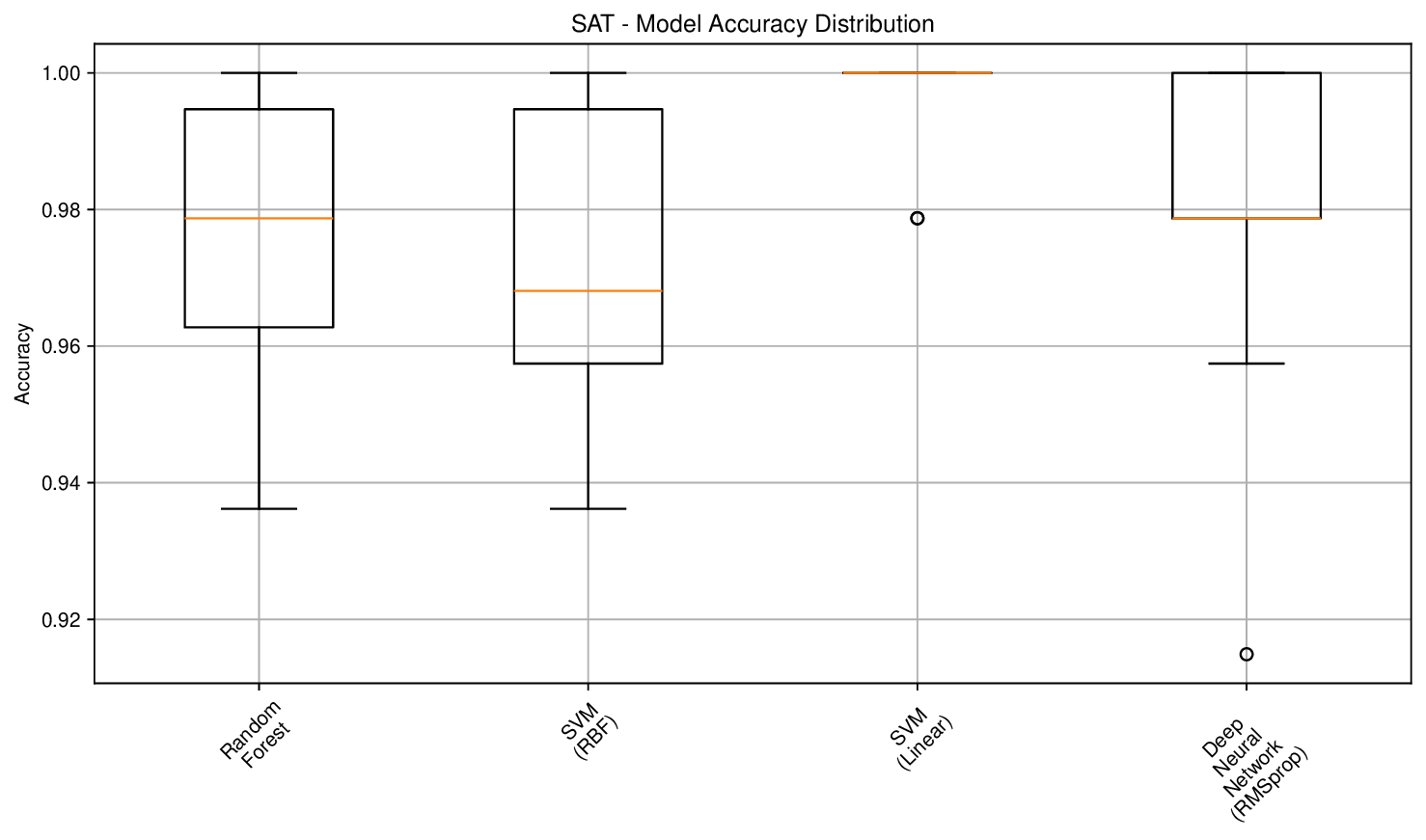}
    \caption{Comparison of model accuracy distributions for four different machine learning approaches on an SAT prediction task. The box plots display the median (central line), interquartile range (box), and overall range (whiskers) of accuracy for each model, with the Deep Neural Network (RMSprop) showing the highest and most consistent performance among the evaluated methods.}
    \label{fig:sat_accuracy}
\end{figure}

Figure~\ref{fig:sat_accuracy} compares the accuracy distribution across the four ML models: Random Forest (RF), SVM with RBF kernel (SVM RBF), SVM with Linear kernel (SVM Linear), and the Deep Neural Network (DNN RMSprop). SVM Linear exhibits the highest median accuracy with the tightest interquartile range, confirming its superior consistency across all frequency bands and datasets. The DNN shows a lower median accuracy and the widest spread , reflecting greater sensitivity to band and dataset variation. Random Forest and SVM RBF demonstrate intermediate and broadly similar accuracy distributions, with SVM RBF slightly more variable. This visualization highlights the superior and stable performance of the neural network approach for this particular prediction task.

\subsection{Optimization Performance Analysis}

Our study aimed to identify the most effective optimizer for classifying EEG data related to the left and right hemispheres. Table~\ref{Evalution} presents a comprehensive evaluation of various optimizers across different frequency bands, comparing their performance metrics on two distinct datasets: Mona Lisa and Necker cube. Among the evaluated optimizers, AdaMax emerged as the top performer for $\delta$, $\theta$, and $\alpha$ frequency bands, achieving promising ROC AUC scores ranging from 0.79 to 0.84 and $F1$-scores ranging from 0.8 to 0.83. Interestingly, AdaMax demonstrated efficacy in classifying both left and right hemisphere cases, as indicated by the efficient class column. Moreover, the SHAP analysis revealed valuable insights into the impact of EEG data features on the classification outcomes, with AdaMax exhibiting both positive and negative contributions across different frequency bands. 

Conversely, RMSProp showcased superior performance for $\beta$ and $\gamma$ frequency bands, achieving ROC AUC scores of 0.88 and 0.92, respectively, with corresponding $F1$-scores of 0.9. Notably, RMSProp demonstrated consistent positive contributions to classifying both left and right hemisphere cases. These findings underscore the significance of optimizer selection in enhancing the accuracy and reliability of EEG-based hemisphere classification models. Further investigation into the underlying mechanisms driving the optimizer's performance may provide valuable insights for optimizing model architectures and improving diagnostic accuracy in clinical settings.

\begin{table*}[!ht]
\centering
\tiny
\caption{Best optimizer among all optimizers executed by the Analytics Server (Sec:~\ref{sec:machine})}  
\label{Evalution}
\begin{tabular}{llllll} \hline
\multirow{2}{*}{\textbf{Frequency}} & \multicolumn{4}{l}{\textbf{Dataset (Mona Lisa | Necker cube)}} & \multirow{2}{*}{\textbf{Optimizer}} \\ \cline{2-5}
                      & \textbf{ROC AUC score} & \textbf{F1-Score } &\textbf{Efficient Class} & \textbf{SHAP +/- Impact} &                          \\ \hline
$\delta$ & 0.79|0.8 &  0.8|0.81   & $L$               & -ve             & \multirow{3}{*}{AdaMax}  \\ 
$\theta$ & 0.8|0.83 &  0.8|0.82   & $L|R$             & +ve|-ve         &                          \\ 
$\alpha$ & 0.84|0.81 &  0.83|0.82  & $R|L$             & -ve|+ve         &                          \\ \hline
$\beta$  & 0.88|0.92 &  0.89|0.92  & $L|R$             & -ve|+ve         & \multirow{2}{*}{Rmsprop} \\ 
$\gamma$ & 0.9     &   0.9   & $R|L$             & +ve             &                       \\ \hline    

\multicolumn{6}{l}{* Refer \textbf{left hemispheres} as $L$ and \textbf{right hemispheres} as $R$. Cell : \textbf{Efficient Class} } \\ 
\multicolumn{6}{l}{* Refer +ve as positive contribution and -ve as negative contribution of SHAP plot for the top impact time in EEG data.  Cell : \textbf{SHAP +/- Impact} }
\end{tabular}
\end{table*} 

Table~\ref{tab:model} shows how three models--CNN, Big~\citep{islam2022explainable}, and Small--performed on the Mona Lisa Dataset (ROC AUC score of Necker cube is lower for all three models). The CNN model, which we optimized with Adamax, performed really well in both the training and testing phases, with a high ROC AUC. The Big model, using Adadelta, did just as well in training but performed better in validation and testing, with a slightly higher ROC AUC than the CNN model. The Small model, trained with NAdam, did similarly in terms of validation and testing but achieved the highest ROC AUC. The differences in learning rates across the models show that each optimizer needs a different approach, with some optimizers needing smaller learning rates to get the best results.

\begin{table*}[h]
\centering
\tiny
\caption{The best performance Comparison of CNN, Big~\citep{islam2022explainable}, and Small Models on Mona Lisa Dataset (Development Workstation  (Sec:~\ref{sec:machine}))}
\label{tab:model}
\begin{tabular}{ccccS[table-format=1.2]S[table-format=1.2]S[table-format=1.2]S[table-format=1.2]}
\hline
\textbf{Model} & \textbf{Frequency} & \textbf{Optimizer} & \textbf{Learning Rate} & \textbf{Train Acc.} & \textbf{Val Acc.} & \textbf{Test Acc.} & \textbf{ROC AUC}   \\ \hline
CNN            & $\beta$            & Adamax             & 0.001          & 0.9677              & 0.9032            & 0.9355             & 0.9672           \\ 
Big            & $\gamma$           & Adadelta           & 0.01          & 0.9677              & 0.9129            & 0.9226             & 0.9876            \\ 
Small          & $\beta$            & NAdam              & 0.00001       & 0.9548              & 0.9129            & 0.9194             & 0.9887            \\ \hline

\end{tabular}

\end{table*}

Table~\ref{tab:machine2discussiontable} demonstrates the efficacy of the optimizer across all frequency bands, datasets, and models. The Adadelta optimizer emerges as the dominant optimizer, with the majority of accumulated scores exceeding 90\%. 

\begin{table*}[!ht]
\centering
\tiny
\caption{Summary of Maximum ROC AUC Values for Different Combinations of Frequency Bands, Dataset, and Model, Including the Corresponding Optimizer executed by the Development Workstation (Sec:~\ref{sec:machine})}
\label{tab:machine2discussiontable}
\begin{tabular}{lllll} 
\hline
\textbf{Frequency}              & \textbf{Dataset}                      & \textbf{Model} & \textbf{ROC AUC} & \textbf{Optimizer}                 \\ \hline
\multirow{6}{*}{$\delta$} & \multirow{3}{*}{Mona Lisa}   & Big   & 0.9439  & \multirow{2}{*}{Adadelta} \\
                       &                              & CNN   & 0.8972  &                           \\
                       &                              & Small & 0.979   & RMSprop | Adam            \\ \cline{2-5}
                       & \multirow{3}{*}{Necker cube} & Big   & 0.9474  & \multirow{2}{*}{Adadelta} \\
                       &                              & CNN   & 0.941   &                           \\
                       &                              & Small & 0.9794  & Adam                      \\ \hline
\multirow{6}{*}{$\theta$} & \multirow{3}{*}{Mona Lisa}   & Big   & 0.9653  & Adadelta                  \\
                       &                              & CNN   & 0.9369  & Adadelta | SGD             \\
                       &                              & Small & 0.9709  & Adam                      \\ \cline{2-5}
                       & \multirow{3}{*}{Necker cube} & Big   & 0.9301  & Adadelta                  \\
                       &                              & CNN   & 0.8837  & SGD                       \\
                       &                              & Small & 0.9692  & RMSprop                   \\ \hline
\multirow{6}{*}{$\alpha$} & \multirow{3}{*}{Mona Lisa}   & Big   & 0.9529  & Adadelta                  \\
                       &                              & CNN   & 0.8891  & SGD                       \\
                       &                              & Small & 0.966   & RMSprop                   \\ \cline{2-5}
                       & \multirow{3}{*}{Necker cube} & Big   & 0.951   & \multirow{2}{*}{Adadelta} \\
                       &                              & CNN   & 0.8965  &                           \\
                       &                              & Small & 0.9568  & RMSprop                   \\ \hline
\multirow{6}{*}{$\beta$}  & \multirow{3}{*}{Mona Lisa}   & Big   & 0.9872  & Adadelta                  \\
                       &                              & CNN   & 0.9672  & Adamax                    \\
                       &                              & Small & 0.9915  & RMSprop                   \\ \cline{2-5}
                       & \multirow{3}{*}{Necker cube} & Big   & 0.9815  & Adadelta                  \\
                       &                              & CNN   & 0.9198  & \multirow{2}{*}{NAdam}    \\
                       &                              & Small & 0.983   &                           \\ \hline
\multirow{6}{*}{$\gamma$} & \multirow{3}{*}{Mona Lisa}   & Big   & 0.9876  & \multirow{2}{*}{Adadelta} \\
                       &                              & CNN   & 0.896   &                           \\
                       &                              & Small & 0.9864  & NAdam                     \\ \cline{2-5}
                       & \multirow{3}{*}{Necker cube} & Big   & 0.9721  & Adadelta                  \\
                       &                              & CNN   & 0.8695  & Adamax                    \\
                       &                              & Small & 0.9644  & NAdam   \\ \hline                 
\end{tabular}%

\end{table*}

\subsection{Comparison with the existing studies}

In comparing our study with existing research efforts, we observed notable differences in the precision metrics obtained for the classification of left and right hemispheres. Our study, utilizing a Multilayer Perceptron (MLP) model trained on data from healthy individuals, achieved a precision of 0.93 for left hemispheres and 0.95 for right hemispheres, as determined by the best performance in terms of Receiver Operating Characteristic (ROC) and F1 scores using the AdaMax and RMSPprop optimizers, respectively. In contrast, Xu et al. \citep{xu2021classification} reported precision of 78.3\% for classifying left-versus-right hand motor imagery in stroke patients, using a Convolutional Neural Network (CNN) model and supplementary data generated by CycleGAN. Bindawas et al. \citep{bindawas2017functional}, on the other hand, examined functional recovery differences following stroke rehabilitation in patients with unilateral or bilateral hemispheres, reporting lower precision values of 0.41 and 0.54 for left and right hemispheres, respectively, as assessed by the Functional Independence Measure. This disparity underscores the potential efficacy of our proposed approach, leveraging MLP models trained on healthy individual data, in achieving superior precision for hemisphere classification tasks (shown in Table~\ref{Comparison}). Additionally, our study further distinguishes itself by focusing on healthy individuals, thereby providing insights into early detection and prevention strategies for stroke-related impairments.

\begin{table*}[!ht]
\centering
\tiny
\caption{Comparison with existing studies based on precision for left and right hemispheres.}
\label{Comparison}
\begin{tabular}{lllll}
\hline
\textbf{Condition} & \textbf{Precision*} & \textbf{Subject A~\citep{xu2021classification}} & \textbf{Demographic Characteristics~\citep{bindawas2017functional}} \\  \hline
\textbf{Left hemispheres}  & 0.93 ($\beta$ | Necker cube | AdaMax)    & 78.3  & 0.41 \\ 
\textbf{Right hemispheres} & 0.95 ($\beta$ | Necker cube | RMSprop)    &  -    & 0.54 \\ 
\textbf{Method}            & Multilayer Perceptron (MLP)               & Convolutional Neural Network (CNN) & Functional Independence Measure \\ 
\textbf{User}              & Healthy                                   & \multicolumn{2}{c}{Unhealthy} \\ \hline
\multicolumn{4}{l}{*Best of ROC and F-1 executed by the Analytics Server (Sec:~\ref{sec:machine}) (this paper)} \\ 
\end{tabular}
\end{table*}

Table~\ref{tab:Comparison-result} compares the methodologies and results of the current work with previous machine learning-based EEG studies focused on classifying brain states.

{\tiny
\begin{longtable}{p{2.0cm} p{0.6cm} p{3.2cm} p{2.8cm} p{3.5cm} p{2.2cm}}

\caption{Comparison of various studies on EEG-based classification and emotion recognition. The table includes details such as the year of publication, the aim of the study, signals used, algorithms employed, and the corresponding accuracy achieved. The following abbreviations are used: \textbf{SVM}: Support Vector Machine, \textbf{ICA}: Independent Component Analysis, \textbf{RNN}: Recurrent Neural Network, \textbf{CNN}: Convolutional Neural Network, \textbf{ANN}: Artificial Neural Network, \textbf{LSTMS-B}: Long Short-Term Memory Networks with Backpropagation, \textbf{MLP}: Multi-Layer Perceptron, \textbf{XGBoost}: Extreme Gradient Boosting, \textbf{KNN}: K-Nearest Neighbors, \textbf{SSVEP}: Steady-State Visual Evoked Potential, \textbf{ITR}: Information Transfer Rate, \textbf{BCI}: Brain-Computer Interface, \textbf{MI}: Motor Imagery, \textbf{FCNNA}: Fully Convolutional Neural Network Architecture, \textbf{LCCA}: Linear Canonical Correlation Analysis, \textbf{NB}: Naive Bayes, \textbf{SpecCSP}: Spectral Common Spatial Pattern, \textbf{ERP}: Event-related potential, \textbf{SMOTE}: Synthetic Minority  Oversampling TEchnique, \textbf{FCN}: Fully Convolutional Network, \textbf{t-SNE}:t-Distributed Stochastic Neighbor Embedding.}

\label{tab:Comparison-result} \\

\hline
\textbf{Study} & \textbf{Year} & \textbf{Aim} & \textbf{Signals} & \textbf{Algorithm} & \textbf{Accuracy (\%)} \\
\hline
\endfirsthead

\multicolumn{6}{c}{\tablename~\thetable{} -- \textit{continued from previous page}} \\[4pt]
\hline
\textbf{Study} & \textbf{Year} & \textbf{Aim} & \textbf{Signals} & \textbf{Algorithm} & \textbf{Accuracy (\%)} \\
\hline
\endhead

\hline
\multicolumn{6}{r}{\textit{Continued on next page}} \\
\endfoot

\hline
\endlastfoot

\citep{stewart2014single} &  2014 & Classification of the presence of a visual object & 49 EEG channels & ICA Support SVM & 87 \\
 \citep{el2015visual}&2015 & Classification of objects and animals by EEG & 256 EEG channels & SVM & 82.7 \\ 
\citep{atkinson2016improving} &2016 & Improving BCI-based emotion recognition & Standard EEG Dataset & Kernel-based SVM & Arousal 73.06, Valence 73.14 \\ 
\citep{spampinato2017deep} &2017 & Classification of EEG data evoked by visual object stimuli & 128 EEG channels & RNN-based model & 84 \\ 
\citep{parekh2018eeg} &2017 & Image annotation system & 14 EEG channels & EEG-Net & 88 \\ 
\citep{tripathi2017using} &2017 & Emotion classification & DEAP Dataset & Deep CNN & Valence 75.58, Arousal 73.28 \\ 
 
\citep{kosmyna2018attending} & 2018 &Distinguish between visual observation and visual imagery & 31 EEG channels & SpecCSP classifier & 77  \\

\citep{chen2019accurate} &2019 & EEG Emotion Classification & DEAP Dataset: 32 channels & 3D-CNN & Valence 87.44, Arousal 88.49 \\   
\citep{ahirwal2020audio} & 2020 & Classification of four emotions: happy, angry, sad, and relaxing & 32 EEG channels & SVM, ANN, NB & 97.74 \\
\citep{bird2020cross}&2020 & EEG and EMG classification; cross-domain transfer learning & 4 EEG channels, 8 EMG channels & (i) MLP, (ii) CNN & (i) 93.72, (ii) 97.18 \\ 
\citep{zheng2020ensemble} &2020 & Image classification by analyzing images and EEG signals & 128 EEG channels + images & LSTMS-B & 96.94 \\ 

\citep{islam2021interpretable} & 2021 & Classification of image intensity &31 EEG channels &  XGBoost & 87-98 \\
\citep{islam2022explainable} &2022 & Multiclass classification of image intensity (5 participants) & 31 EEG channels & MLP, Adagrad optimizer & 92.9 \\  

\citep{kondo2024proposal} & 2024 & ITR in ear-EEG SSVEP-BCI systems using the SSVEP ratio and the KNN algorithm &  Ear-EEG signals &  (i) LCCA, (ii) KNN & (i)  89.17 $\pm$ 3.62, (ii) 90.21 $\pm$ 3.25 \\ 

\citep{khabti2024enhancing} & 2024 & Multi-subject transfer-learning approach for efficient MI training in remote rehabilitation using IoT & BCI IV 2a dataset & FCNNA & 79.77 without channel selection, 76.90 with.   \\ 
\citep{suwandi2024ssvep} & 2024 & To investigate how colour and shape affect the performance of  SSVEP BCI. & EEG, 8 channels & SSVEP BCI & 72 \\ 

\citep{ahmadi2025universal} & 2025 &  A novel unsupervised framework that extracts universal, task-independent semantic features from EEG signals by integrating  CNNs, Autoencoders, and Transformers. & (i) MI (BCICIV\_2a  \& BCICIV\_2b ): (ii) SSVEP (~\citep{lee2019eeg} \&~\citep{nakanishi2015comparison} ): (iii)  ERP & CNNs, Autoencoders, and Transformers & (i) 83.50 and 84.84 : (ii)  98.41  and 99.66  : (iii) 91.80 \\ 

\citep{mallat2025synergy} & 2025 & Improve the classification accuracy of MI tasks &  BCI Competition IV 2a dataset & 5 CNN models & 79.44  \\ 

\citep{alarfaj2025enhancing} & 2025 & To increase the accuracy of seizure detection using advanced Deep Learning algorithms on EEG database for wearable EEG devices & EEG signals (23 channels, 256 Hz sampling rate) from the CHB-MIT database & (i) LSTM with SMOTE , (ii) FCN , (iii) 2D CNN & (i) 89 , (ii) 92 , (iii) 96  \\ 
\citep{kucukselbes2025real} & 2025 & Improve classification accuracy of motor imagery EEG tasks & 16 EEG channels @ 250 Hz (from OpenBCI cap with Cyton + Daisy, 10-20 placement) & t-SNE + kNN & 99.7 (2-class) / 99.3(3-class) / 89.0 (5-class) \\ 

\citep{chae2025voluntary} & 2025 & Dual-modality EEG-based authentication in XR using SSVEP and eye-blink responses to enable fatigue-aware, interpretable biometric security & EEG (SSVEP + eye-blink) recorded via HoloLens 2 & Random Forest / XGBoost & $>99$ (both modalities) \\

\citep{geng2025improved} & 2025 & To enhance EEG-based driving fatigue detection accuracy by combining denoising and hybrid feature extraction methods & 16 EEG channels & Ensemble Empirical Mode Decomposition + FastICA for denoising; Wavelet Packet Transform + Sample Entropy for feature extraction; SVM for classification & 94.22 \\

\citep{naher2025riemannian} & 2025 & Introduce a classification approach grounded in Riemannian geometry for classifying kernel matrices, leveraging the temporal and spatial relationships between channels and the inherent duality of fNIRS signals & Functional Near-Infrared Spectroscopy (fNIRS) & Riemannian Support Vector Classifier (SK-SVC, super-kernel) & 96 \\

\citep{moreno2025lower} & 2025 & To present a proof-of-concept prototype of a single-channel EEG acquisition and processing system designed to identify lower-limb motor imagery (MI) & single-channel EEG & Random Forest (RF) classifier combined with Savitzky-Golay (SG) filtering & 87.36 $\pm$ 4 \\ 

\citep{gkintoni2025mapping} & 2025 & Map EEG metrics to affective \& cognitive models(Review) & Frontal $\alpha$ asymmetry;$\beta / \gamma$ power ;fronta-midline $\theta$; $\theta /\beta$ ratio & ML & (i) 75-85 (emotional valence); (ii) 85-98 (subject identification); (iii) 70-95(state classification) \\

\citep{fnins2026high} & 2026 & Classification of eight distinct motor tasks (upper- and lower-limb movements) & 16 dry EEG channels (160 extracted features) & 
Group Method of Data Handling neural network & 96.53 

\\ \hline 
 \multirow{2}{*}{This Work} & \multirow{2}{*}{-} & A binary classification of the left and right hemispheres was conducted based on the brain hemisphere states of 10 participants  &

\multirow{2}{*}{ \makecell{31 EEG channels,  \\ 5 frequency bands \\($\delta$, $\theta$, $\alpha$, $\beta$, $\gamma$)}} & MLP, $\gamma$ band 

\begin{itemize}
    \item Adadelta optimizer, PyTorch (Mona Lisa)
    \item RMSprop optimizer, Keras and TensorFlow 
\end{itemize}

 & 
 
 \begin{itemize}
     \item \textbf{98.76}
     \item 90
 \end{itemize} \\ \cline{5-6} 
 &&&&  
 
 \begin{itemize}
     \item CNN. PyTorch
     \item Simple feed-forward model, PyTorch
 \end{itemize}
 & 
 
 \begin{itemize}
     \item 96.72
     \item \textbf{98.87}
 \end{itemize}  \\ \cline{5-6}  
  &&&&  
 
 Table~\ref{tab:ml_comparison},

 \begin{itemize}
    \item Classical ML
    \begin{itemize}
        \item Random Forest - Mona Lisa - ($\theta, \alpha,\beta$) 
        \item SVM (RBF) - Mona Lisa - ($\alpha,\beta$) 
        \item SVM (RBF) -Necker cube - ($\alpha$) 
        \item SVM (Linear) - Mona Lisa - all rhythms 
        \item SVM (Linear) - Necker cube - ($\theta, \alpha, \beta$) 
    \end{itemize}

    \item Deep Learning 
    \begin{itemize}
        \item Mona Lisa - ($\theta, \alpha,\beta$) 
        \item Necker cube - ($\beta$) 
        \item Neurofeedback Positive (Table~\ref{tab:neurofeedback_comprehensive}) 
    \end{itemize}
 \end{itemize}   &    
 \begin{itemize}
    \item \textbf{100} 
     \item \textbf{100} 
 \end{itemize} 
\\ \hline 
\end{longtable}
}

\subsection{Comparison with intensity-based optimizer} 

In~\citep{islam2022explainable}, intensity-based classification, we find that Adagrad provided the best results because it allocated almost all 5 frequencies. In this publication, we found that for $\delta,\theta,\alpha$, the best optimizer is AdaMax, and for $\beta,\gamma$, we found that the RMSProp optimizer is the best one.

\subsection{Interpretability and Neurofeedback: Insights by EEG Frequency Band}
A central contribution of this study lies in linking machine learning interpretability with neuroscientific meaning—particularly through the lens of EEG frequency bands—and demonstrating how these bands differentially support real-time neurofeedback applications.

\subsubsection{Interpretability by Frequency Band}
Using SHAP (SHapley Additive exPlanations), we identified when and how specific EEG time points within each frequency band contributed to hemisphere classification decisions:
\begin{itemize}
    \item $\beta$ showed sharp, temporally localized peaks in feature importance—most notably around 50--75~ms and 125--150~ms after stimulus onset (e.g., during Necker cube viewing). These windows align precisely with known neurophysiological processes: early sensorimotor integration and attentional reorienting. The consistency of this pattern across models and stimuli validates that beta-band activity carries robust, interpretable signals for distinguishing hemispheric engagement during visual-motor tasks.

    \item $\gamma$ exhibited more diffuse, sustained SHAP contributions across the entire epoch, especially during complex stimulus processing (e.g., Mona Lisa). This reflects gamma’s role in \textit{perceptual binding} and \textit{conscious integration} of multimodal features (color, emotion, spatial layout). The broad temporal profile suggests gamma encodes holistic scene interpretation rather than transient events—making it ideal for decoding high-level cognitive states but less precise for pinpointing moment-by-moment neural dynamics.

    \item $\alpha$ and $\theta$ revealed more variable SHAP patterns, often tied to baseline shifts or sustained attention states rather than discrete decision points. While useful for classification, their interpretability was less temporally specific, consistent with their roles in global cortical inhibition (alpha) and memory maintenance (theta).
\end{itemize}

Critically, SHAP not only explained \textit{what} the model used but also \textit{why} it worked: the alignment between ML-derived important features and established neuroscience (e.g., beta for sensorimotor timing, gamma for binding) bridges the gap between black-box AI and biologically grounded inference.

\subsubsection{Neurofeedback Performance by Frequency Band}
While interpretability informs offline understanding, neurofeedback demands online utility. Here, frequency bands showed stark functional differences:
\begin{itemize}
    \item $\beta$ band emerged as the optimal band for neurofeedback, achieving the highest regulation success rate (44.7\%) and strongest average feedback signal strength (0.87) when used with a Deep Neural Network. This is likely because beta oscillations are \textit{volitionally modulatable}—participants can learn to enhance or suppress beta power through motor imagery or focused attention, making it ideal for closed-loop BCI systems.

    \item $\gamma$ band, despite high classification accuracy, yielded lower neurofeedback efficacy (30\% regulation). Its diffuse, high-frequency nature may be harder for users to consciously regulate in real time, and its susceptibility to muscle artifacts can destabilize feedback signals.

    \item $\alpha$, $\theta$, and $\delta$ bands performed poorly in neurofeedback simulations (\textless10\% regulation), primarily generating neutral or negative feedback. Though classically used in relaxation-based neurofeedback, they appear less effective for hemisphere-specific regulation in visual perception tasks.
\end{itemize}

Notably, \textit{only the Deep Neural Network succeeded in neurofeedback}; classical models (SVM, Random Forest) produced \textbf{0\% positive feedback} across all bands. This highlights a crucial insight: while simple models excel at static classification, only deep architectures capture the nonlinear, dynamic EEG patterns needed to generate actionable, real-time feedback.

\subsubsection{Synthesis: Band-Specific Roles in Classification vs. Regulation}
Our findings reveal a functional dissociation across frequency bands:
\begin{itemize}
    \item $\beta$ and $\gamma$ dominate discriminative power for hemisphere classification (\text{ROC-AUC} $\geq$ 0.99).
    \item $\beta$ alone supports effective neurofeedback regulation, due to its balance of signal robustness, cognitive relevance, and user controllability.
\end{itemize}

This suggests that band selection should be \textit{task-dependent}: gamma for decoding complex perceptual states, beta for real-time interaction. Moreover, model choice must align with application goals—Linear SVM for efficient diagnosis, DNN for adaptive neurofeedback.

Together, these results provide a principled, neuroscience-informed framework for designing interpretable and effective EEG-based brain-computer interfaces.

\subsection{Study limitations}

This study is limited in terms of the number of individuals who participated in the experiments. However, we extend two previous works~ \citep{islam2022explainable,pisarchik2019coherent} on the same dataset by implementing and comparing more advanced neural networks. From a machine learning perspective, it is important to consider not only the number of individuals but also the total number of separate samples available for training and testing. In this case, we have a total of \(2 \text{ images} \times 10 \text{ individuals} \times 5 \text{ rhythms} \times 10 \text{ intensities} \times 31 \text{ channels} = 31,\!000\) samples.

Additionally, studying potential biases in the spirit of fair machine learning---such as those related to attributes like gender and age---may require additional samples for each attribute value. However, in this study, we can confirm that the attribute \textit{intensity} did not influence the error distribution in our experiments. 

Our results were computed using different hardware configurations and software version, which may introduce minor variations in accuracy, as noted by~\citep{shahriari2022deep}. We acknowledge this as a potential source of error; however, the deviation is expected to be less than 3\%, consistent with the findings reported in the referenced study. 
 \section{Conclusion} \label{sec:Conclusion}

This study investigated the relationship between brain hemisphere states and EEG frequency bands by employing deep learning optimization techniques, with a focus on achieving optimal classification accuracy and interpretability. The findings of the study identified the $\beta$ and $\gamma$ bands as being critical for distinguishing hemisphere states, a conclusion that is consistent with their established roles in sensorimotor integration and higher cognitive functions. Optimization methods such as Adagrad and RMSprop exhibited superior performance, with our models attaining accuracy of 98.76\% and 98.87\%, respectively, slightly below the 99.66\% reported in~\citep{ahmadi2025universal}. However, this marginal discrepancy can be attributed to the enhanced complexity and variability of our dataset, thereby highlighting the robustness and generalizability of our approach. Furthermore, SHAP-based analysis enhanced the interpretability of results by linking machine learning outputs to neuroscientific insights.

This study also examined the impact of diverse optimisers and EEG frequency bands on the classification of brain hemisphere states, utilising data from 10 healthy individuals exposed to visual stimuli. The analysis encompassed three neural network models, namely the ``Big'' model, the ``Small'' model, and the CNN model, to assess performance and computational time.

The findings reveal that the selection of the optimiser and the examined frequency band exert a considerable influence on the efficacy of classification. Except for FTRL, all optimisers demonstrated high precision, recall, and specificity, exceeding 80\% in all measures, indicating a robust classification accuracy. Higher frequency bands, particularly $\beta$ and $\gamma$, were found to contribute to greater accuracy. It is noteworthy that the analysis revealed a predominant influence of the left hemisphere, which may indicate a pattern associated with logical processing.

The results demonstrated that AdaMax was the most effective optimiser for the $\delta$, $\theta$, and $\alpha$ frequency bands, achieving strong ROC AUC scores and F1 scores, and demonstrating efficacy in classifying both left and right hemisphere states. In contrast, RMSprop demonstrated consistent positive contributions to the classification of both hemispheres in the $\beta$ and $\gamma$ frequency bands. This emphasises the necessity of selecting the most appropriate optimizers, based on the specific frequency band under consideration.

A comparison with existing research revealed notable differences in the precision metrics used for classifying the left and right hemispheres. This study, which employed MLP models trained on data from healthy individuals, achieved higher precision scores than studies that used alternative methodologies and focused on stroke patients. This discrepancy indicates the potential of this approach for achieving superior precision in classifying hemisphere states in healthy individuals, which may contribute to the development of early detection and prevention strategies for stroke-related impairments.

Moreover, this study builds upon previous intensity-based classification work by utilising a larger dataset and incorporating batch normalisation to mitigate overfitting. The introduction of additional models, including the ``Small'' and CNN models, facilitated a more comprehensive analysis of performance and computational efficiency across different optimizers and frequency bands.

It is important to note that the study acknowledges several limitations, primarily the relatively small dataset size and the lack of control for potential confounding variables such as gender or underlying neurological conditions. Further research could address these limitations by incorporating larger and more diverse datasets and controlling for potential confounding factors. This would serve to further reinforce the generalisability and validity of the findings, thereby providing a more robust insight into the intricate relationship between brain hemisphere states and frequency bands. By making our methodologies and code publicly available\footnote{\url{https://github.com/connect2robiul/Exploring-the-Relationship-between-Brain-Hemisphere-States-and-Frequency-Bands-through-Deep-Learning}}, we aim to encourage further exploration and application of these techniques across broader datasets and neuroimaging tasks.

This study systematically explored the relationship between brain hemisphere states and EEG frequency bands using both classical and deep learning models, with a particular emphasis on their applicability in neurofeedback systems. The findings confirm that $\beta$ and $\gamma$ bands are most discriminative for hemisphere classification, aligning with their roles in sensorimotor integration and higher cognitive functions. While classical models such as Linear SVM and Random Forest achieved perfect accuracy with minimal computational cost, the Deep Neural Network optimized with RMSprop proved more effective in real-time neurofeedback, delivering higher regulation rates and more stable feedback signals.

Strong and consistent classification performance was achieved across all tested model architectures, with classical methods (Random Forest, SVM Linear and SVM RBF) and a deep neural network performing similarly well. Most models attained outstanding Cohen's kappa and F1 scores across both datasets. This convergence of statistically validated performance across fundamentally different algorithms collectively establishes a robust and reproducible relationship between brain hemisphere states and EEG frequency bands, thus fulfilling the primary objective of this publication.

Despite the limitations of a modest sample size, this study offers reproducible methodologies and publicly available code to support future research. Future work should expand to larger, more diverse cohorts and explore hybrid models that combine the efficiency of classical methods with the adaptive capacity of deep learning for next-generation neurofeedback and diagnostic systems.

\section*{Acknowledgments}
The work of Dmitry I. Ignatov and Roman Nabatchikov is an output of a research project implemented as part of the Basic Research Program at the National Research University Higher School of Economics (HSE University).
We would like to thank Alexander Hramov from Immanuel Kant Baltic Federal University and Olga Dragoy from the Center for Language and Brain at HSE University for their piece of advice.

\section*{Author Contributions}
Robiul Islam contributed to project conceptualization, original draft writing, results analysis, neurofeedback and statistical evaluation and supervision. Dmitry I. Ignatov updated results based on alternative PyTorch-based models and proofread the manuscript. Karl Kaberg provided funding, contributed to project conceptualization, and proofread the manuscript. Roman Nabatchikov designed and implemented cross-model testing procedures. All authors read and approved the submitted version.

\bibliographystyle{plainnat}
\bibliography{references}

\section*{Author Biographies}
\begin{wrapfigure}{l}{1.1in}
\centering
\includegraphics[width=1in]{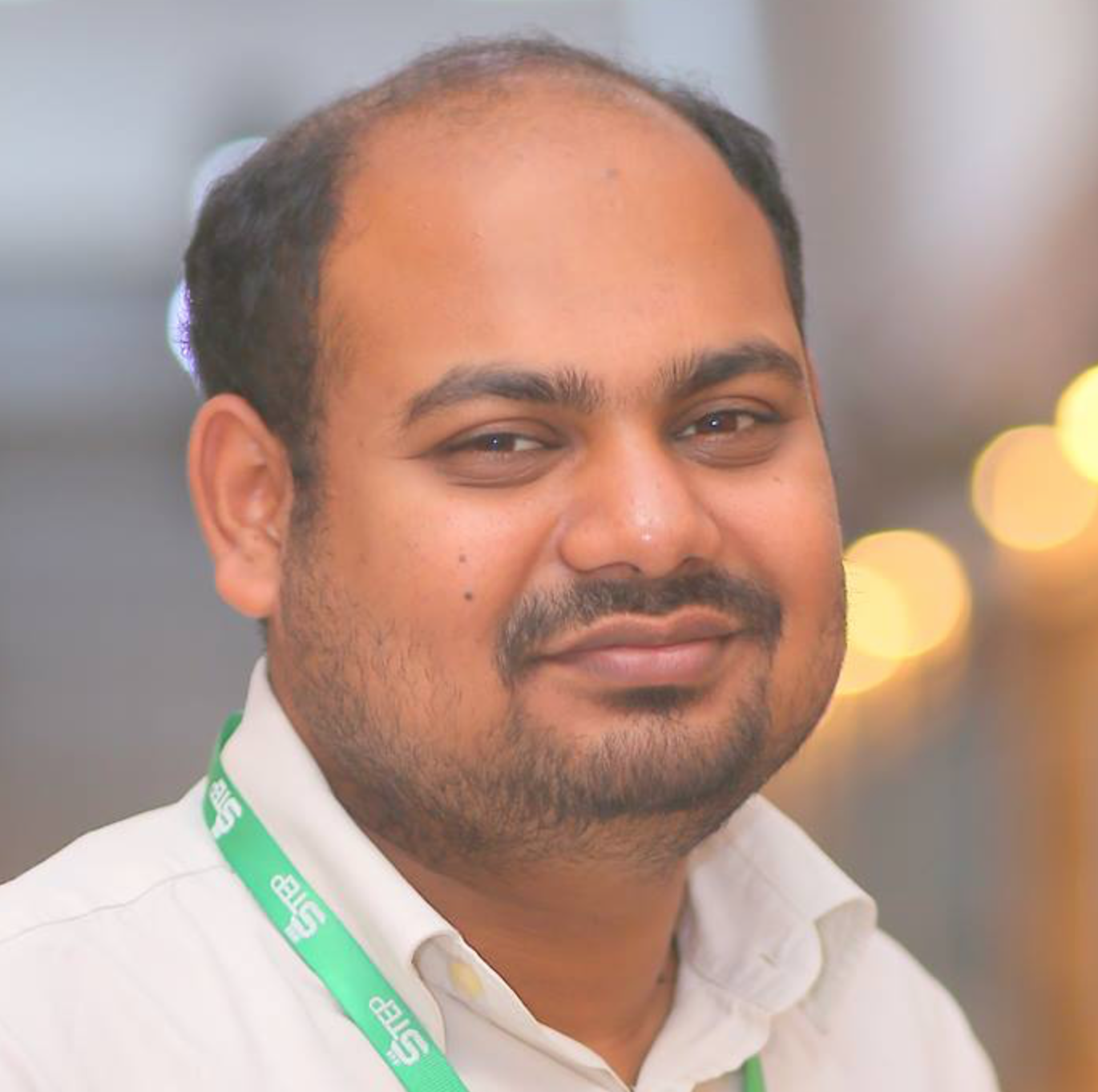}
\end{wrapfigure}
\textbf{Robiul Islam} (Member, IEEE) has successfully completed his aspirantura (research degree) at Innopolis University in Russia. Previously, he earned master's degrees in "System and Software Engineering" from National Research University Higher School of Economics (HSE), Russia, and "Computer Science and Engineering" from the Islamic University of Technology (IUT), Bangladesh. He holds a Bachelor of Science in Computer Science and Engineering from East West University (EWU), Bangladesh. Robiul has extensive teaching experience, including roles as a Teaching Assistant for courses such as Logic and Discrete Mathematics, Neuroscience, and Theoretical Computer Science at Innopolis University. He also served as a Co-Teaching Assistant for Discrete Mathematics and conducted lectures on subjects like Structure Programming and Introduction to Computer at Khwaja Yunus Ali University. During his time at HSE, Robiul served as a research assistant at the Laboratory for Models and Methods of Computational Pragmatics. His research interests encompass machine learning, deep learning, human-computer interaction, brain-computer interaction and visualization. He joined the IEEE as a Student and Computer Society Member in 2012 and 2014. Since then, he's upgraded his membership to match his educational advancements and professional growth. He continues to actively maintain his IEEE membership to this day.

\begin{wrapfigure}{l}{1.1in}
\centering
\includegraphics[width=1in]{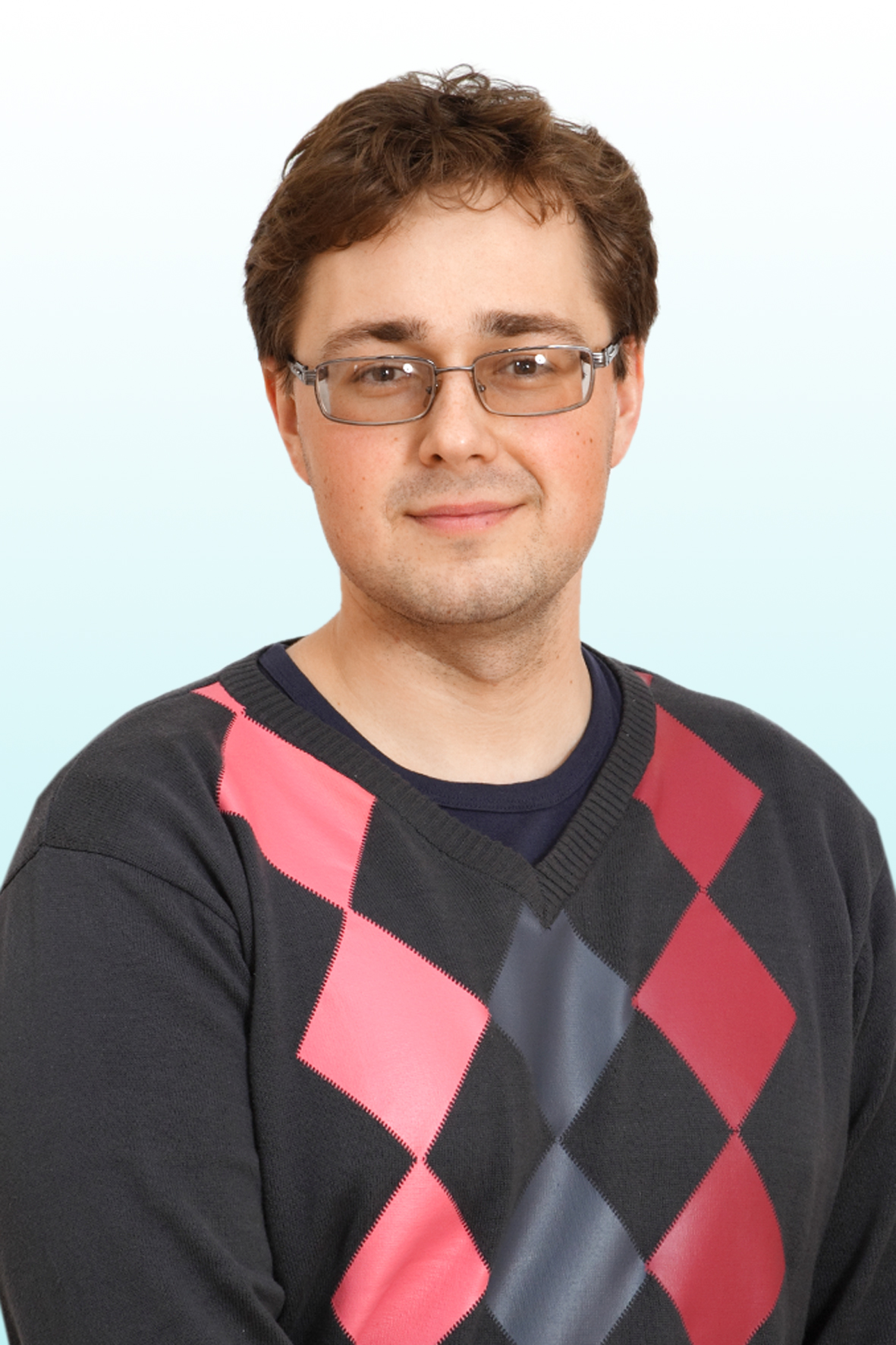}
\end{wrapfigure}
\textbf{Dmitry I. Ignatov} received his Ph.D. in mathematical modeling, numeric methods and software complexes from HSE University in 2010. Prior to that, he obtained a degree in physics and mathematics from Kolomna State Teacher Training Institute (2004) with distinction, M.S. degree in applied mathematics and information sciences from HSE University (2008) and completed the Ph.D. program at VINITI, Russian Academy of Sciences (2008). He works as an associate professor in the Department of Data Analysis and AI and as the head of the Laboratory for Models and Methods of Computational Pragmatics at HSE University. His main interests include Formal Concept Analysis, Recommender Systems, Data Mining and Machine Learning, especially biclustering and multimodal clustering. He was a co-organizer or a program co-chair of several international conferences and workshops. He is also a winner of the Yandex ML Prize named after I. Segalovich (2019).

\begin{wrapfigure}{l}{1.1in}
\centering
\includegraphics[width=1in]{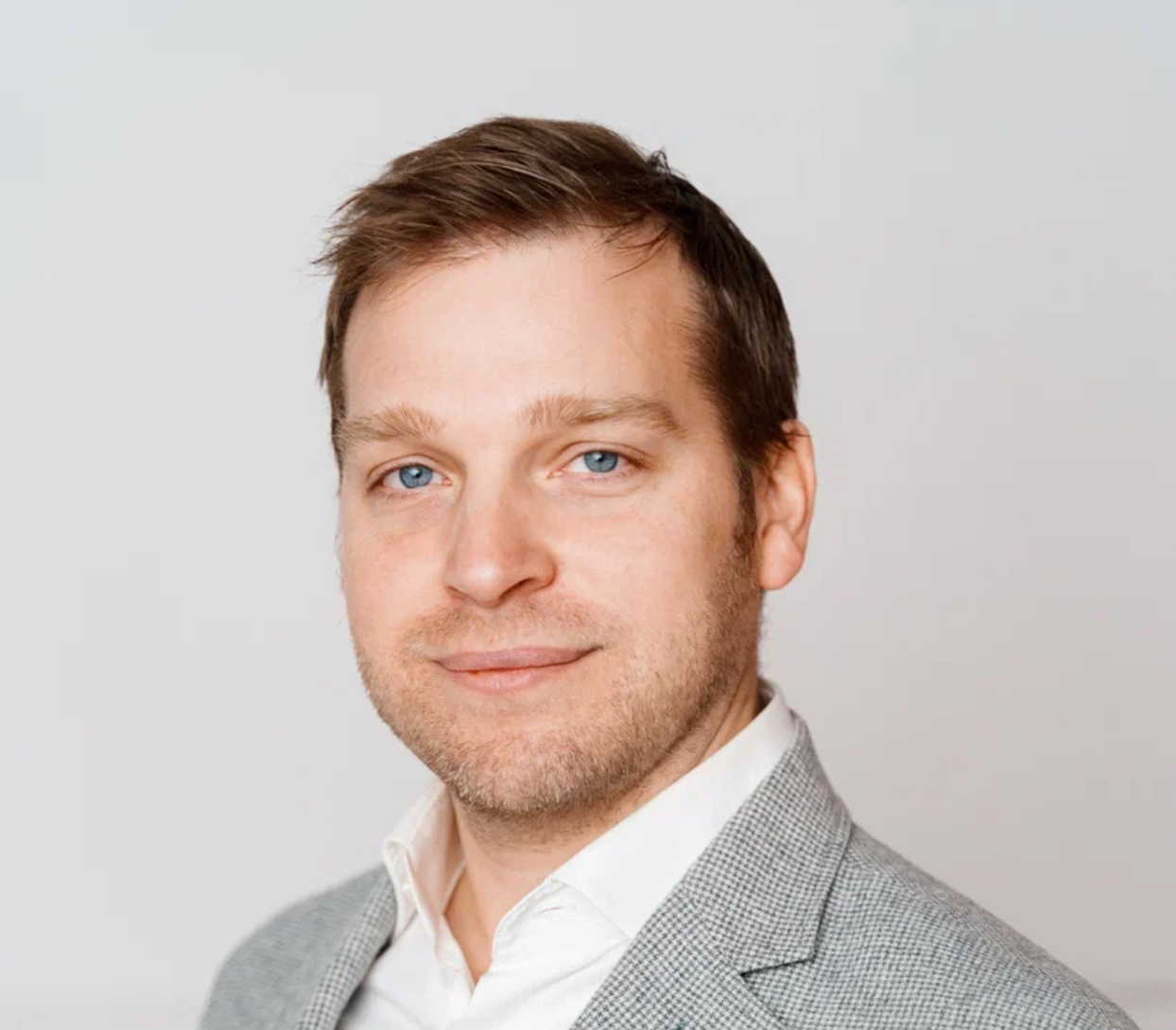}
\end{wrapfigure}
\textbf{Karl Kaberg} is a postdoc researcher at the Lab of Software and Service Engineering at Innopolis University, Russia. He completed his PhD in robotics at School of Advanced Studies Sant'Anna, Italy, in 2013 and has later worked as an independent inventor and researcher for several years. His research interests include motoric knowledge transfer, neurorobotics, model-based optimization, computational neuroscience and AI ethics.




\begin{minipage}[t]{0.15\textwidth}
\vspace{0pt}
\includegraphics[width=\linewidth]{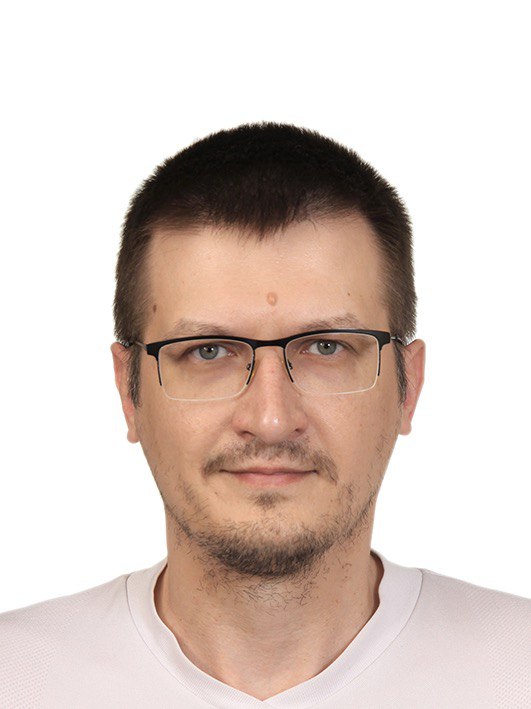}
\end{minipage}
\begin{minipage}[t]{0.8\textwidth}
\vspace{0pt}
\textbf{Roman Nabatchikov} is an IT engineer, freelancer, and entrepreneur with extensive experience in web development, website promotion, and digital marketing campaigns. He graduated from the Kolomna Polytechnic College specializing in Software for Computational Devices and Automated Systems in 2001 (with distinction). He also has a Master’s degree in Organizational Management from the Moscow State Open University (2007). Currently, he specializes in website creation, social media and SEO promotion, as well as fine-tuning of contextual and media advertising. An expert in traffic arbitrage and advertising campaign analytics, Roman is currently focused on exploring and implementing cutting-edge technologies such as LLMs (large language models) and neural networks, staying at the forefront of technological innovation. He is an external member of MMCP lab at HSE University.
\end{minipage}

\end{document}